\documentclass{article}

\usepackage{PRIMEarxiv}

\usepackage[utf8]{inputenc}
\usepackage[T1]{fontenc}
\usepackage{hyperref}
\usepackage{url}
\usepackage{booktabs}
\usepackage{amsfonts}
\usepackage{amsmath}
\usepackage{nicefrac}
\usepackage{microtype}
\usepackage{fancyhdr}
\usepackage{graphicx}
\usepackage{algorithm}
\usepackage{algorithmic}
\usepackage{array}
\usepackage{multirow}
\usepackage{threeparttable}
\usepackage{makecell}
\usepackage{wrapfig}

\graphicspath{{figures/}}

\title{Visualizing Uncertainty: Spatial Maps of Missing and Conflicting Evidence in Deep Learning
}

\author{
  Dong Hyun Jeong \\
  University of the District of Columbia \\
  Department of Computer Science \\
  Washington, DC, USA \\
  \texttt{djeong@udc.edu} \\
  \And
  Feng Chen \\
  University of Texas at Dallas \\
  Department of Computer Science \\
  Richardson, TX, USA \\
  \texttt{feng.chen@utdallas.edu} \\
  \And
  Jin-Hee Cho \\
  Virginia Tech \\
  Department of Computer Science \\
  Blacksburg, VA, USA \\
  \texttt{jicho@vt.edu} \\
  \And
  Lance M. Kaplan \\
  U.S. Army DEVCOM \\
  Army Research Laboratory \\
  Adelphi, MD, USA \\
  \texttt{lance.m.kaplan.civ@army.mil} \\
  \And
  Audun Jøsang \\
  University of Oslo \\
  Department of Informatics \\
  Oslo, Norway \\
  \texttt{josang@ifi.uio.no} \\
  \And
  Soo-Yeon Ji \\
  Bowie State University \\
  Department of Computer Science \\
  Bowie, MD, USA \\
  \texttt{sji@bowiestate.edu} \\
}

\begin{document}
\maketitle

\begin{abstract}
Understanding when and why deep neural networks are uncertain is crucial for deploying reliable machine learning systems in safety-critical domains. While existing uncertainty quantification methods provide scalar measures of model confidence, they offer limited insight into which spatial regions of an input contribute to different types of uncertainty. We propose a novel visualization framework, Uncertainty Activation Map (UAM), that combines Evidential Deep Learning (EDL) with Full-Gradient Class Activation Mapping (FullGrad) to generate interpretable spatial uncertainty activation maps. Our approach distinguishes between two fundamental types of uncertainty: vacuity, representing lack of evidence, and dissonance, capturing conflicting evidence between competing hypotheses. By leveraging the complete gradient decomposition property of FullGrad and the principled uncertainty quantification of Subjective Logic, our method produces theoretically grounded visualizations that highlight specific image regions responsible for model uncertainty. With this framework, vacuity and dissonance activation maps are generated by computing belief-weighted attributions, enabling identification of where models lack knowledge versus where they encounter ambiguous evidence. Extensive evaluations across multiple benchmark datasets demonstrate that the proposed framework effectively addresses the critical gap between uncertainty quantification and explainability, providing intuitive visual feedback to assess model reliability in complex visual recognition tasks.
\end{abstract}


\section{Introduction}

Deep neural networks have achieved remarkable performance across diverse computer vision tasks, yet their deployment in safety-critical applications remains constrained by fundamental questions of reliability and interpretability. While these models can produce highly accurate predictions, understanding when they are uncertain and why they make specific decisions remains challenging. Both quantifying uncertainty and explaining model behavior are essential for building reliable systems, yet existing methods typically address only one of these aspects.

Uncertainty quantification methods, including Bayesian neural networks~\cite{mackay1992practical, neal2012bayesian}, Monte Carlo dropout~\cite{gal2016dropout}, and ensemble approaches~\cite{lakshminarayanan2017simple}, provide scalar measures of model confidence but offer limited spatial insight into which input regions contribute to uncertainty. Conversely, visualization techniques such as Class Activation Mapping (CAM)~\cite{zhou2016learning} and Gradient-weighted CAM (Grad-CAM)~\cite{selvaraju2017grad} highlight important regions for class predictions but do not explicitly address uncertainty. This separation leaves practitioners without tools to answer critical questions: Which parts of an input image cause the model to be uncertain? Does uncertainty arise from missing evidence or from conflicting evidence between competing hypotheses? How do different spatial regions contribute to different types of uncertainty?

We address this gap by proposing a unified framework, Uncertainty Activation Map (UAM), that combines Evidential Deep Learning (EDL)~\cite{NEURIPS2018_a981f2b7} with Full-Gradient Class Activation Mapping (FullGrad)~\cite{srinivas2019full} to generate interpretable spatial uncertainty visualizations. Grounded in Dempster-Shafer Theory (DST) and Subjective Logic (SL)~\cite{Josang2018}, EDL provides a principled approach to uncertainty quantification. SL distinguishes two types of uncertainty: vacuity, representing lack of evidence when the model has insufficient information to form confident beliefs, and dissonance, capturing conflicts when the model encounters competing evidence supporting multiple hypotheses. Hu et al.~\cite{hu2021multidimensional} explicitly leveraged vacuity and dissonance to distinguish out-of-distribution (OOD) samples. While EDL-based approaches quantify these uncertainties as scalar values, our framework extends them into spatial heatmaps that localize uncertainty sources within input images. UAM consists of two complementary components: the Vacuity Activation Map, which highlights regions where the model lacks evidence through belief-weighted attribution aggregation, and the Dissonance Activation Map, which identifies locations with conflicting evidence via pairwise analysis of competing class attributions. These methods maintain consistency with SL principles while providing intuitive visual feedback about different uncertainty sources.

Our choice of FullGrad as the visualization foundation is motivated by its theoretical completeness. Unlike traditional gradient-based methods that suffer from non-uniqueness problems due to softmax shift-invariance~\cite{srinivas2021rethinking}, FullGrad provides exact output decomposition for ReLU networks by accounting for both input gradient contributions and bias gradient contributions. This decomposition property ensures that attributions capture the complete influence of network parameters on predictions. This completeness is essential for uncertainty visualization because uncertainty represents a holistic property of the model's overall confidence across all classes, not merely a class-specific attribution. When visualizing OOD inputs where class-specific gradients may become unreliable, the theoretical consistency provided by FullGrad ensures that our uncertainty visualizations remain aligned with the underlying evidential framework. Although the UAM framework is best suited to FullGrad, it can be integrated with other gradient-based techniques, such as Integrated Gradients (IG) and SHAP. We performed an extensive evaluation comparing these approaches to demonstrate the advantages of the framework.

Our contributions are threefold. First, we propose a framework that unifies evidential uncertainty quantification with gradient-based visualization to produce spatial uncertainty maps that explicitly distinguish vacuity and dissonance. Second, we introduce theoretically grounded algorithms that leverage FullGrad's complete decomposition property to ensure reliable uncertainty attribution even for out-of-distribution inputs. Third, we demonstrate through extensive experiments on various datasets that our visualizations effectively reveal uncertainty patterns across diverse visual domains and image resolutions, providing interpretable explanations for model behavior in both familiar and unfamiliar scenarios.

The remainder of this paper is organized as follows. In Section \ref{sec:Previous Works}, we review related work on gradient-based visualization methods and uncertainty quantification approaches. Section \ref{sec:Quantifying Uncertainties with EDL} presents our theoretical framework, detailing how EDL and SL principles are integrated with FullGrad for uncertainty visualization. Section \ref{sec:Uncertainty Visualization} describes our experimental setup and presents comprehensive evaluations demonstrating the effectiveness of our approach across multiple datasets and uncertainty scenarios. Section \ref{sec:Discussion} discusses limitations and future directions, and Section \ref{sec:Conclusion} concludes this paper.

\section{Previous Works}
\label{sec:Previous Works}

\subsection{Saliency Mapping}\label{sec:gradcams}

Gradient-based visualization techniques highlight the most relevant features leading to classification decisions in deep neural networks. The idea originated as pixel-level saliency maps~\cite{simonyan2014} and evolved into object-level class activation mapping (CAM)~\cite{zhou2016learning}, which introduced visual explanations for neural network decisions by highlighting discriminative image regions relevant to specific classes. However, CAM's practical application was limited by architectural constraints because it required a global average pooling layer before the final fully connected layer. Thus, applying it to existing models often required network modifications and retraining. Despite these limitations, CAM established the foundation for subsequent visualization techniques by demonstrating that convolutional feature maps retain spatial information that can be leveraged for localization. Gradient-weighted CAM (Grad-CAM) addressed CAM's limitations by generalizing the approach to any convolutional neural network (CNN) architecture without requiring structural modifications or retraining~\cite{selvaraju2017grad}. By computing gradients of a target class with respect to a target convolutional layer, Grad-CAM produces coarse localization maps that highlight important regions for prediction. This technique maintains class-discriminative properties while offering broader applicability, making it one of the most widely adopted visualization methods in deep learning interpretability. The core innovation lies in using gradient information to weight each feature map, effectively capturing which features most influence the model's decision for a given class.

Building upon this foundation, numerous variants have been developed. Grad-CAM++~\cite{chattopadhay2018grad} improves object localization for multiple instances and small objects by using a weighted combination of pixel-wise gradients. Score-CAM~\cite{wang2020score} removes gradient dependence by evaluating each feature map's contribution to the target class confidence score, thus providing more stable visualizations. Guided Grad-CAM~\cite{selvaraju2017grad} combines Grad-CAM's class-discriminative localization with Guided Backpropagation's~\cite{springenberg2014striving} fine-grained detail to generate high-resolution explanations. These techniques collectively form a robust toolkit for CNN interpretability, enabling researchers and practitioners to gain insights into model behavior, diagnose failures, identify dataset biases, and develop more trustworthy AI systems~\cite{ribeiro2016should}.

FullGrad~\cite{srinivas2019full} extends the gradient-based visualization by focusing on activation maps with respect to the input image rather than the output class. While traditional methods like Grad-CAM primarily highlight regions relevant to a specific output class, FullGrad provides a more comprehensive understanding of how the entire network processes the input data. By computing gradients through all layers back to the input space, this approach captures feature sensitivities across the full network depth, revealing how early convolutional layers respond to input features before information is aggregated in later layers. This input-focused perspective is particularly valuable for analyzing how transformations of the input propagate through the network, identifying potential vulnerabilities to adversarial attacks~\cite{goodfellow2014explaining}, and understanding feature extraction processes in complex architectures. FullGrad thus complements output-focused methods by providing insights into the complete input-to-output information flow within the network~\cite{srinivas2019full, chattopadhay2018grad}.

\subsection{Uncertainty Visualization in Deep Learning}
\label{sec:Uncertainty Visualization in Deep Learning}

While activation mapping techniques focus on explaining model decisions, uncertainty visualization in CNN predictions addresses a complementary aspect of model interpretability. Uncertainty can be categorized into epistemic (model) and aleatoric (data) uncertainty. Monte Carlo Dropout (MC-Dropout)~\cite{gal2016dropout} has emerged as a popular approach for estimating epistemic uncertainty, which captures the model's ignorance about the data distribution. By activating dropout layers during inference, MC-Dropout approximates Bayesian posterior distributions via multiple stochastic forward passes, generating a distribution of predictions. The variance across these predictions can be visualized as heatmaps to identify regions of epistemic uncertainty~\cite{kendall2017uncertainties}, with high-variance areas indicating where the model lacks sufficient training data or encounters unfamiliar patterns.

Other approaches for visualizing epistemic uncertainty include Bayesian Neural Networks~\cite{blundell2015weight}, which provide uncertainty maps by sampling from posterior weight distributions, and deep ensemble methods~\cite{lakshminarayanan2017simple} that visualize disagreement among multiple independently trained models. For aleatoric uncertainty, which represents inherent data noise or ambiguity, heteroscedastic neural networks~\cite{kendall2017uncertainties} can be employed to predict input-dependent variance alongside the primary output, enabling visualization of regions with intrinsic ambiguity. Some researchers have also developed methods to decompose and separately visualize epistemic and aleatoric uncertainty components~\cite{depeweg2018decomposition}, providing a more nuanced understanding of prediction reliability.

These uncertainty visualization techniques complement activation-mapping methods by addressing distinct aspects of model interpretability. While CAM and Grad-CAM highlight features influencing decisions, uncertainty visualization reveals where those decisions may be unreliable. This combination is particularly valuable in safety-critical applications like medical imaging~\cite{leibig2017leveraging} and autonomous driving~\cite{feng2018towards}, where understanding model confidence is essential. Visualizing uncertainty through heatmaps provides intuitive interfaces for non-experts to assess model reliability, supporting the responsible deployment of deep learning systems and potentially identifying when human intervention is necessary in automated decision-making processes.

\subsection{Uncertainty-Aware Saliency and Attribution Methods}\label{sec:uncertainty-saliency}

Recent research has recognized that saliency maps, while effective for explaining classification decisions, can model parameters, training data, and inference conditions. This has motivated the development of uncertainty-aware saliency methods that incorporate uncertainty estimates into the visualization process.

Chakraborty et al.~\cite{chakraborty2021augmenting} proposed augmenting traditional saliency maps with uncertainty estimates by computing both mean and variance saliency maps across ensembles of Bayesian Neural Networks. This approach enables practitioners to distinguish between consistently important features (low variance) and those whose importance fluctuates across the model ensemble (high variance), providing insight into the stability of explanations. By visualizing both the expected saliency and its uncertainty, this method reveals which regions are reliably important and which are less so, with their importance varying across model instantiations.

Wang et al.~\cite{wang2023gradient} introduced gradient-based uncertainty attribution for explainable Bayesian deep learning, which decomposes prediction uncertainty into contributions from individual input features. Their approach integrates uncertainty quantification into gradient-based attribution methods to identify input regions that contribute to predictive uncertainty in Bayesian neural networks. Salvi et al.~\cite{salvi2024explainability} further explored the integration of explainability and uncertainty in healthcare applications, demonstrating how combining saliency-based explanations with uncertainty estimates can enhance model interpretability and support clinical decision-making.

While these methods quantify uncertainty within Bayesian inference frameworks or combine existing explainability and uncertainty techniques, our proposed approach distinguishes and spatially localizes two fundamentally different types of uncertainty: vacuity (lack of evidence) and dissonance (conflicting evidence), providing richer semantic interpretations of uncertainty sources in the input space.

\section{Uncertainty Activation Map Framework}
\label{sec:Uncertainty Activation Map}

We propose a unified framework, Uncertainty Activation Map (UAM), to generate interpretable spatial uncertainty visualizations using EDL and SL. The framework has two components: the Vacuity Activation Map and the Dissonance Activation Map.

\subsection{Quantifying Uncertainties with EDL}
\label{sec:Quantifying Uncertainties with EDL}

In Deep Learning, various uncertainty quantification methods have been proposed. Bayesian neural networks~\cite{mackay1992practical, neal2012bayesian} use probability distributions over network weights rather than point estimates, enabling the model to capture epistemic uncertainty arising from limited training data or model misspecification. Since exact Bayesian inference is computationally intractable in deep neural networks, approximation techniques have been developed, such as variational inference~\cite{blundell2015weight, graves2011practical} and Markov Chain Monte Carlo methods~\cite{welling2011bayesian}. Ensemble methods~\cite{lakshminarayanan2017simple, fort2019deep} train multiple models independently with different random initializations, training data subsets (bootstrap aggregating), or architectures, using the variance or disagreement among predictions as an uncertainty measure. As discussed in Section \ref{sec:Uncertainty Visualization in Deep Learning}, MC-Dropout~\cite{gal2016dropout, gal2016uncertainty} offers a computationally efficient alternative by treating dropout as a Bayesian approximation. EDL~\cite{NEURIPS2018_a981f2b7, sensoy2018evidential} provides an alternative framework by modeling uncertainty through higher-order distributions. Calibration methods~\cite{guo2017calibration, kumar2019verified, minderer2021revisiting} adjust prediction confidence to better align with empirical probabilities.

We employ EDL~\cite{NEURIPS2018_a981f2b7} to quantify uncertainties. It is based on the Dempster-Shafer Theory (DST) of Evidence, which assigns belief masses to sets of classes rather than probabilities to single classes as in traditional probability theory. This framework represents uncertainty through belief masses assigned to subsets of the power set of the frame of discernment \(\Theta\), where \(\Theta\) is the set of all possible outcomes or classes. While the sum of masses across all sets equals 1, individual sets may have masses that do not sum to 1. It enables explicit uncertainty representation through a dedicated uncertainty mass. Traditional DST lacks a mechanism to directly learn these mass assignments from raw data, historically requiring either heuristic rule-sets or the combination of multiple independent evidence sources using Dempster's rule of combination. In contrast, EDL parameterizes a Dirichlet distribution with a single neural network, enabling it to quantify epistemic uncertainty within a highly efficient, end-to-end learning framework without the computational overhead of data fusion or ensemble modeling.

EDL addresses this limitation by utilizing SL~\cite{Josang2018} to reason under uncertainty. Unlike DST, SL provides a framework for quantifying uncertainty based on individual model predictions without requiring multiple models. SL extends traditional probability theory by introducing the concept of an ``opinion,'' which consists of belief, disbelief, base rate, and uncertainty components. In SL, two types of uncertainty are distinguished: vacuity and dissonance. Vacuity, represented as `uncertainty mass,' indicates a lack of evidence, while dissonance occurs when evidence supports multiple conflicting hypotheses or is ambiguous. In deep learning~\cite{sensoy2018evidential}, vacuity represents a lack of confidence in predictions due to output probabilities being spread thinly across classes, making it difficult to determine the actual output with certainty.

In EDL, multinomial opinions are used to determine the degree of belief (belief mass) in each possible outcome or class. For a domain \(\mathbb{X}\) with \(K\) classes, each multinomial opinion is represented by \(\omega_X = (b_X, u_X, a_X)\), where \(b = [b_1, ..., b_K]\) denotes belief mass for each class, \(u\) is a scalar representing overall uncertainty (vacuity of evidence), and \(a = [a_1, ..., a_K]\) indicates the base rate distribution over \(\mathbb{X}\). The projected probability distribution is given by:

\begin{equation}\label{eqn:p}
\textbf{p}_k = b_k + a_k u, \quad \sum_{k=1}^{K}\textbf{p}_k = 1, \quad \forall_k\in \mathbb{X}.
\end{equation}

In neural networks, where prior knowledge for each class is typically unknown, the base rate is set to \(a_k = \frac{1}{K}\) to ensure equal prior beliefs across all classes, leading to a uniform distribution of uncertainty.

EDL uses the Dirichlet distribution to model the uncertainty associated with predictions. The Dirichlet distribution is a multivariate probability distribution defined over the simplex, parameterized by \(K\) parameters \(\alpha_1, \alpha_2, ..., \alpha_K\) for \(K\) classes. The relationship between belief masses and the Dirichlet distribution is established through the parameterization \(\alpha_k = e_k + 1\), where \(e_k \ge 0\) represents the evidence for class \(k\). The evidence \(e = [e_1, e_2, ..., e_K]\) corresponds to the neural network output before being transformed into probabilities, representing the confidence in the classified outcome. For a given input \(x\) and network parameters \(\Theta\), the evidence is computed as:

\begin{equation}\label{eqn:e}
e = f(\mathrm{x} | \Theta), \quad e = [e_1, e_2, ..., e_K], \quad e_k \geq 0,
\end{equation}
where different activation functions, such as ReLU or Softplus, can be used at the final layer to ensure non-negativity.

The Dirichlet strength is defined as \(S = \sum_{k=1}^{K}\alpha_k = \sum_{k=1}^{K}(e_k + 1)\). The constant term of 1 added to each evidence value acts as regularization, ensuring that when no evidence exists (\(e_k = 0\) for all \(k\)), the uncertainty becomes maximal (\(u = 1\)). Belief masses and uncertainty are computed as:

\begin{equation}\label{eqn:b}
b_k = \frac{e_k}{S}, \quad u = \frac{K}{S}.
\end{equation}

The overall opinion satisfies:

\begin{equation}\label{eqn:u}
u + \sum_{k=1}^{K}b_k = 1
\end{equation}
ensuring that the resulting opinion is valid according to SL principles. The expected probability for each class is then \(\textbf{p}_k = b_k + a_k u = \frac{e_k + 1}{S}\), which can be interpreted as the output projected probability distribution.

\subsection{Uncertainty Visualization}
\label{sec:Uncertainty Visualization}

To generate uncertainty visualizations, we implement an approach based on FullGrad~\cite{srinivas2019full} and SL~\cite{josang2016subjective}. This method leverages the mathematical properties of CNNs with ReLU activations to decompose network outputs and identify regions of uncertainty.

\subsubsection{Why FullGrad?}

Among the gradient-based visualization methods discussed in Section \ref{sec:gradcams}, we employ FullGrad for our uncertainty visualization approach. While traditional methods like Grad-CAM are designed to highlight regions relevant to specific output classes, our framework requires a fundamentally different approach because uncertainty is a holistic property that encompasses the model's overall confidence across all classes rather than attribution to a single class.

As demonstrated by Srinivas and Fleuret~\cite{srinivas2021rethinking}, standard logit-gradients suffer from a fundamental non-uniqueness problem. For a neural network function where \(\tilde{f}_i(\cdot) = f_i(\cdot) + g(\cdot)\) for any arbitrary function \(g\), we obtain \(\nabla_x\tilde{f}_i(\cdot) = \nabla_xf_i(\cdot) + \nabla_xg(\cdot)\), while the corresponding loss values and loss-gradients remain unchanged due to the shift-invariance property of softmax. This means that individual logit gradients \(\nabla_xf_i(x)\) can be arbitrarily structured and may be uninformative about the underlying discriminative model.

FullGrad addresses this limitation through its complete decomposition property. For a ReLU neural network \(f\) with biases \(b \in \mathbb{R}^F\), the following exact decomposition holds~\cite{srinivas2019full}:

\begin{equation}
f(x; b) = \nabla_x f(x; b)^T x + \nabla_b f(x; b)^T b.
\end{equation}

This equation demonstrates that the output can be precisely represented as the sum of the input-gradient contribution \(\nabla_x f(x; b)^T x\) and the bias-gradient contribution \(\nabla_b f(x; b)^T b\). This decomposition derives from the positive homogeneity property inherent to ReLU networks~\cite{bach2015pixel}.

This completeness property is essential for our uncertainty visualization framework. Uncertainty measures, such as vacuity and dissonance, represent the model's overall confidence across all classes, not class-specific attributions. When visualizing uncertainty for out-of-distribution inputs, the completeness of attribution becomes even more important, as class-specific gradients may become increasingly unreliable. The theoretical consistency provided by FullGrad ensures that our uncertainty visualizations are aligned with the underlying EDL framework, where uncertainty measures are derived from Dirichlet distribution parameters across the entire output space.

While we demonstrate our approach using FullGrad attribution maps, the UAM framework is compatible with any gradient-based attribution method that produces spatial importance maps, including IG, SHAP, or Grad-CAM. We term these ``activation maps'' as they highlight spatial regions where specific types of uncertainty are activated or manifested in the model's decision-making process.

\subsubsection{Vacuity Activation Map}

In SL~\cite{josang2016subjective}, an opinion about a proposition is represented as a triplet \(\{b, d, u\}\) where \(b\) is belief, \(d\) is disbelief, and \(u\) is uncertainty (vacuity), with the constraint \(b + d + u = 1\). For multi-class classification, this extends to \(\sum_{k=1}^{K} b_k + u = 1\), where each \(b_k\) represents the belief mass for class \(k\). In the context of the uncertainty framework with FullGrad, we compute spatial belief distributions by weighting FullGrad attributions with their corresponding belief masses, then derive uncertainty as the complement of total belief mass.

\begin{equation}
B_{\mathrm{total}}(x,y) = \sum_{k=1}^{K} b_{k} \cdot \mathcal{N}(A_{k}^{\text{FullGrad}}(x,y)),
\end{equation}

where \(B_{\mathrm{total}}(x,y)\) represents the total belief mass at spatial location \((x,y)\) across all classes. \(K\) denotes the total number of target classes in the classification problem. The term \(b_{k}\) represents the belief mass for target class \(k\) (scalar weight from SL), and \(A_{k}^{\text{FullGrad}}(x,y)\) is the FullGrad attribution map for class \(k\) at spatial location \((x,y)\), computed using the previously described decomposition. The function \(\mathcal{N}(\cdot)\) is a normalization operator that ensures attribution maps are on comparable scales.

Uncertainty at each spatial location is then computed as:

\begin{equation}
u(x,y) = 1 - B_{\mathrm{total}}(x,y).
\end{equation}

Here, \(u(x,y)\) represents vacuity (uncertainty) at spatial location \((x,y)\), indicating areas where the model lacks confidence. The value 1 represents a complete opinion mass in the SL framework. Substituting the expression for \(B_{\mathrm{total}}(x,y)\), we obtain the mathematical formulation for vacuity FullGrad:

\begin{equation}
u(x,y) = 1 - \sum_{k=1}^{K} b_{k} \cdot \mathcal{N}(A_{k}^{\text{FullGrad}}(x,y)).
\end{equation}

This combines belief-weighted attribution aggregation with uncertainty computation. The FullGrad attributions provide a complete decomposition of network predictions into input and bias contributions. The approach ensures that high-belief regions with strong attributions contribute more to total belief, while low-belief or low-attribution regions contribute more to uncertainty.

The fundamental constraint of SL for each spatial location is maintained through:

\begin{equation}
u(x,y) + \sum_{k=1}^{K} b_k(x,y) = 1,
\end{equation}
where \(b_k(x,y)\) represents the effective belief mass for class \(k\) at location \((x,y)\) after belief-weighted FullGrad processing. This constraint ensures that uncertainty plus all class beliefs sum to unity at every spatial location, validating that the vacuity maintains mathematical consistency with SL principles.

\begin{algorithm}
\caption{Vacuity Activation Map}
\label{alg:vacuity_full_gradcam}
\begin{algorithmic}[1]
\REQUIRE FullGrad attribution maps $\{A_{k}^{\text{FullGrad}}\}$, Belief masses $\{b_{k}\}$, Number of classes $K$
\ENSURE Vacuity map $u(x,y)$
\STATE Initialize $B_{\mathrm{total}}(x,y) \leftarrow \mathbf{0}$ for all spatial locations $(x,y)$
\FOR{each target class $k = 1$ to $K$}
    \STATE $A_{\text{norm}} \leftarrow \mathcal{N}(A_{k}^{\text{FullGrad}}(x,y))$
    \STATE $B_{\mathrm{total}}(x,y) \leftarrow B_{\mathrm{total}}(x,y) + b_{k} \cdot A_{\text{norm}}(x,y)$
\ENDFOR
\STATE $u(x,y) \leftarrow 1 - B_{\mathrm{total}}(x,y)$
\RETURN Vacuity map $u(x,y)$
\end{algorithmic}
\end{algorithm}

Algorithm \ref{alg:vacuity_full_gradcam} presents the vacuity activation map generation. It takes as input FullGrad attribution maps \(\{A_{k}^{\text{FullGrad}}\}\) for each class \(k\), along with corresponding belief masses \(\{b_{k}\}\). It aggregates normalized belief-weighted attributions across classes to produce the vacuity map \(u(x,y)\), where high values indicate regions with low model confidence. This approach differs from traditional uncertainty quantification methods by incorporating SL principles directly into attribution maps. The resulting uncertainty heatmaps highlight regions where belief mass is low, indicating areas where the model has insufficient evidence to form strong opinions about class-relevant features.

\subsubsection{Dissonance Activation Map}

We introduce a dissonance activation map (Algorithm \ref{alg:dissonance_full_gradcam}), which quantifies and visualizes conflicts between attribution maps of different classes that exceed a confidence threshold. The core mathematical formulation for dissonance computation incorporates pairwise conflict analysis:

\begin{equation}
D_{\mathrm{total}}(x,y) = \frac{1}{n_c} \sum_{\substack{i,j \in \mathcal{I} \\ i < j}} \text{Bal}_{i,j}(x,y) \cdot I_{i,j}(x,y) \cdot M_{i,j}(x,y) \cdot w_{i,j}
\end{equation}
where \(\mathcal{I} = \{k : b_k > \tau\}\) is the set of classes with beliefs exceeding threshold \(\tau\), \(n_c = \binom{|\mathcal{I}|}{2}\) is the number of pairwise comparisons, \(\text{Bal}_{i,j}(x,y) = 1 - \frac{|G_i(x,y) - G_j(x,y)|}{G_i(x,y) + G_j(x,y) + \epsilon}\) is a balance function measuring attribution similarity, \(I_{i,j}(x,y) = \frac{G_i(x,y) + G_j(x,y)}{2}\) is the average importance of a spatial location, \(M_{i,j}(x,y) = \mathbb{I}(G_i(x,y) > 0) \wedge \mathbb{I}(G_j(x,y) > 0)\) is a joint activation mask, and \(w_{i,j} = \frac{b_i + b_j}{2}\) is the average belief weight for the class pair.

The normalized activation maps are computed as:

\begin{equation}
G_k(x,y) = \mathcal{N}(A_{k}^{\text{FullGrad}}(x,y)),
\end{equation}

where \(\mathcal{N}\) represents normalization to the [0,1] range. The algorithm sets a belief threshold \(\tau = 0.1\) that determines the minimum confidence required for a class to be considered in conflict analysis. For each class \(k\), it checks whether the class belief exceeds the threshold. Only classes with belief exceeding the threshold are included in the candidate set \(\mathcal{C}\).

\begin{algorithm}
\caption{Dissonance Activation Map}
\label{alg:dissonance_full_gradcam}
\begin{algorithmic}[1]
\REQUIRE FullGrad attribution maps $\{A_{k}^{\text{FullGrad}}\}$, Belief masses $\{b_k\}$, Classes $K$
\ENSURE Dissonance map $D_{\mathrm{total}}(x,y)$
\STATE Set belief threshold $\tau \leftarrow 0.1$
\STATE Initialize $D_{\mathrm{total}}(x,y) \leftarrow \mathbf{0}$ for all spatial locations $(x,y)$
\STATE Initialize candidate set $\mathcal{C} \leftarrow \emptyset$
\FOR{each target class $k = 1$ to $K$}
    \IF{$b_k > \tau$}
        \STATE $G_k(x,y) \leftarrow \mathcal{N}(A_{k}^{\text{FullGrad}}(x,y))$
        \STATE Add $(G_k, b_k, k)$ to candidate set $\mathcal{C}$
    \ENDIF
\ENDFOR
\IF{$|\mathcal{C}| < 2$}
    \RETURN $D_{\mathrm{total}}(x,y) \leftarrow \mathbf{0}$
\ENDIF
\STATE Initialize comparison counter $n_c \leftarrow 0$
\FOR{each pair $(G_i, b_i, i)$ and $(G_j, b_j, j)$ in $\mathcal{C}$ where $i < j$}
    \STATE $\text{Bal}_{i,j}(x,y) \leftarrow 1 - \frac{|G_i(x,y) - G_j(x,y)|}{G_i(x,y) + G_j(x,y) + \epsilon}$
    \STATE $M_{i,j}(x,y) \leftarrow \mathbb{I}(G_i(x,y) > 0) \wedge \mathbb{I}(G_j(x,y) > 0)$
    \STATE $I_{i,j}(x,y) \leftarrow \frac{G_i(x,y) + G_j(x,y)}{2}$
    \STATE $w_{i,j} \leftarrow \frac{b_i + b_j}{2}$
    \STATE $C_{i,j}(x,y) \leftarrow \text{Bal}_{i,j}(x,y) \cdot I_{i,j}(x,y) \cdot M_{i,j}(x,y) \cdot w_{i,j}$
    \STATE $D_{\mathrm{total}}(x,y) \leftarrow D_{\mathrm{total}}(x,y) + C_{i,j}(x,y)$
    \STATE $n_c \leftarrow n_c + 1$
\ENDFOR
\STATE $D_{\mathrm{total}}(x,y) \leftarrow \frac{D_{\mathrm{total}}(x,y)}{n_c}$
\RETURN Dissonance map $D_{\mathrm{total}}(x,y)$
\end{algorithmic}
\end{algorithm}

If the candidate set contains fewer than two classes, the algorithm returns a zero-valued dissonance map, as meaningful conflict requires at least two competing explanations. Otherwise, the algorithm proceeds to examine each pair of classes in the candidate set. For each pair, several components contributing to conflict are computed: the balance function, which measures attribution similarity with higher values indicating greater similarity; the joint activation mask, which ensures conflict is only computed where both classes show meaningful attribution; the average importance, which captures the combined significance of a spatial location; and the belief weight, which weights the conflict by the average belief of both classes. Then, the pairwise conflict is computed as the product of these components and accumulated into the total dissonance map. After processing all pairs, the accumulated conflict is averaged to produce the final dissonance map. This approach systematically identifies regions where multiple high-confidence classes attribute significant importance, indicating potential feature overlap or model confusion.

The dissonance measure maintains several important properties. First, it is bounded: \(0 \leq D_{\mathrm{total}}(x,y) \leq \max_{i,j \in \mathcal{I}} w_{i,j}\), with the upper bound depending on maximum belief weights, ensuring interpretable scaling. Second, the belief threshold filtering is mathematically justified because classes with low beliefs produce weak attribution maps that contribute negligibly to meaningful conflict, as \(\lim_{b_k \to 0} C_{i,j}(x,y) \approx 0\) for any pair involving class \(k\). This filtering focuses the dissonance computation on genuinely competing high-confidence explanations.

\section{Implementation}

\subsection{Datasets}

For a comprehensive evaluation of our uncertainty estimation framework, we employ four widely used benchmark datasets: MNIST~\cite{lecun1998mnist}, SVHN~\cite{netzer2011reading}, CIFAR10~\cite{krizhevsky2009learning}, and Imagenette~\cite{howard2019imagenette}, along with one medical imaging dataset for Alzheimer's Disease (AD) classification~\cite{Zolfaghari2025-az}. MNIST contains 70,000 grayscale images of handwritten digits (28×28 pixels) across 10 classes. Our initial testing revealed that MNIST's simplicity led to very low or near-zero dissonance values across many samples, making it difficult to determine the effectiveness of our uncertainty-estimation approach for visualizing conflicting evidence. To address this, we incorporate SVHN, which contains over 600,000 32×32 color images of house numbers with natural scene backgrounds and varying lighting conditions, resulting in a richer distribution of uncertainty patterns effective for testing our proposed approach. CIFAR10 provides 60,000 32×32 color images across 10 classes with substantial intra-class variation and diverse backgrounds. Imagenette extends our evaluation to high-resolution images (320×320 pixels) from 10 easily classified ImageNet classes. The AD dataset contains 6,400 grayscale brain MRI scans (208×176 pixels) across 4 dementia severity classes, introducing diagnostic ambiguity characteristic of real-world clinical scenarios.

\subsection{Model Training}

We employed neural network architectures tailored to each dataset and training configuration to evaluate UAM across diverse visual complexity and uncertainty conditions. For MNIST, we used a LeNet variant with two convolutional layers of 16 filters each, followed by max pooling, and three fully connected layers (256, 128, and output nodes, respectively), with dropout at 0.5. For other datasets, we employed a ResNet18 architecture~\cite{he2016deep} with four main blocks containing residual connections, batch normalization, and a dropout rate of 0.2 before classification. All models were trained for 200 epochs using the Adam optimizer~\cite{Ovadia2019, Kingma2015} with an annealing schedule for the KL divergence regularization term, achieving high accuracies (0.95--0.99) across all datasets. For robust evaluation under perturbation testing (see Section~\ref{sec:evaluation}), models were additionally trained with pseudo-OOD sample augmentation, which maintained comparable accuracy levels while improving uncertainty calibration.

\subsection{Utilizing FullGrad}

As explained above, FullGrad combines complete gradient attribution with class-discriminative visualization capabilities. It produces heatmaps by computing importance weights for each feature channel in the final convolutional layer based on global-average-pooled gradients, then applying these weights to the feature maps and using ReLU activation to highlight positive contributions. Unlike standard GradCAM, which often focuses exclusively on gradients from the last convolutional layer, FullGrad incorporates both feature map activations and bias gradients across the entire network.

In our framework, we generate attribution maps for each class in the output space. For a classification task with \(K\) classes, we obtain \(K\) separate attribution maps, each highlighting spatial regions important for predicting that specific class. To create the belief activation map, we select the attribution map corresponding to the class with the highest belief mass \(b_k\) as computed by the EDL model. This ``best belief'' attribution map represents the spatial evidence supporting the model's most confident prediction, serving as the foundation for computing vacuity and dissonance maps that reveal where and why the model exhibits uncertainty.

\subsection{Vacuity and Dissonance Activation Maps}

The Vacuity Activation Map visualizes regions contributing to model uncertainty due to a lack of evidence. In EDL, vacuity quantifies the absence of supporting evidence for each class, computed as the complement of the total belief from the Dirichlet distribution parameters. Thus, maps are visualized as heatmaps normalized to [0,1]: blue indicates strong evidence (low vacuity), transitioning through cyan and yellow to red representing maximum epistemic uncertainty (high vacuity). Unlike traditional attribution methods that highlight features for specific predictions, vacuity maps reveal regions where the model is fundamentally uncertain across all classifications.

Dissonance Activation Map captures conflicting evidence between classes, complementing vacuity. While vacuity measures lack of evidence, dissonance quantifies contradictory evidence, creating confusion between multiple hypotheses. The map identifies regions where multiple high-confidence classes attribute importance to the same locations, revealing feature overlap or ambiguous visual patterns across class boundaries.

\begin{figure}[htp]
\centering
\scriptsize
\setlength{\tabcolsep}{1pt}
\newlength{\imgwidth}
\setlength{\imgwidth}{0.11\textwidth}
\begin{tabular}{@{}ccccc@{}}
\includegraphics[width=\imgwidth]{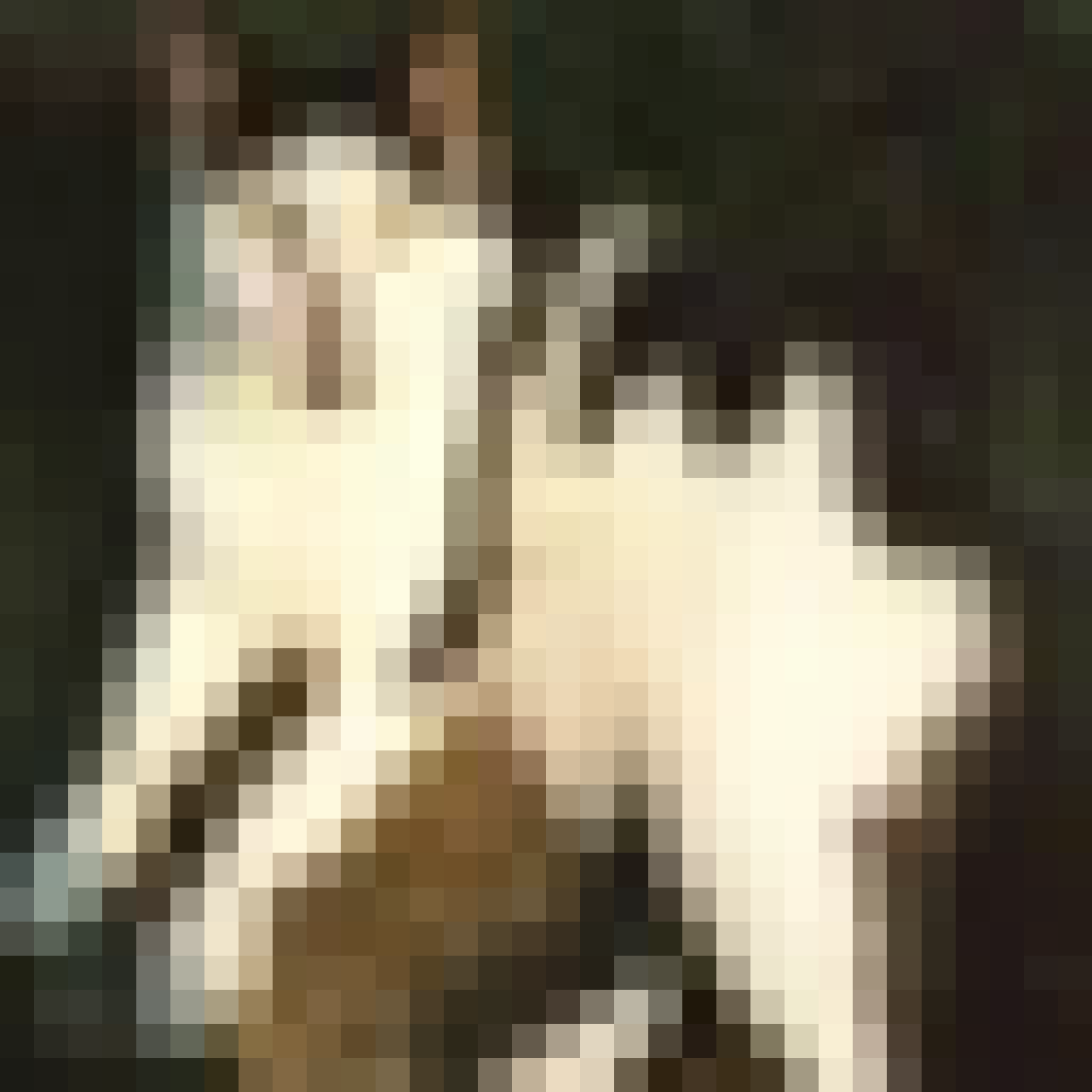} &
\includegraphics[width=\imgwidth]{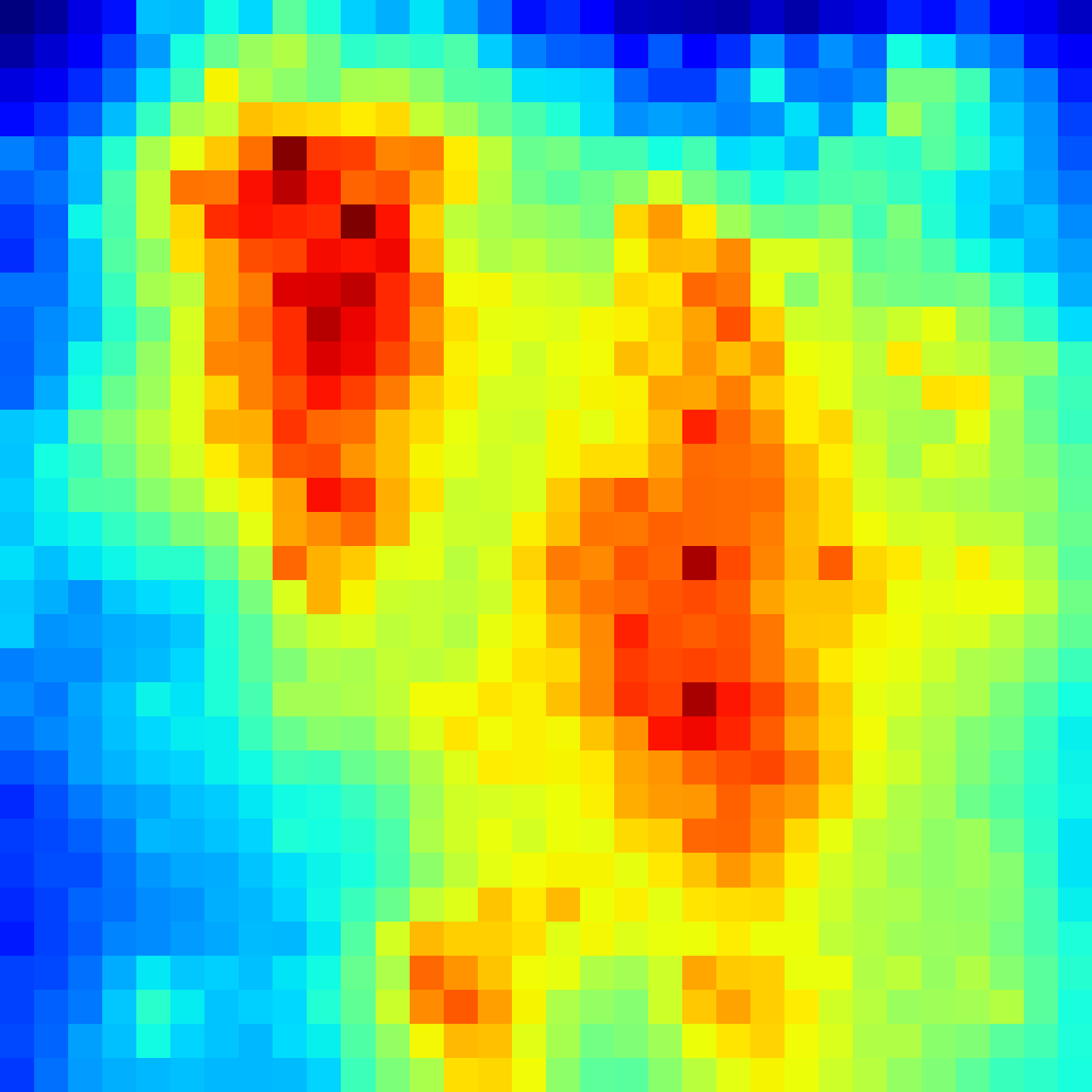} &
\includegraphics[width=\imgwidth]{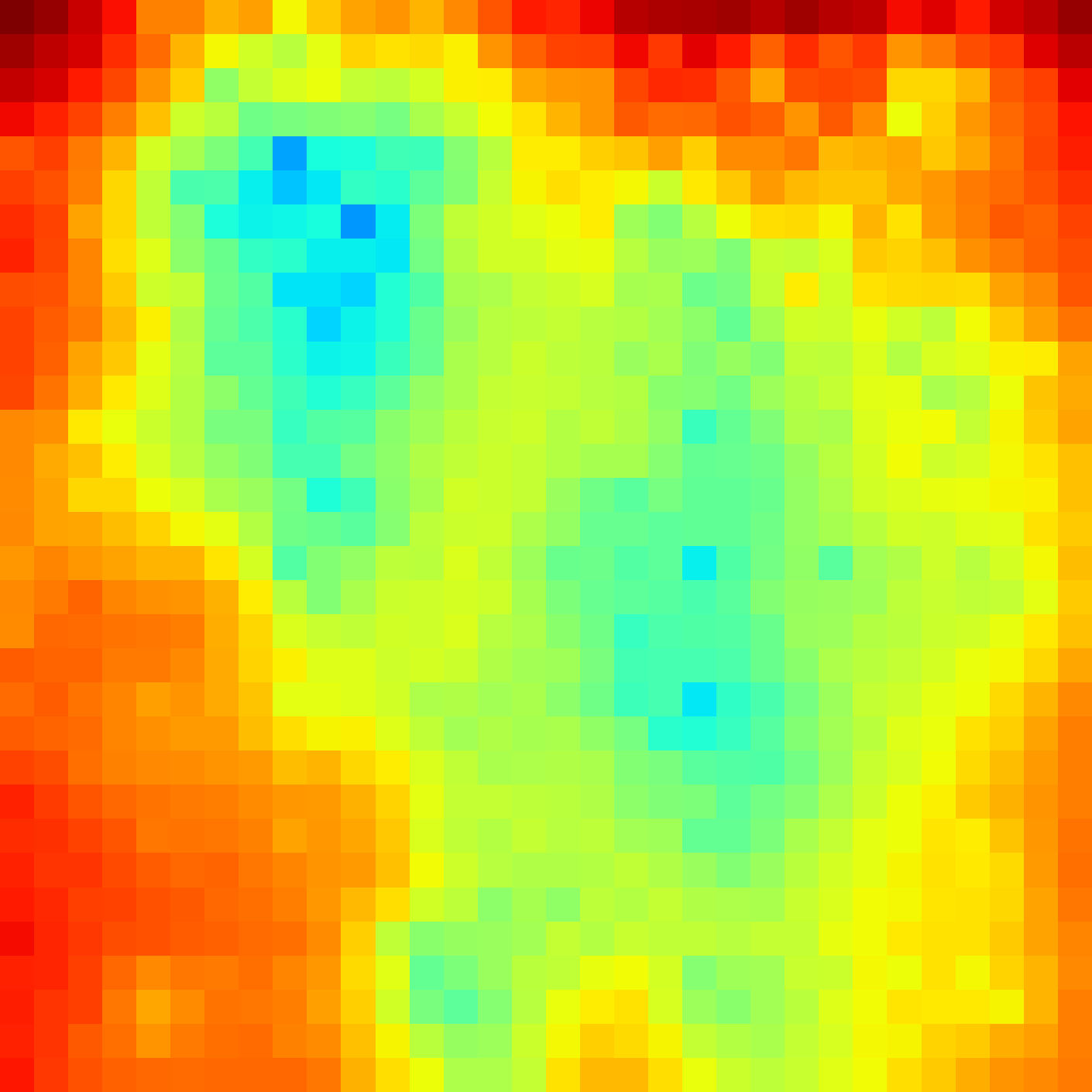} &
\includegraphics[width=\imgwidth]{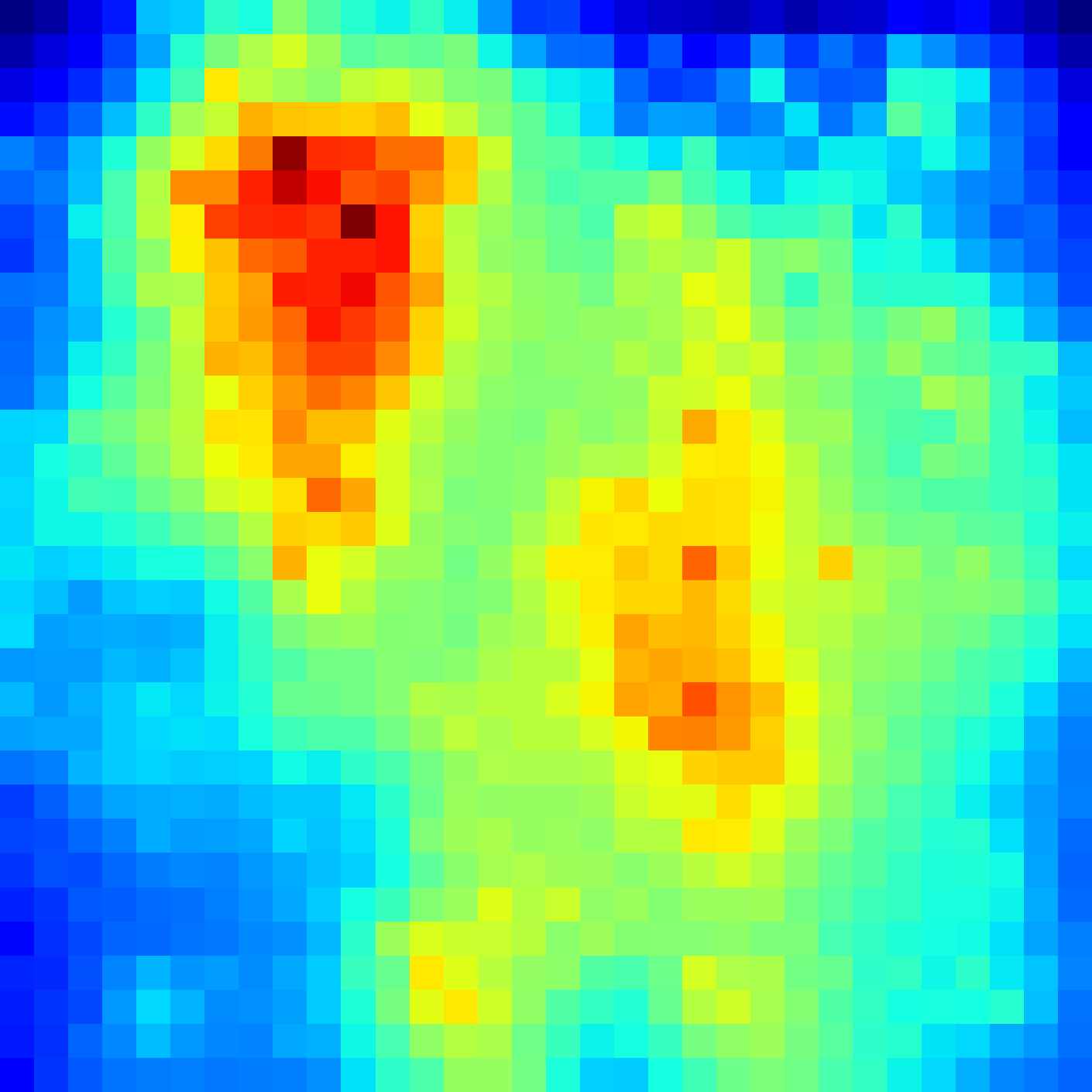} &
\includegraphics[width=\imgwidth]{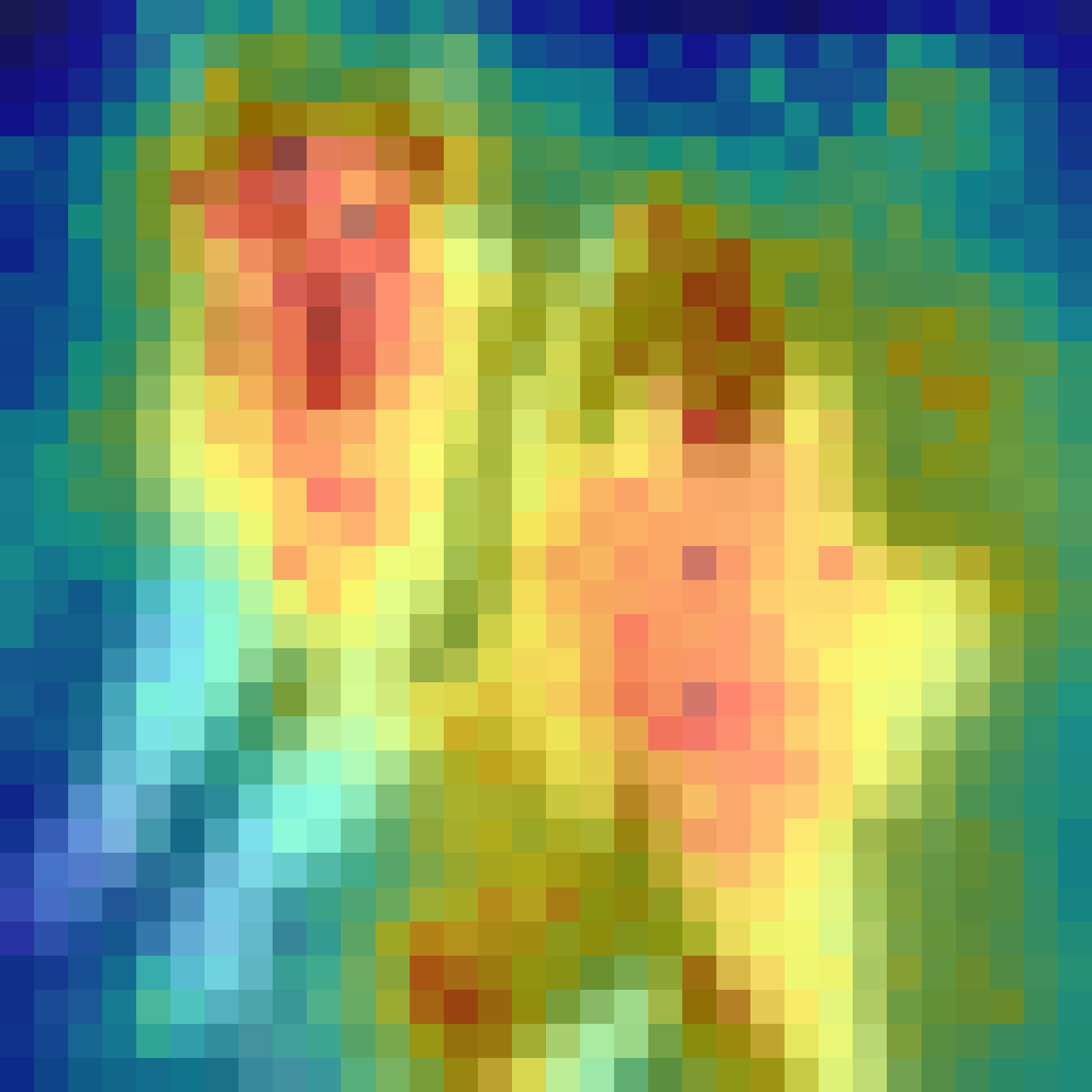} \\
{\tiny Original} & {\tiny Belief (\(b\): 0.37)} & {\tiny Vacuity (\(v\): 0.28)} & {\tiny Dissonance (\(d\): 0.70)} & {\tiny Belief (Overlay)} \\[0.3em]

\includegraphics[width=\imgwidth]{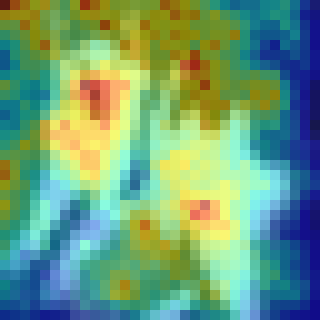} &
\includegraphics[width=\imgwidth]{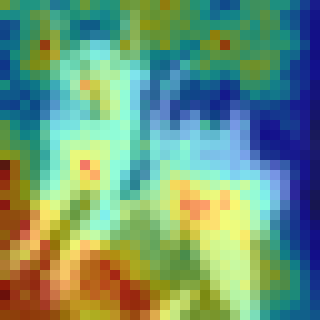} &
\includegraphics[width=\imgwidth]{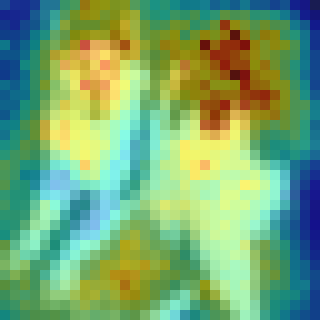} &
\includegraphics[width=\imgwidth]{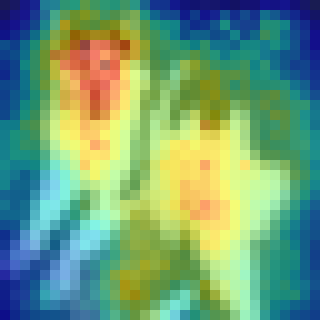} &
\includegraphics[width=\imgwidth]{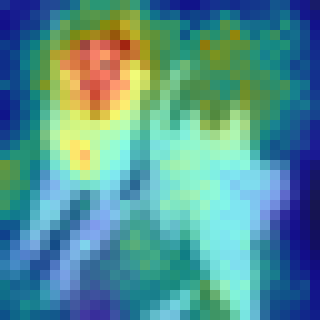} \\
{\tiny Airplane (\(b\): 0.0)} & {\tiny Automobile (\(b\): 0.0)} & {\tiny Bird (\(b\): 0.0)} & {\tiny Cat (\(b\): 0.35)} & {\tiny Deer (\(b\): 0.0)} \\[0.3em]

\includegraphics[width=\imgwidth]{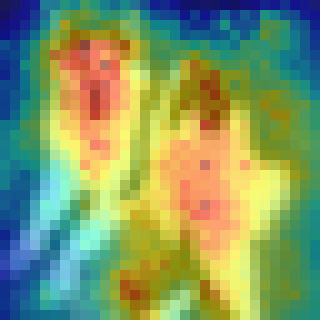} &
\includegraphics[width=\imgwidth]{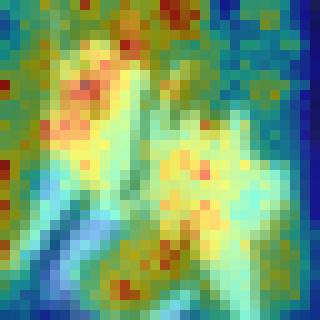} &
\includegraphics[width=\imgwidth]{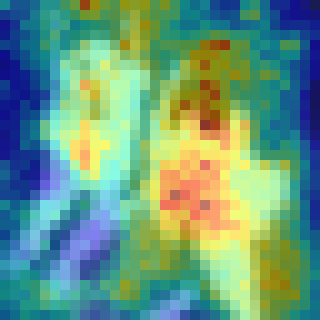} &
\includegraphics[width=\imgwidth]{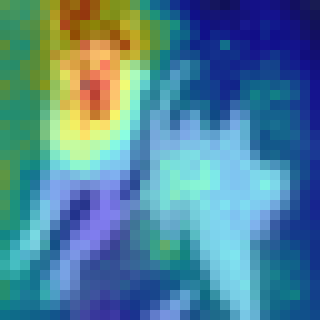} &
\includegraphics[width=\imgwidth]{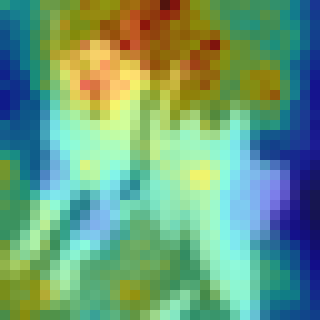} \\
{\tiny Dog (\(b\): 0.37)} & {\tiny Frog (\(b\): 0.0)} & {\tiny Horse (\(b\): 0.0)} & {\tiny Ship (\(b\): 0.0)} & {\tiny Truck (\(b\): 0.0)}
\end{tabular}
\caption{Semantic ambiguity visualization on CIFAR-10. Belief splits between cat (0.35) and dog (0.37) with high dissonance (0.70) localized to facial and upper-body regions where distinguishing features overlap between semantically related categories.}
\label{fig:fullgrad_uncertainty_cat}
\end{figure}

Fig.~\ref{fig:fullgrad_uncertainty_cat} demonstrates a cat image from CIFAR10 where belief is nearly evenly distributed between cat (0.35) and dog (0.37). The high dissonance (0.70) indicates strong cat-dog conflict, while moderate vacuity (0.28) shows some uncertainty in the evidence. The dissonance heatmap emphasizes facial and upper-body regions where distinguishing characteristics overlap. The cat attribution map focuses on facial features, while the dog map activates both facial and body areas, demonstrating differential spatial attention despite similar confidence. Class-specific overlays confirm zero belief for other classes, indicating dissonance arises specifically from confusion between semantically related animal categories rather than broader uncertainty.

\begin{figure}[htp]
\centering
\scriptsize
\setlength{\tabcolsep}{1pt}
\newlength{\imgwidthb}
\setlength{\imgwidthb}{0.11\textwidth}
\begin{tabular}{@{}ccccc@{}}
\includegraphics[width=\imgwidthb]{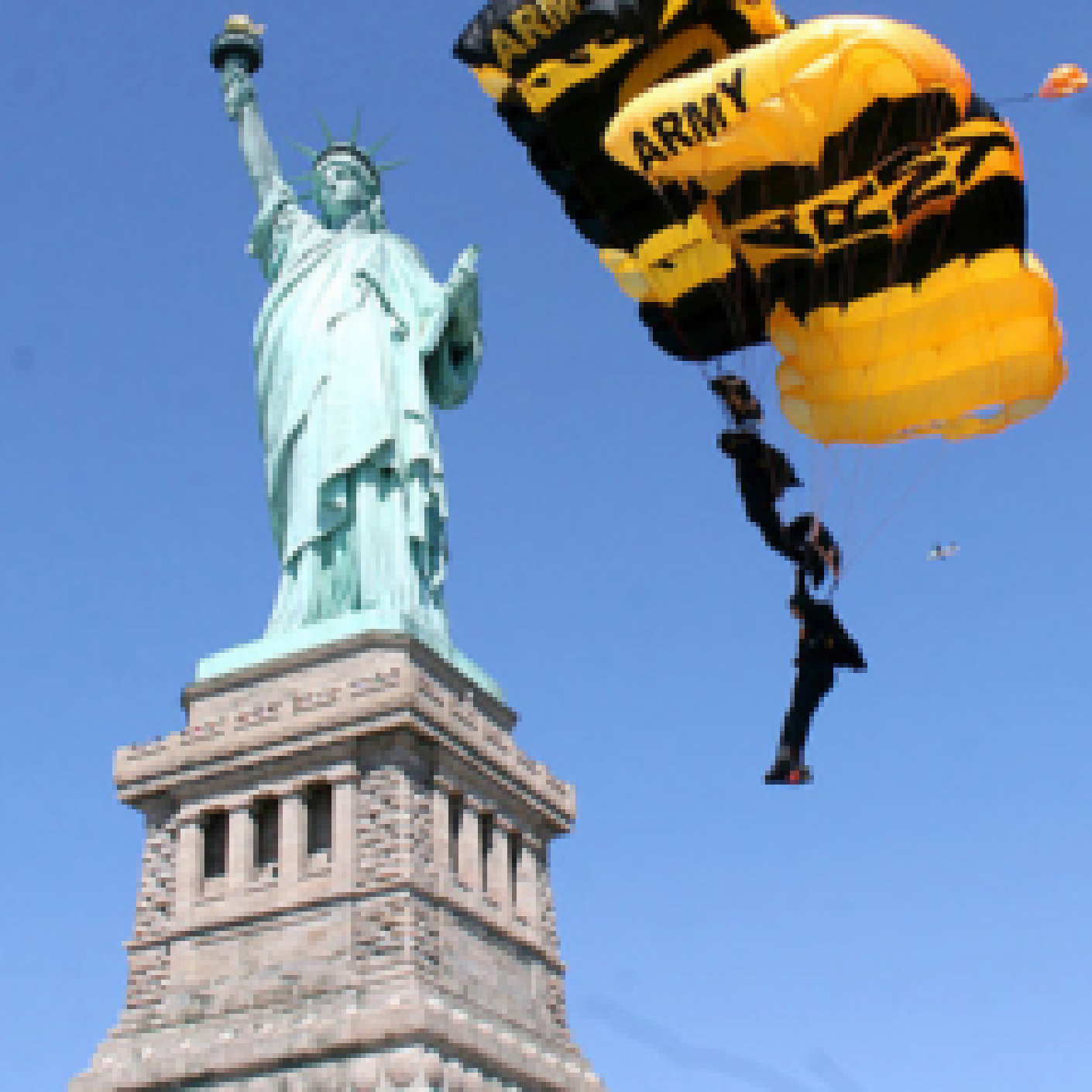} &
\includegraphics[width=\imgwidthb]{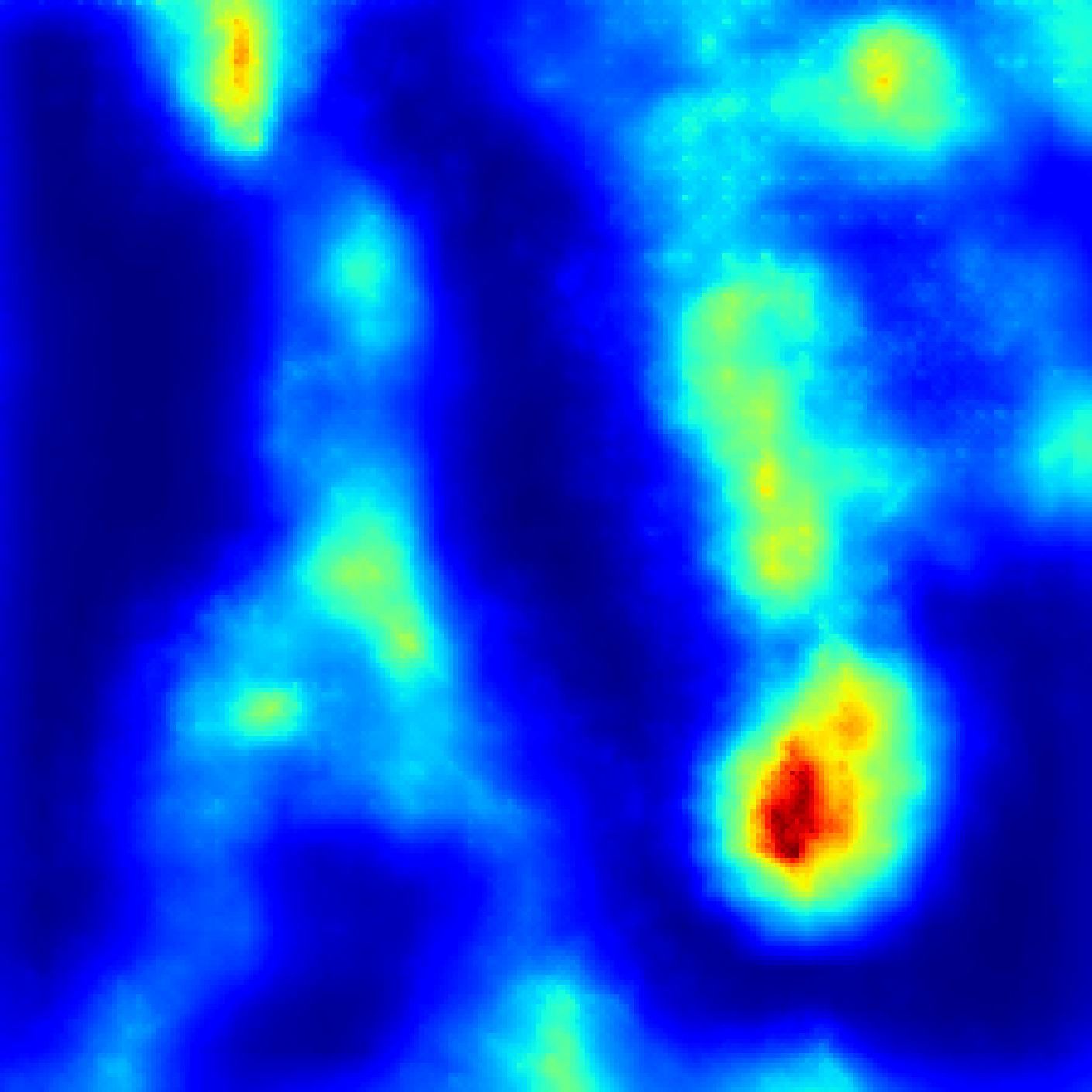} &
\includegraphics[width=\imgwidthb]{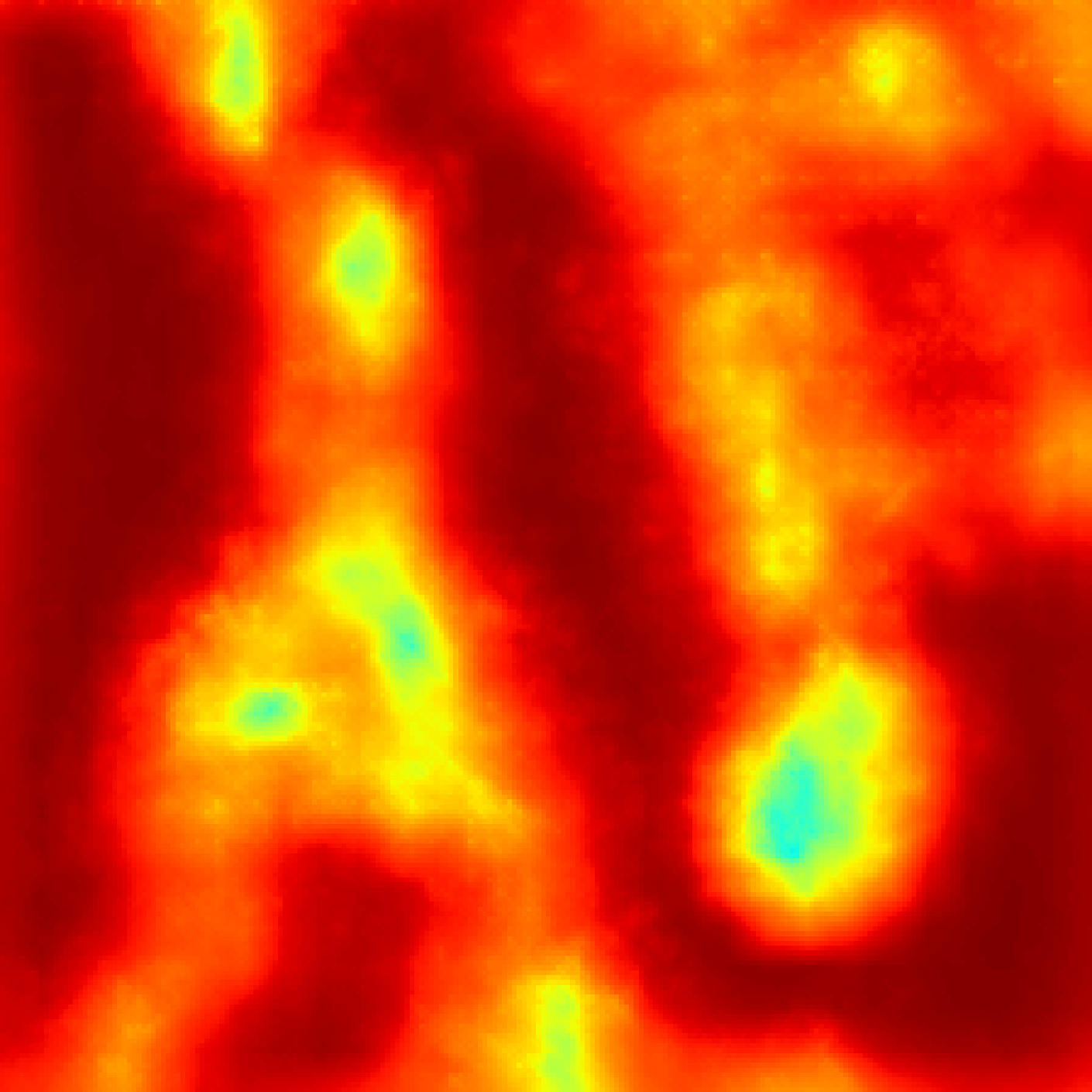} &
\includegraphics[width=\imgwidthb]{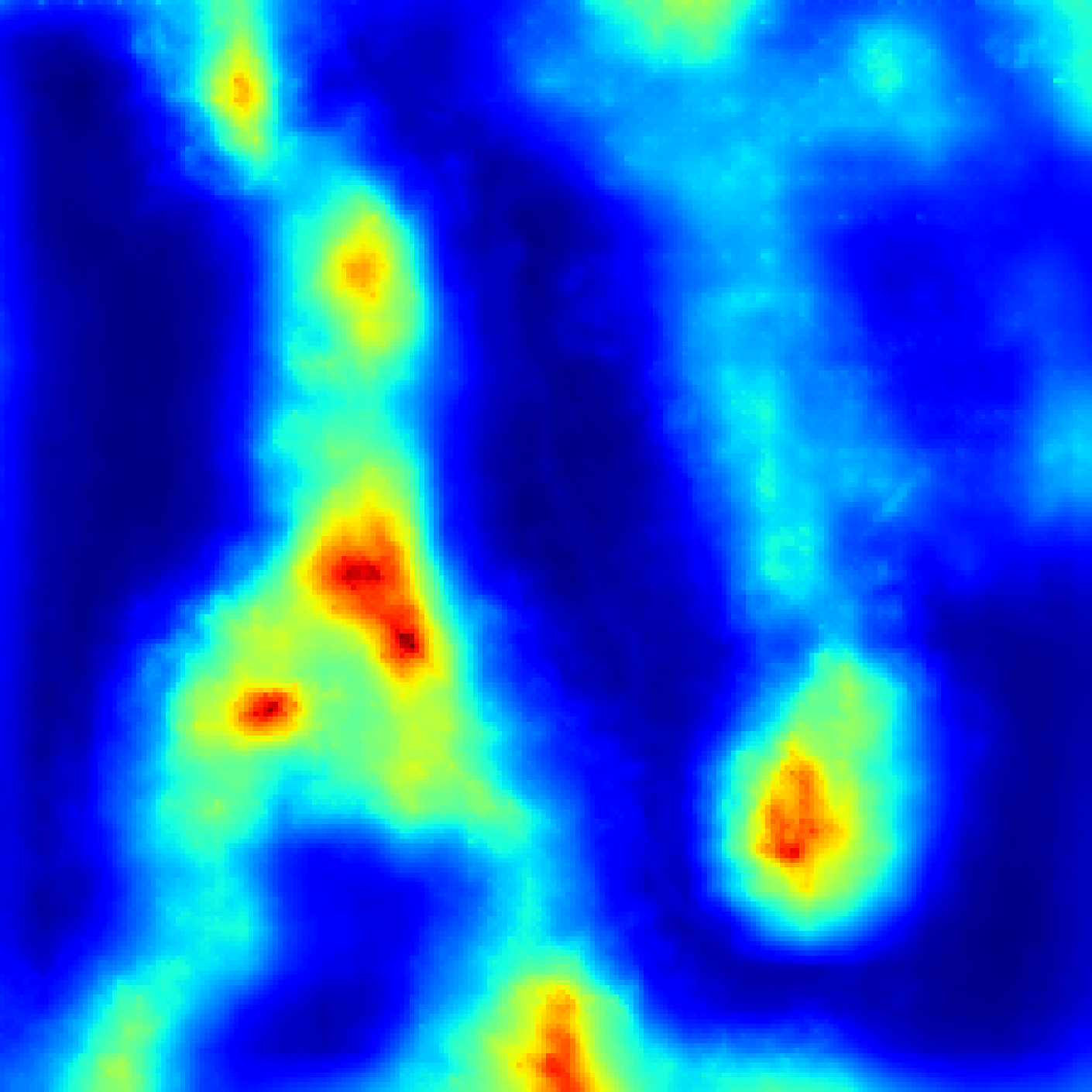} &
\includegraphics[width=\imgwidthb]{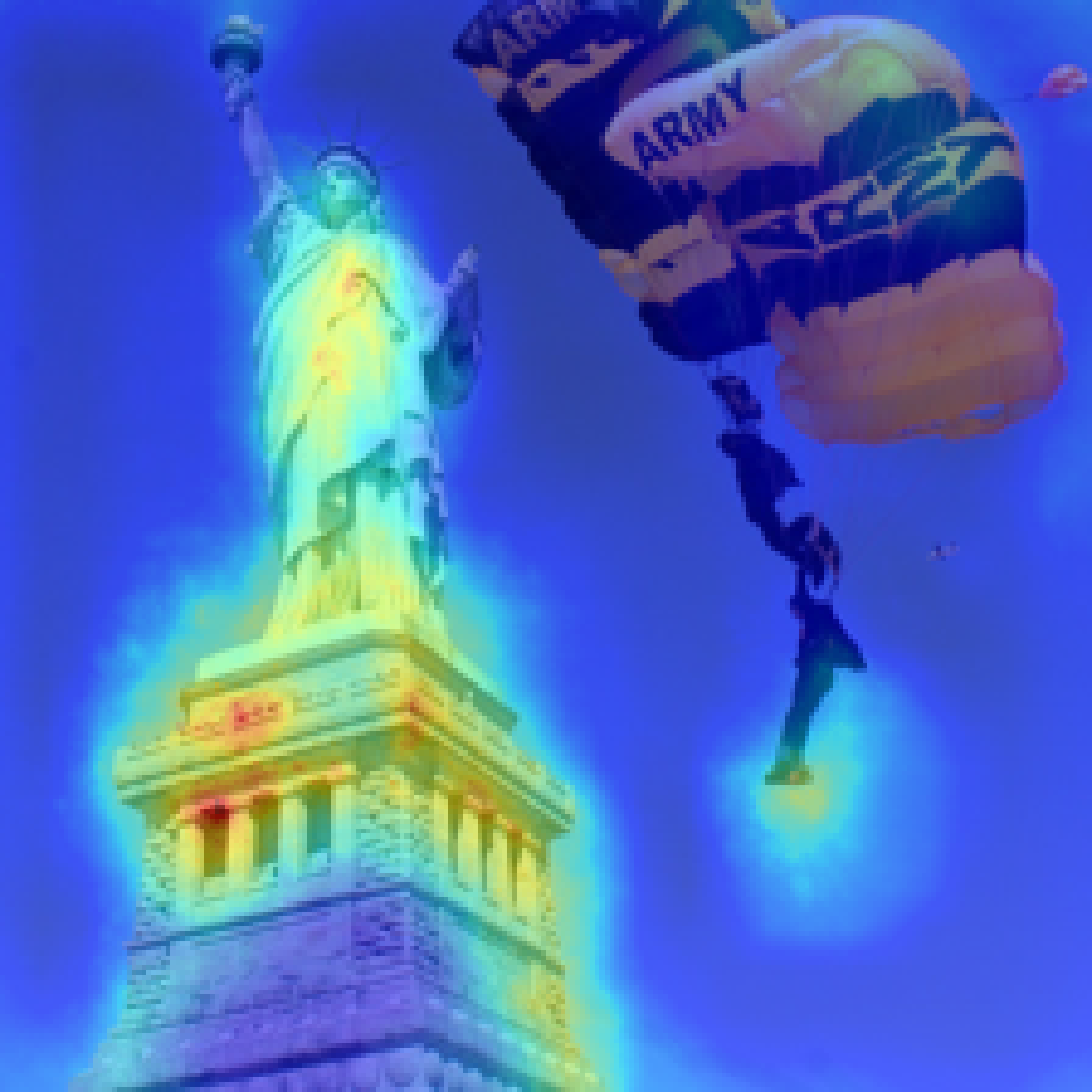} \\
{\tiny Original} & {\tiny Belief (\(b\): 0.47)} & {\tiny Vacuity (\(v\): 0.21)} & {\tiny Dissonance (\(d\): 0.65)} & {\tiny Belief (Overlay)} \\[0.3em]

\includegraphics[width=\imgwidthb]{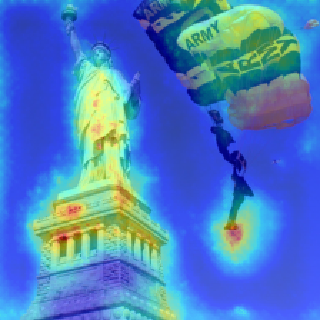} &
\includegraphics[width=\imgwidthb]{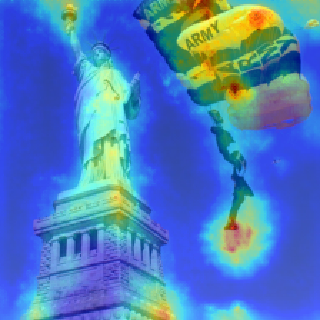} &
\includegraphics[width=\imgwidthb]{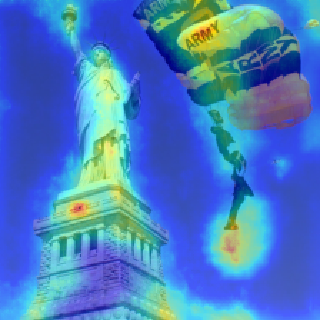} &
\includegraphics[width=\imgwidthb]{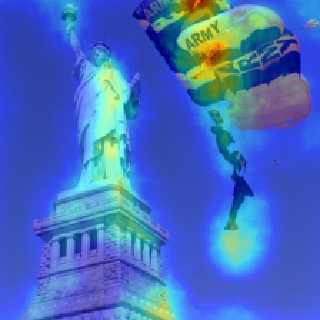} &
\includegraphics[width=\imgwidthb]{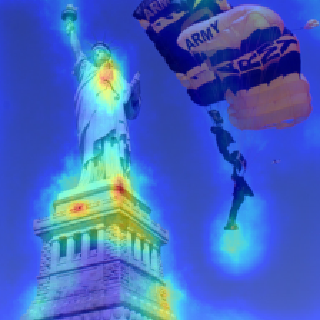} \\
{\tiny Tench (\(b\): 0.0)} & {\tiny English springer (\(b\): 0.0)} & {\tiny Cassette player (\(b\): 0.0)} & {\tiny Chain saw (\(b\): 0.0)} & {\tiny Church (\(b\): 0.33)} \\[0.3em]

\includegraphics[width=\imgwidthb]{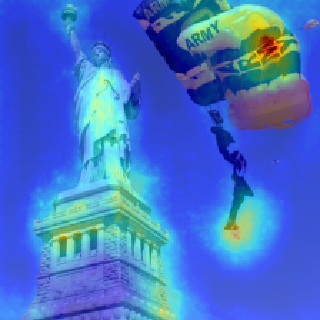} &
\includegraphics[width=\imgwidthb]{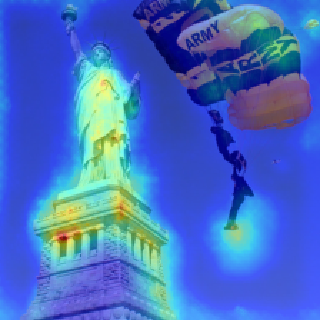} &
\includegraphics[width=\imgwidthb]{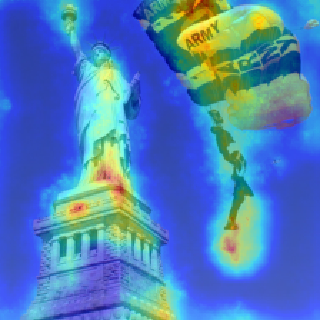} &
\includegraphics[width=\imgwidthb]{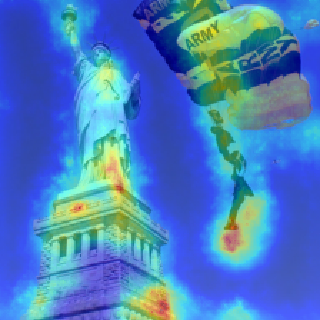} &
\includegraphics[width=\imgwidthb]{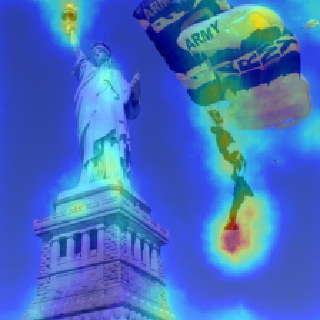} \\
{\tiny French horn (\(b\): 0.0)} & {\tiny Garbage truck (\(b\): 0.0)} & {\tiny Gas pump (\(b\): 0.0)} & {\tiny Golf ball (\(b\): 0.0)} & {\tiny Parachute (\(b\): 0.47)}
\end{tabular}
\caption{Multi-object scene analysis on Imagenette. High-resolution image of Statue of Liberty with parachutist reveals belief split between parachute (0.47) and church (0.33), with high dissonance (0.65) spatially localized to distinct visual elements.}
\label{fig:fullgrad_uncertainty_imagenette}
\end{figure}

Fig.~\ref{fig:fullgrad_uncertainty_imagenette} presents a high-resolution Imagenette image containing the Statue of Liberty and a parachutist. Without a ``statue'' class in the dataset, the model splits belief between parachute (0.47) and church (0.33). High dissonance (0.65) reflects conflict between these hypotheses. The statue's architectural features---particularly its vertical structure---resemble church spires and towers, explaining the secondary church hypothesis. Parachute activation focuses on the descending parachutist in the upper region, while church activation concentrates on the statue structure in the lower region. The dissonance heatmap reveals conflict localized to these two distinct objects, with minimal overlap between the competing interpretations. The moderate vacuity (0.21) indicates the model commits most of its belief to these two hypotheses despite their spatial separation. This spatial decoupling of evidence suggests the model independently processes distinct visual elements rather than treating the scene holistically. The case demonstrates how the model handles out-of-distribution objects by mapping them to the nearest available classes, and how uncertainty differs across spatial regions when class coverage is incomplete. Such spatially localized conflicts are particularly valuable for identifying regions where the available classes do not adequately represent the visual content.

\begin{figure}[htp]
\centering
\scriptsize
\setlength{\tabcolsep}{1pt}
\newlength{\imgwidthc}
\setlength{\imgwidthc}{0.11\textwidth}
\begin{tabular}{@{}ccccc@{}}
\includegraphics[width=\imgwidthc]{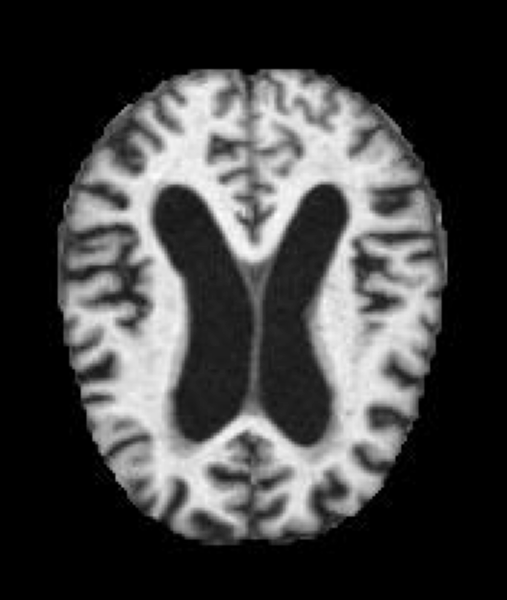} &
\includegraphics[width=\imgwidthc]{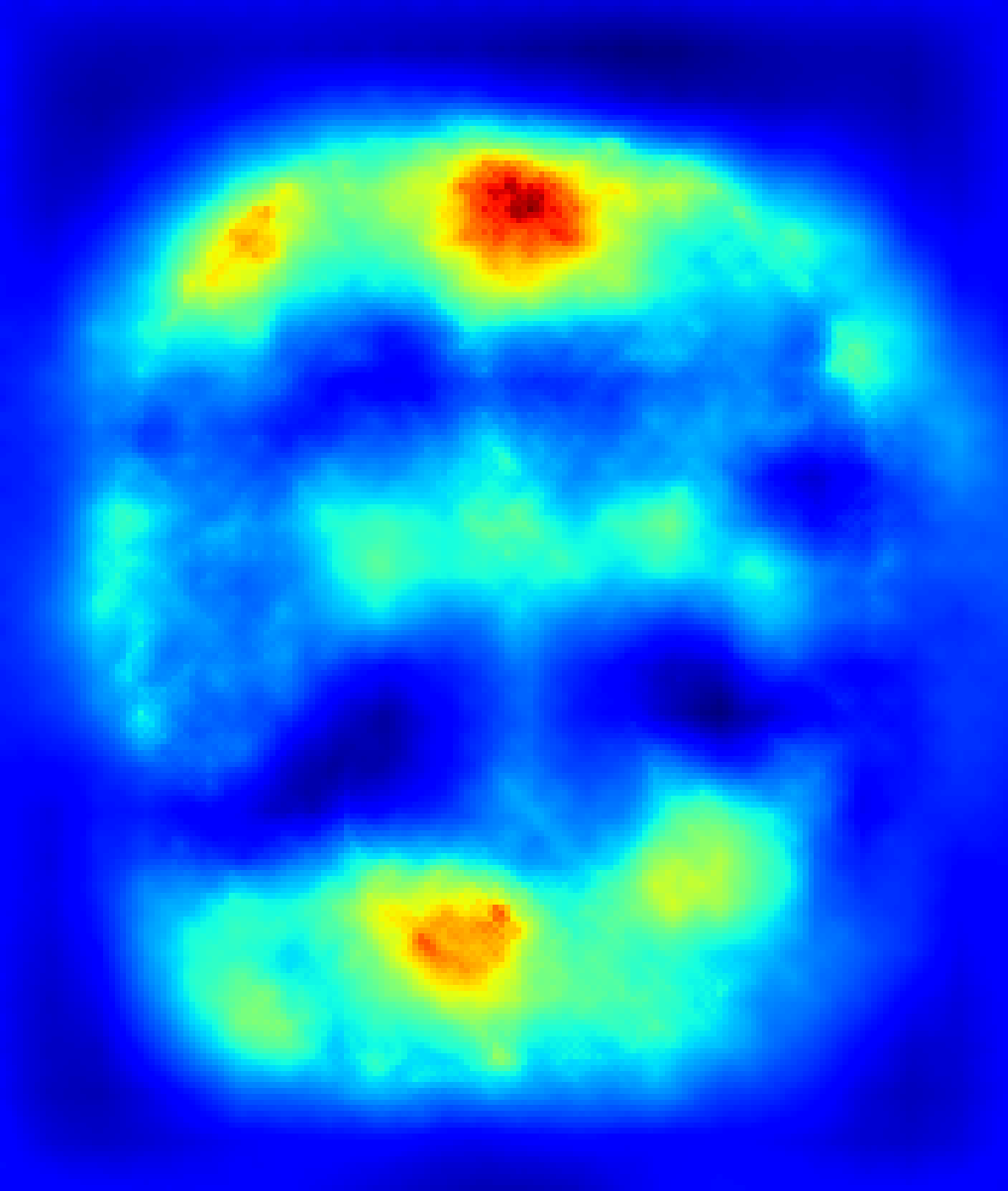} &
\includegraphics[width=\imgwidthc]{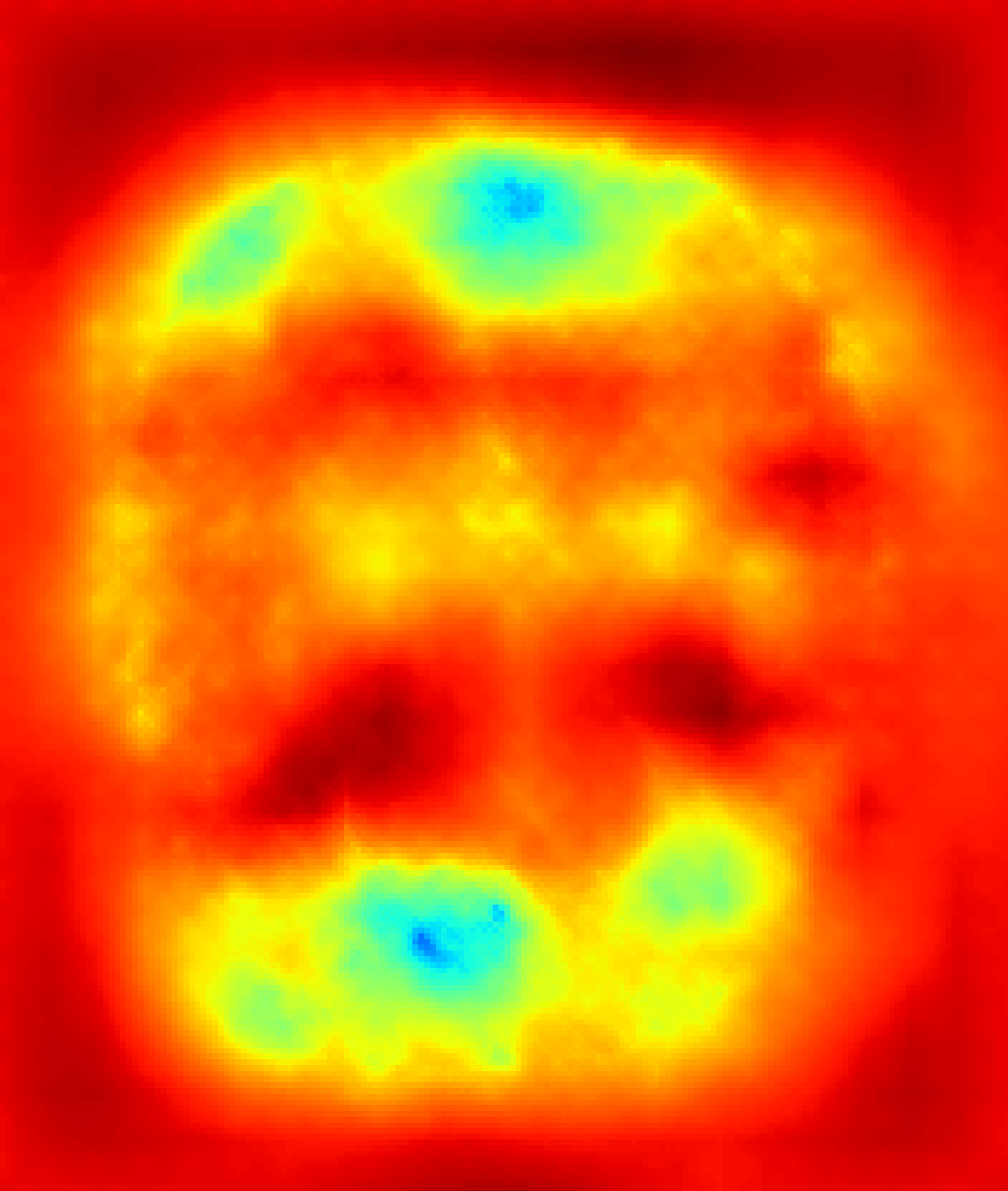} &
\includegraphics[width=\imgwidthc]{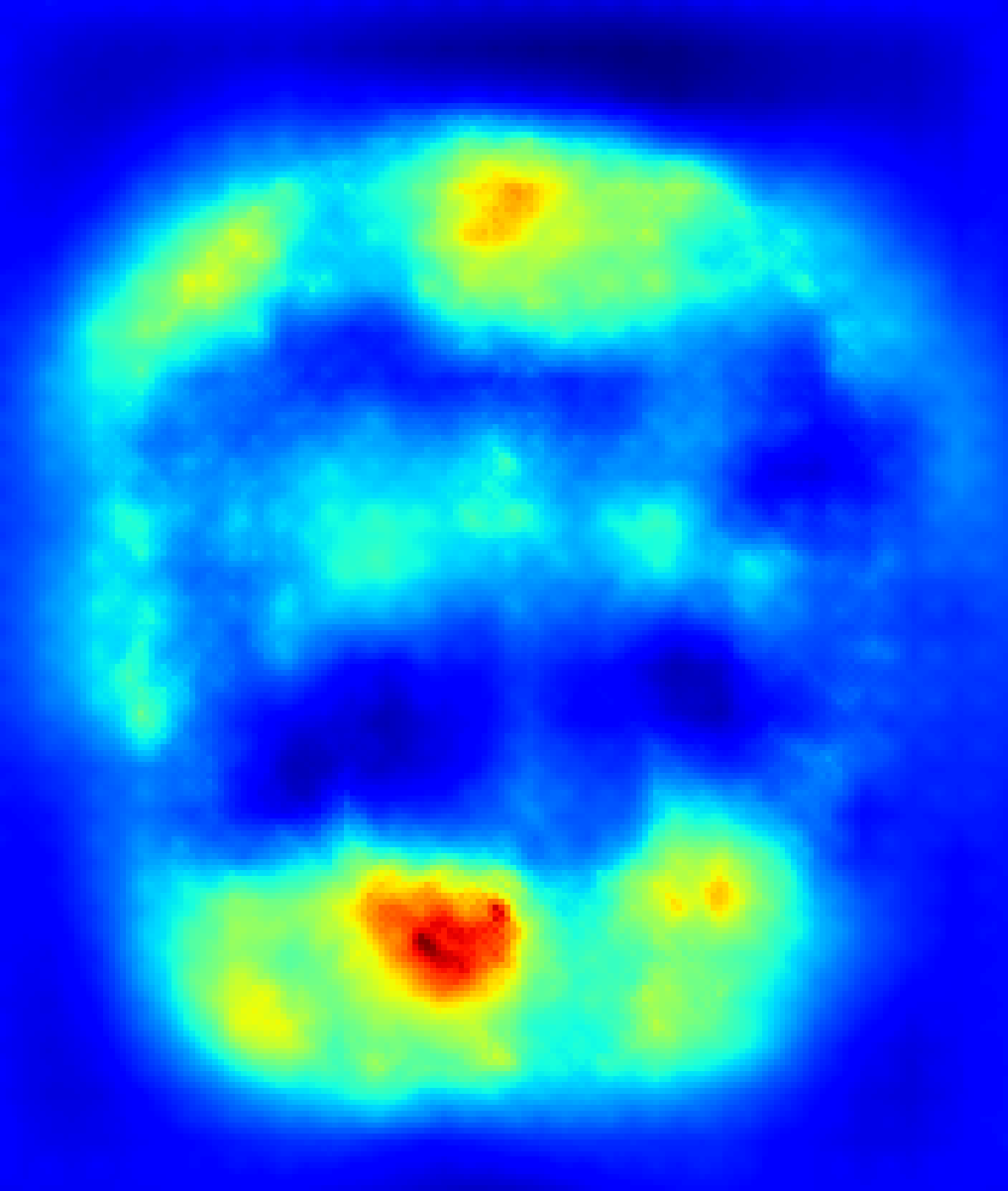} &
\includegraphics[width=\imgwidthc]{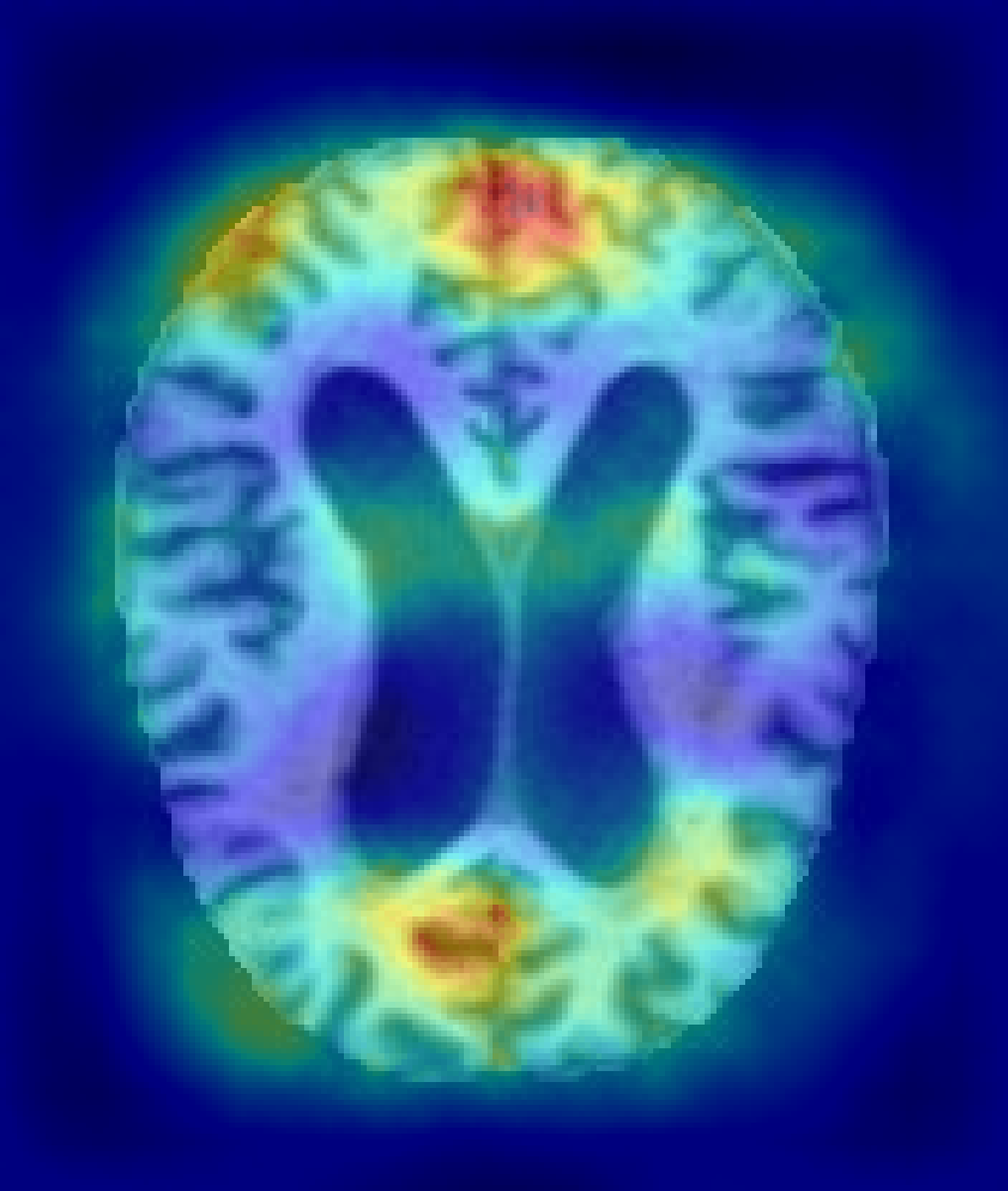} \\
{\tiny Original} & {\tiny Belief (\(b\): 0.49)} & {\tiny Vacuity (\(v\): 0.37)} & {\tiny Dissonance (\(d\): 0.71)} & {\tiny Belief (Overlay)} \\[0.5em]

\multicolumn{5}{@{}c@{}}{%
\setlength{\imgwidthc}{0.14\textwidth}
\begin{tabular}{@{}cccc@{}}
\includegraphics[width=\imgwidthc]{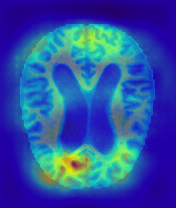} &
\includegraphics[width=\imgwidthc]{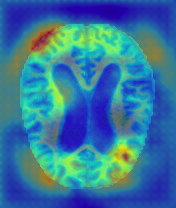} &
\includegraphics[width=\imgwidthc]{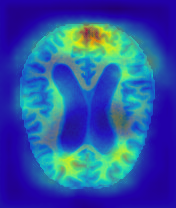} &
\includegraphics[width=\imgwidthc]{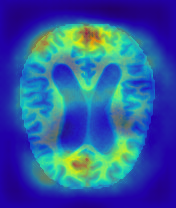} \\
{\tiny Mild (\(b\): 0.36)} & {\tiny Moderate (\(b\): 0.0)} & {\tiny Non-Demented (\(b\): 0.0)} & {\tiny Very Mild (\(b\): 0.49)}
\end{tabular}
}
\end{tabular}
\caption{Clinical uncertainty quantification in Alzheimer's MRI classification. A moderate belief (0.49) with high dissonance (0.71) and vacuity (0.37) indicates conflicting and insufficient evidence, and belief patterns correlate with diagnostic difficulty across disease severity levels.}
\label{fig:medical_uncertainty}
\end{figure}

Fig.~\ref{fig:medical_uncertainty} demonstrates Alzheimer's MRI classification, where moderate belief (0.49) combines with high dissonance (0.71) and substantial vacuity (0.37), indicating both conflicting and insufficient evidence. The dissonance heatmap reveals specific brain regions exhibiting overlapping signals across disease severity stages, while the vacuity map highlights areas where the model lacks sufficient evidence for confident classification. Critically, belief values correlate with diagnostic difficulty: near-zero for clinically distinct cases like moderate dementia (0.0) and non-demented (0.0), versus elevated for ambiguous early-stage cases like mild (0.36) and very mild (0.49) dementia. This pattern aligns with clinical reality, where early-stage Alzheimer's presents subtle, overlapping symptoms that challenge even expert radiologists. The framework thus identifies samples requiring additional clinical review or expert consultation, demonstrating its potential for safety-critical medical applications where understanding model uncertainty is as important as the prediction itself. Additional examples across all datasets and attribution methods are presented in Fig.~\ref{fig:examples}.

\section{Evaluation}
\label{sec:evaluation}

To evaluate the proposed UAM framework, we compare uncertainty maps against established XAI methods. Since no existing methods produce comparable uncertainty-based spatial attributions that distinguish between lack of evidence and conflicting evidence, we use gradient-based XAI methods as validation proxies. We employ two complementary methods: Integrated Gradients (IG)~\cite{sundararajan2017axiomatic} and SHAP GradientExplainer~\cite{lundberg2017unified}, which produce attribution values that can be positive (supporting), negative (opposing), or near-zero (no evidence).

All models were trained with pseudo-OOD sample augmentation~\cite{hendrycks2019using, lee2018simple} to ensure robust uncertainty estimation. For each training batch, we generated pseudo-OOD samples using Gaussian noise, pixel shuffling, intensity inversion, and random rotations~\cite{fort2021exploring}, assigned uniform label distributions (maximum uncertainty), and set an OOD-to-in-distribution ratio of 0.3 and a loss weight of 0.5.

For validation, we hypothesize that positive XAI attribution magnitude should negatively correlate with vacuity: regions with strong supporting evidence (high positive attributions) should show low vacuity, while regions lacking evidence (near-zero attributions) should show high vacuity. We validate through systematic perturbation analysis using the deletion metric~\cite{petsiuk2018rise}, OOD samples via geometric transformations, and Most Relevant First (MoRF)/Least Relevant First (LeRF) degradation tests~\cite{samek2016evaluating}.

\subsection{Validating Belief, Vacuity, and Dissonance Activation Maps}

\subsubsection{Testing by Generating OOD Samples}

\begin{figure}[H]
  \centering
  \includegraphics[width=.6\linewidth]{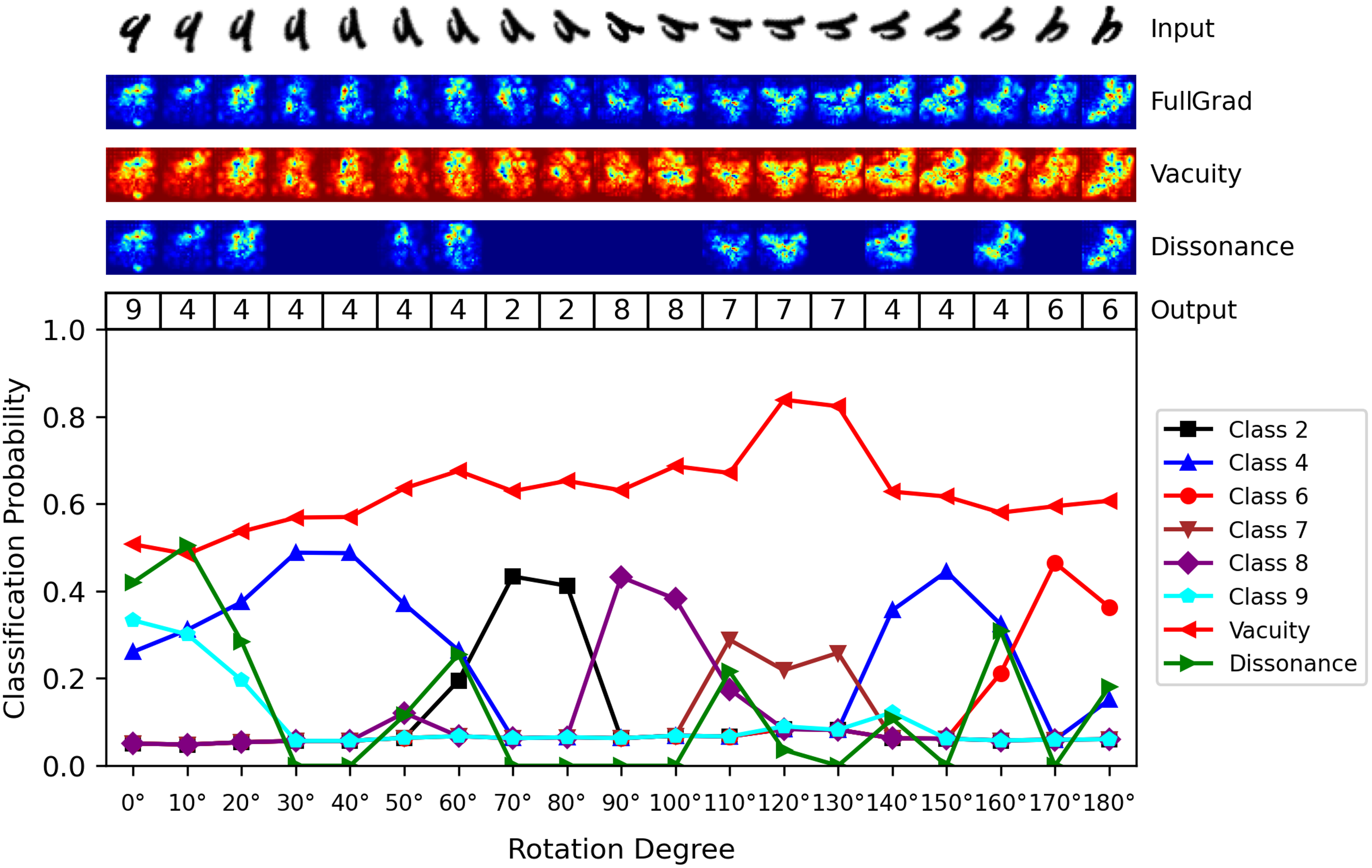}
  \caption{Classification probability and uncertainty visualization for MNIST digit '9' under 0° to 180° rotation.}
\label{fig:rotation}
\end{figure}

We progressively rotate a MNIST digit `9' from 0° to 180° to examine uncertainty evolution under OOD conditions (Fig.~\ref{fig:rotation}). For each rotation angle, we generate belief, vacuity, and dissonance activation maps to validate our framework on samples increasingly distant from the training distribution. The model initially predicts correctly despite elevated uncertainty, but increasingly misclassifies as rotation exceeds the training distribution. Vacuity remains high throughout, indicating significant overall uncertainty for OOD samples, while dissonance peaks at transition points (0°-20° and later stages), revealing conflicting evidence between digit classes when visual features become ambiguous. The activation maps reveal how the model's attention shifts with rotation, while the uncertainty measures provide complementary insights into the nature of classification failures. Notably, the dissonance activation maps show concentrated regions of conflict that shift spatially as the digit rotates, corresponding to features that could belong to multiple digit classes. This analysis validates that our framework successfully generates vacuity and dissonance activation maps for OOD samples, demonstrating how uncertainty visualization can identify when and why models become unreliable under distribution shift.

\subsubsection{Pixel Perturbation Test}

For each input, we obtain baseline EDL outputs (\(b_{\text{belief}}\), \(v_{\text{vacuity}}\), \(d_{\text{dissonance}}\)) and generate three activation map types. Each map guides a separate deletion analysis with pixels ranked by their scores. Following standard practices~\cite{petsiuk2018rise, chattopadhay2018grad}, we use a grayscale background (0.0) for MNIST and the dataset mean for color datasets. We expect that removing high-belief regions decreases true class belief, removing high-vacuity regions reduces uncertainty, and removing high-dissonance regions resolves conflicting evidence.

We employ perturbation processes on MoRF and LeRF regions~\cite{samek2016evaluating} to quantify attribution quality. MoRF removes the most important pixels in descending order, while LeRF removes pixels in ascending order. An effective attribution method should show rapid degradation in MoRF and gradual degradation in LeRF, quantified using AOPC and AUPC metrics. For belief and dissonance, we expect positive \(\Delta\)AOPC (MoRF \(>\) LeRF) and negative \(\Delta\)AUPC, as removing important pixels should decrease belief. For vacuity, we expect negative \(\Delta\)AOPC (MoRF \(<\) LeRF) and positive \(\Delta\)AUPC, as removing important pixels should increase belief by removing uncertainty regions. We analyze up to 200 correctly predicted samples per class (selected by highest belief), reporting mean trajectories with SEM.

\begin{table*}[htbp]
\centering
\caption{Evaluation Using MoRF and LeRF Pixel Perturbation}
\label{tab:enn_complete}
\resizebox{\textwidth}{!}{%
\begin{tabular}{ll|rrrr|rrrr|rrrr}
\toprule
& & \multicolumn{4}{c|}{\textbf{Belief}} & \multicolumn{4}{c|}{\textbf{Vacuity}} & \multicolumn{4}{c}{\textbf{Dissonance}} \\
\cmidrule(lr){3-6} \cmidrule(lr){7-10} \cmidrule(lr){11-14}
\textbf{Dataset} & \textbf{Method} & \textbf{MoRF} & \textbf{LeRF} & \textbf{$\Delta$AOPC} & \textbf{$\Delta$AUPC} & \textbf{MoRF} & \textbf{LeRF} & \textbf{$\Delta$AOPC} & \textbf{$\Delta$AUPC} & \textbf{MoRF} & \textbf{LeRF} & \textbf{$\Delta$AOPC} & \textbf{$\Delta$AUPC} \\
\midrule
\multirow{3}{*}{MNIST} 
& FG & 0.84±0.08 & 0.05±0.03 & \textbf{+0.79} & \textbf{-0.80} & 0.13±0.02 & 0.54±0.09 & -0.42 & +0.42 & 0.11±0.08 & 0.00±0.08 & +0.11 & -0.12 \\
& IG & 0.81±0.08 & 0.06±0.02 & +0.75 & -0.83 & 0.02±0.18 & 0.55±0.08 & -0.53 & +0.58 & 0.11±0.09 & -0.06±0.26 & +0.18 & -0.20 \\
& SHAP & 0.79±0.08 & 0.06±0.01 & +0.73 & -0.81 & -0.03±0.15 & 0.54±0.08 & \textbf{-0.57} & \textbf{+0.63} & 0.12±0.10 & -0.12±0.26 & \textbf{+0.23} & \textbf{-0.26} \\
\midrule
\multirow{3}{*}{CIFAR10} 
& FG & 0.67±0.05 & 0.47±0.07 & \textbf{+0.20} & \textbf{-0.20} & 0.40±0.09 & 0.49±0.10 & -0.10 & +0.10 & 0.41±0.05 & 0.33±0.07 & +0.08 & -0.09 \\
& IG & 0.76±0.07 & 0.64±0.13 & +0.12 & -0.13 & 0.44±0.08 & 0.52±0.06 & -0.08 & +0.09 & 0.42±0.05 & 0.37±0.08 & +0.05 & -0.05 \\
& SHAP & 0.76±0.07 & 0.63±0.13 & +0.13 & -0.14 & 0.48±0.08 & 0.58±0.05 & \textbf{-0.10} & \textbf{+0.11} & 0.42±0.04 & 0.37±0.08 & \textbf{+0.05} & \textbf{-0.06} \\
\midrule
\multirow{3}{*}{SVHN} 
& FG & 0.71±0.04 & 0.44±0.03 & \textbf{+0.27} & \textbf{-0.28} & 0.29±0.10 & 0.36±0.12 & -0.07 & +0.07 & 0.40±0.05 & 0.28±0.05 & \textbf{+0.12} & \textbf{-0.12} \\
& IG & 0.76±0.01 & 0.53±0.05 & +0.22 & -0.24 & 0.28±0.06 & 0.42±0.04 & -0.15 & +0.16 & 0.40±0.05 & 0.34±0.04 & +0.06 & -0.06 \\
& SHAP & 0.78±0.01 & 0.53±0.04 & +0.25 & \textbf{-0.28} & 0.27±0.06 & 0.43±0.04 & \textbf{-0.16} & \textbf{+0.17} & 0.41±0.05 & 0.34±0.04 & +0.07 & -0.07 \\
\midrule
\multirow{3}{*}{Imagenette} 
& FG & 0.68±0.08 & 0.16±0.06 & \textbf{+0.53} & \textbf{-0.53} & 0.26±0.07 & 0.38±0.05 & -0.12 & +0.13 & 0.43±0.09 & 0.20±0.07 & \textbf{+0.23} & \textbf{-0.23} \\
& IG & 0.71±0.07 & 0.48±0.15 & +0.23 & -0.25 & 0.26±0.07 & 0.38±0.05 & -0.12 & +0.13 & 0.43±0.04 & 0.33±0.06 & +0.10 & -0.10 \\
& SHAP & 0.71±0.08 & 0.51±0.14 & +0.21 & -0.23 & 0.29±0.06 & 0.38±0.05 & \textbf{-0.09} & \textbf{+0.10} & 0.43±0.04 & 0.33±0.06 & +0.10 & -0.10 \\
\midrule
\multirow{3}{*}{AD} 
& FG & 0.58±0.43 & 0.44±0.34 & +0.13 & -0.13 & 0.48±0.41 & 0.52±0.44 & -0.04 & +0.04 & 0.38±0.46 & 0.36±0.45 & +0.02 & -0.02 \\
& IG & 0.85±0.01 & 0.19±0.19 & \textbf{+0.66} & \textbf{-0.73} & 0.08±0.11 & 0.70±0.04 & \textbf{-0.62} & \textbf{+0.68} & 0.57±0.04 & 0.19±0.29 & \textbf{+0.38} & \textbf{-0.42} \\
& SHAP & 0.82±0.02 & 0.22±0.05 & +0.60 & -0.66 & 0.13±0.05 & 0.67±0.04 & -0.54 & +0.59 & 0.20±0.04 & 0.26±0.05 & -0.06 & +0.07 \\
\bottomrule
\end{tabular}
}
\begin{tablenotes}
\footnotesize
\item FG = FullGrad; IG = Integrated Gradients. Bold values indicate best performance per dataset/uncertainty type.
\end{tablenotes}
\end{table*}

Table~\ref{tab:enn_complete} presents comprehensive evaluation results across the five datasets. MoRF and LeRF columns show the AOPC values with their standard deviations, while \(\Delta\)AOPC and \(\Delta\)AUPC represent the differences between MoRF and LeRF for each metric, computed as \(\Delta\text{AOPC} = \text{AOPC}_{\text{MoRF}} - \text{AOPC}_{\text{LeRF}}\) and \(\Delta\text{AUPC} = \text{AUPC}_{\text{MoRF}} - \text{AUPC}_{\text{LeRF}}\). Bold values indicate the best performance for each dataset within each uncertainty type: for belief and dissonance, this corresponds to the highest \(\Delta\)AOPC (most positive) and lowest \(\Delta\)AUPC (most negative), while for vacuity, it corresponds to the lowest \(\Delta\)AOPC (most negative) and highest \(\Delta\)AUPC (most positive), reflecting the inverse relationship where effective attribution of uncertainty-driving regions causes rapid increases in vacuity upon removal.

For belief activation maps, all three attribution methods (FullGrad, IG, SHAP) effectively identify important regions with positive \(\Delta\)AOPC and negative \(\Delta\)AUPC values. MNIST shows the strongest performance, with \(\Delta\)AOPC exceeding +0.73 across all methods, and FullGrad achieving the best at +0.79. For CIFAR10 and SVHN, \(\Delta\)AOPC ranges from +0.12 to +0.20 and +0.22 to +0.27 respectively, while FullGrad achieves the highest separation on Imagenette at +0.53. For the AD dataset, IG demonstrates strong performance (\(\Delta\)AOPC = +0.66), while FullGrad exhibits high variance (±0.43) and lower performance (\(\Delta\)AOPC = +0.13) due to limited analyzable samples from the ``Very Mild Demented'' class. FullGrad consistently achieves the best performance across MNIST, CIFAR10, SVHN, and Imagenette for belief attribution.

For vacuity activation maps, negative \(\Delta\)AOPC and positive \(\Delta\)AUPC values indicate that removing low-vacuity (confident) regions first (LeRF) causes greater increases in total uncertainty than removing high-vacuity (uncertain) regions first (MoRF). This confirms effective identification of uncertainty-driving regions. On MNIST, SHAP achieves the strongest performance with \(\Delta\)AOPC = -0.57 and \(\Delta\)AUPC = +0.63, demonstrating clear separation between confident and uncertain regions. Performance differences become smaller on more complex datasets: CIFAR10 and SVHN show \(\Delta\)AOPC ranging from -0.07 to -0.16, while Imagenette shows \(\Delta\)AOPC from -0.09 to -0.12. On the AD dataset, IG achieves the strongest separation (\(\Delta\)AOPC = -0.62, \(\Delta\)AUPC = +0.68), while FullGrad shows minimal performance (\(\Delta\)AOPC = -0.04). This performance difference stems from FullGrad's activation-weighting mechanism, which suppresses regions with low neural responses that correspond to high vacuity. Since vacuity represents a lack of evidence and corresponds to weak activations, IG and SHAP's direct input-space gradient computations better capture the uncertainty-driving regions, particularly in the medical imaging dataset. Overall, the relative performance varies by dataset complexity and domain characteristics.

For dissonance activation maps, positive \(\Delta\)AOPC values indicate effective attribution in datasets with dissonance, though the magnitudes are generally smaller than those for belief and vacuity. FullGrad performs best on SVHN (+0.12) and Imagenette (+0.23), while SHAP achieves the highest \(\Delta\)AOPC on MNIST (+0.23). CIFAR10 shows the smallest differences (\(\Delta\)AOPC from +0.05 to +0.08). On the AD dataset, IG achieves the best performance (\(\Delta\)AOPC = +0.38), while SHAP shows negative separation (\(\Delta\)AOPC = -0.06), indicating limited effectiveness in distinguishing dissonance-driving regions. FullGrad demonstrates minimal performance (\(\Delta\)AOPC = +0.02) with high variance (±0.46), likely due to its sensitivity to weak activations in conflicting evidence scenarios. However, it is important to note that dissonance evaluation is limited by small sample sizes: only 5 samples exhibited positive dissonance on MNIST (distributed across classes 0, 3, 5, and 9), and only 33 samples across all classes on AD (7 from Non Demented, 0 from Very Mild Demented, 5 from Mild Demented, and 21 from Moderate Demented), making the dissonance evaluation on these datasets less comprehensive (see the supplementary material for details).

\subsection{Aggregated Attribution Analysis}

\begin{figure}[!htp]
  \centering
  \setlength{\tabcolsep}{2pt}
  \renewcommand{\arraystretch}{1.2}
  \scriptsize
  \begin{tabular}{cccc}
    & Belief & Vacuity & Dissonance \\
    \raisebox{0.5cm}{\rotatebox{90}{FullGrad}} &
      \includegraphics[width=0.31\linewidth]{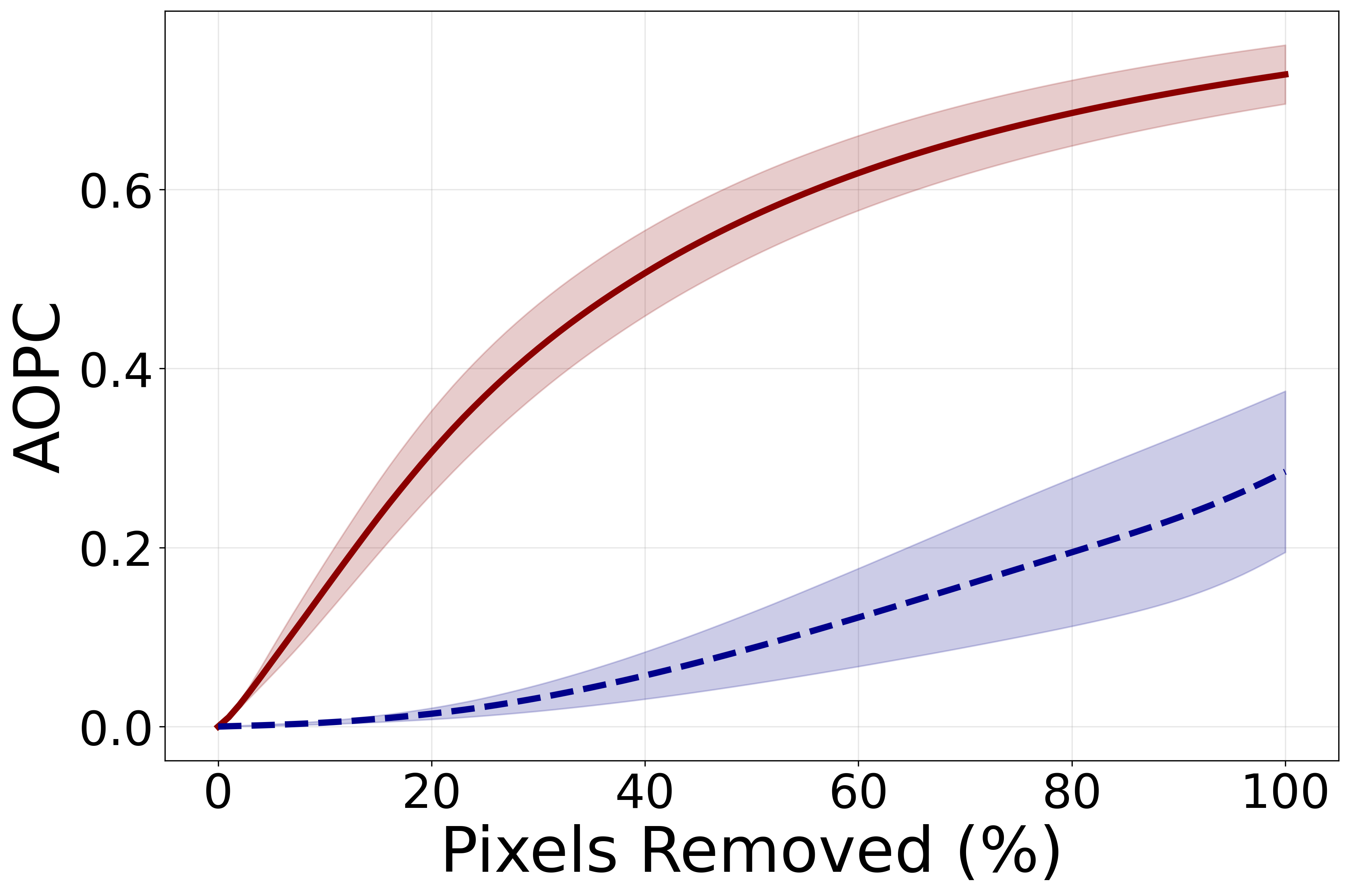} &
      \includegraphics[width=0.31\linewidth]{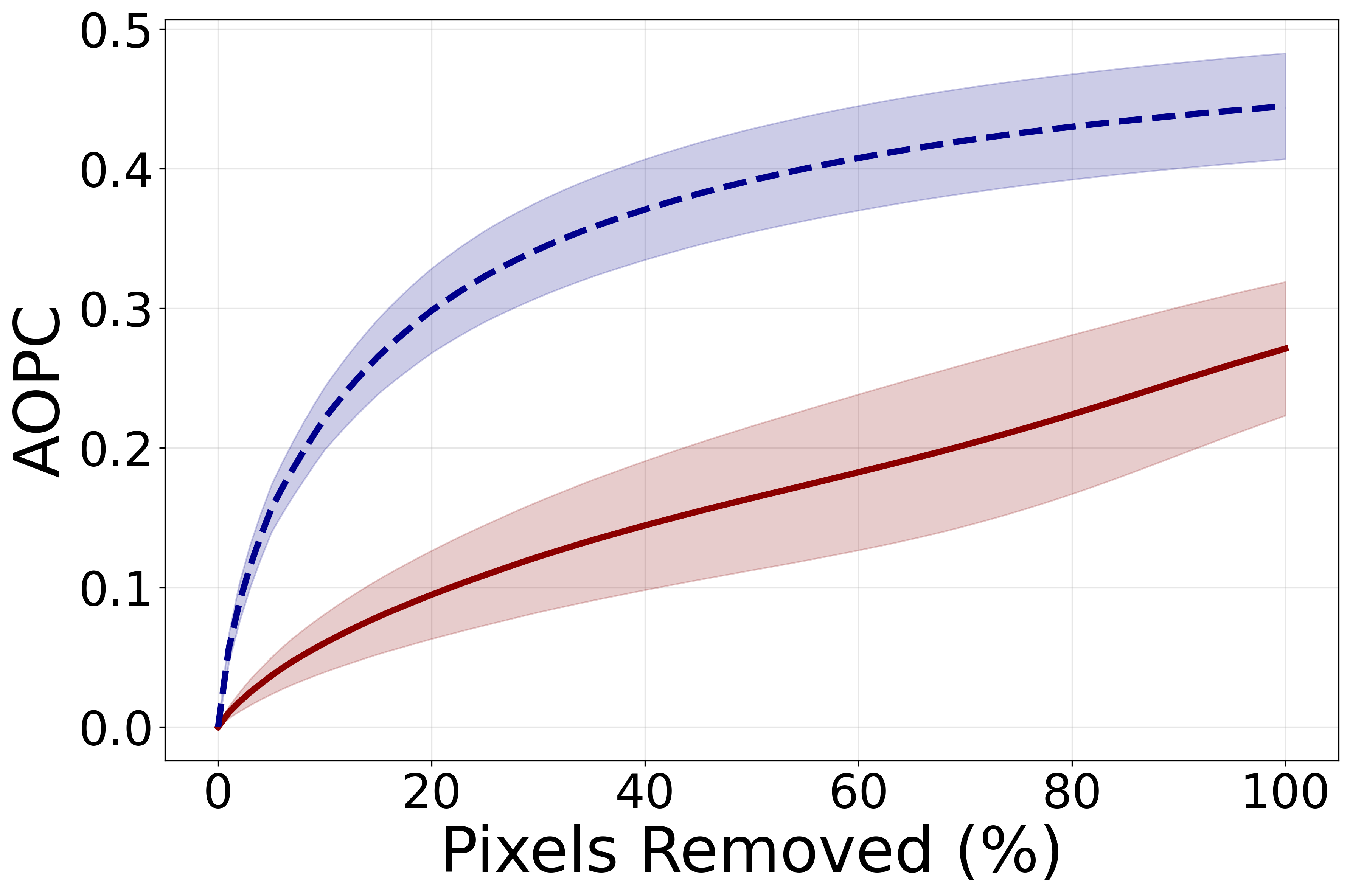} &
      \includegraphics[width=0.31\linewidth]{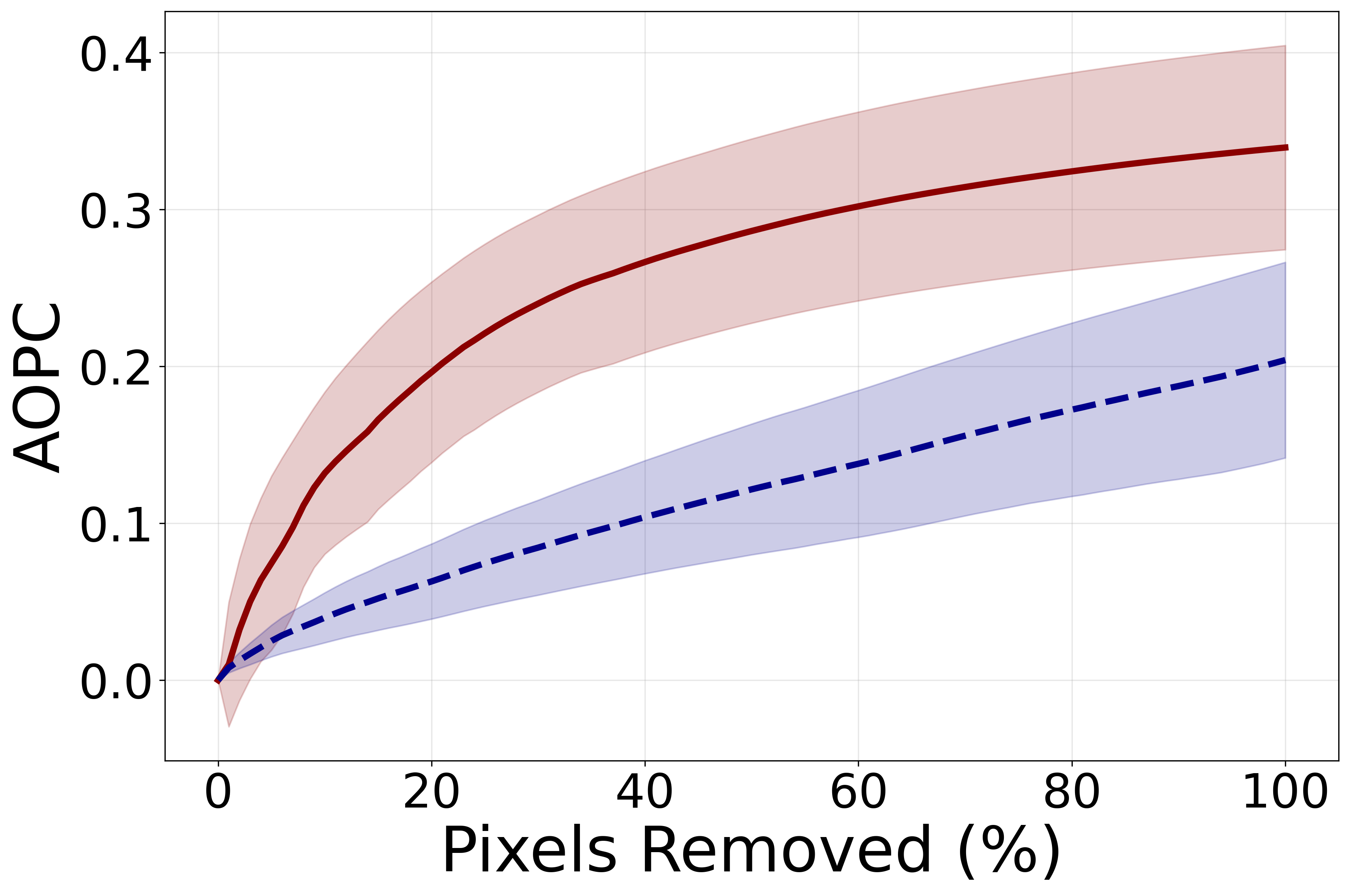} \\
    \raisebox{0.7cm}{\rotatebox{90}{IG}} &
      \includegraphics[width=0.31\linewidth]{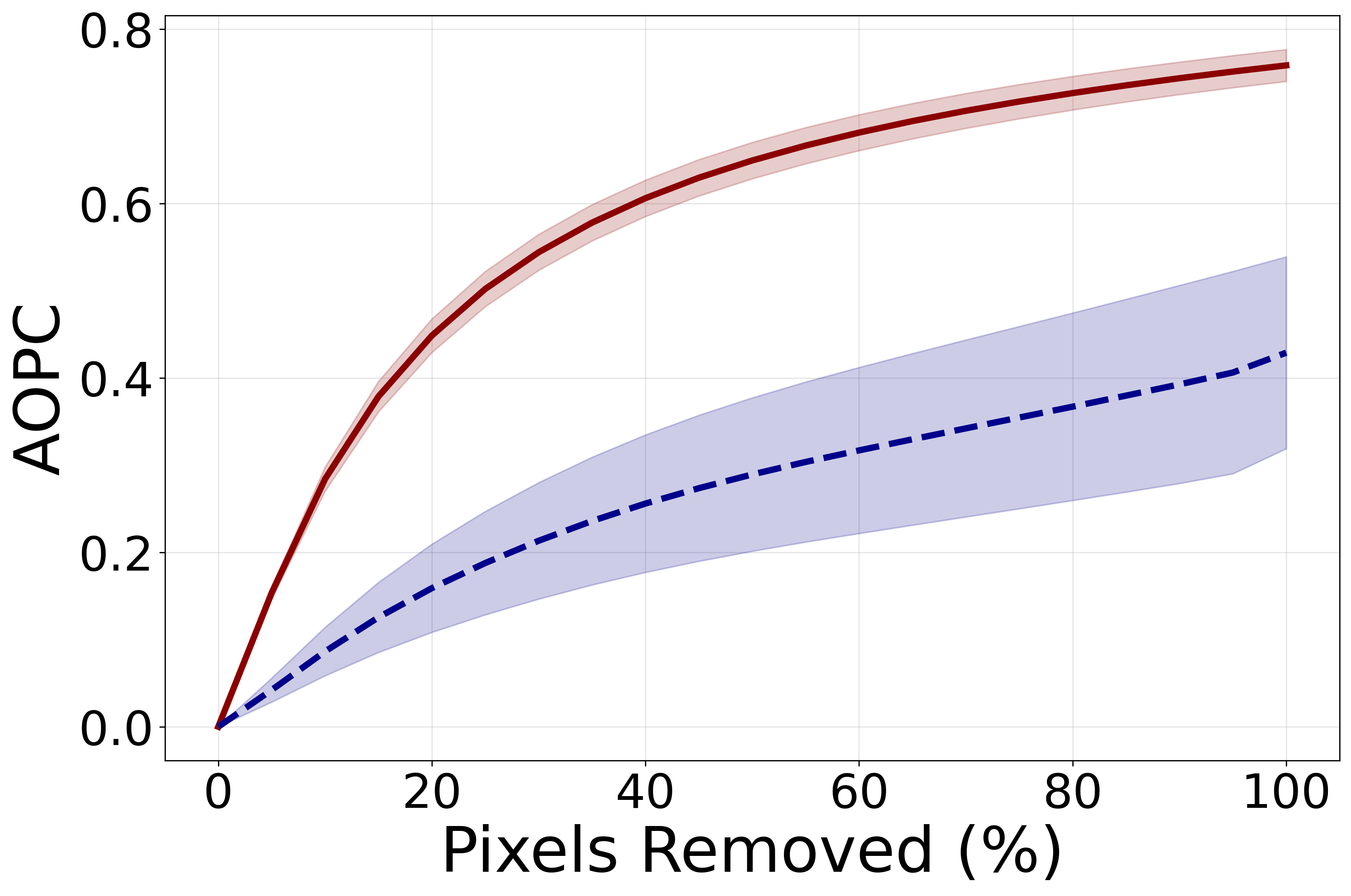} &
      \includegraphics[width=0.31\linewidth]{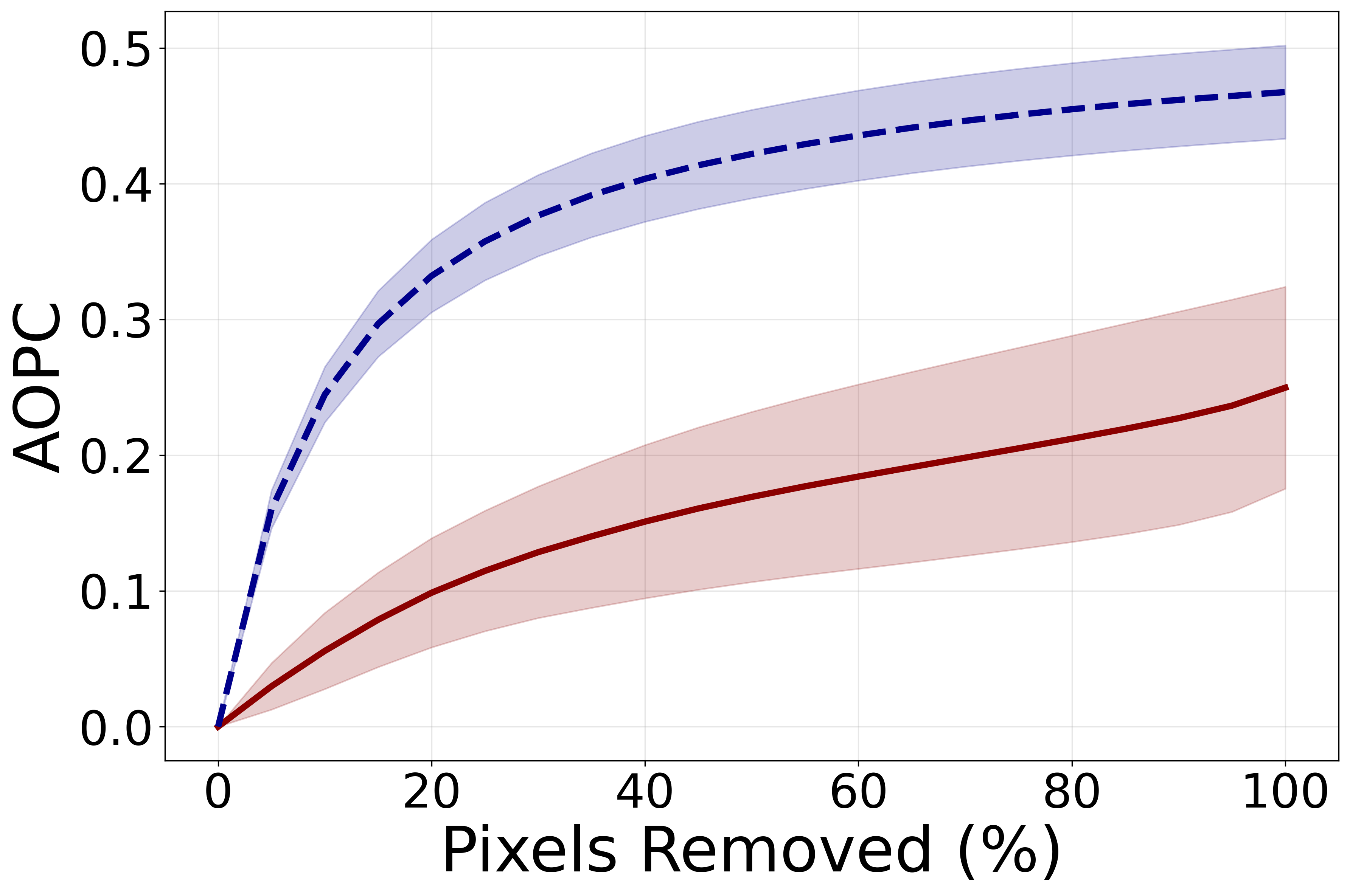} &
      \includegraphics[width=0.31\linewidth]{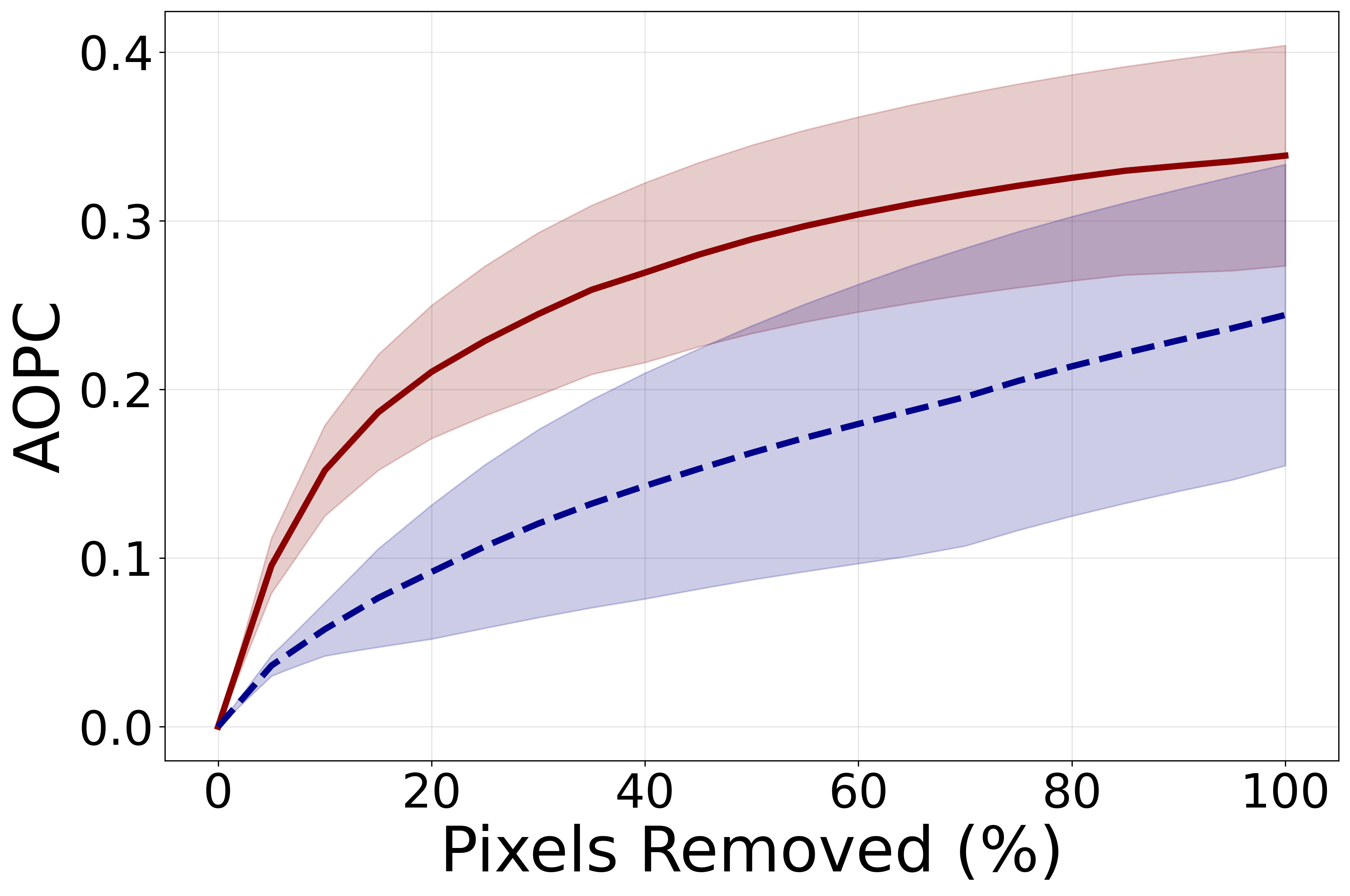} \\
    \raisebox{0.6cm}{\rotatebox{90}{SHAP}} &
      \includegraphics[width=0.31\linewidth]{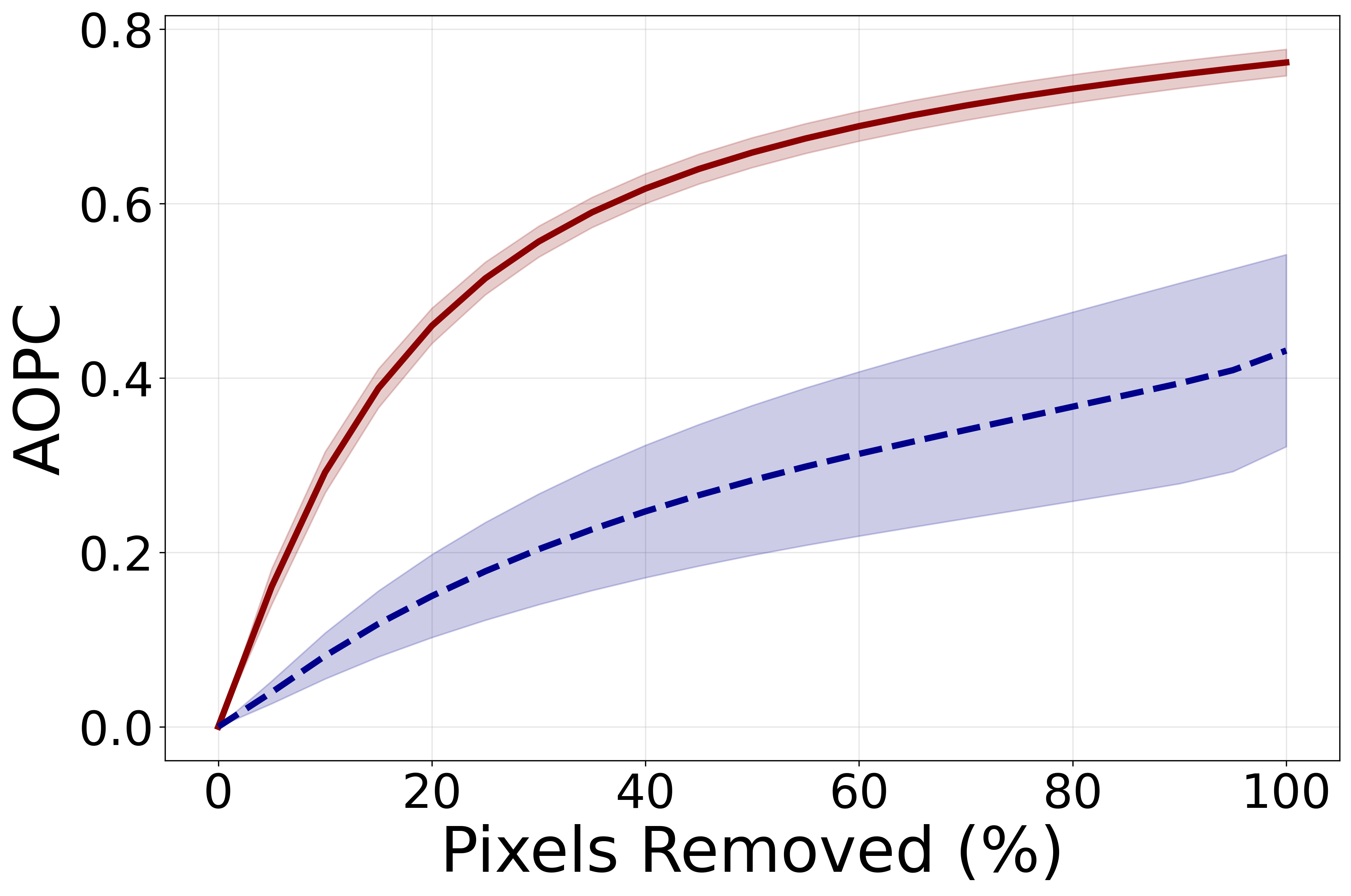} &
      \includegraphics[width=0.31\linewidth]{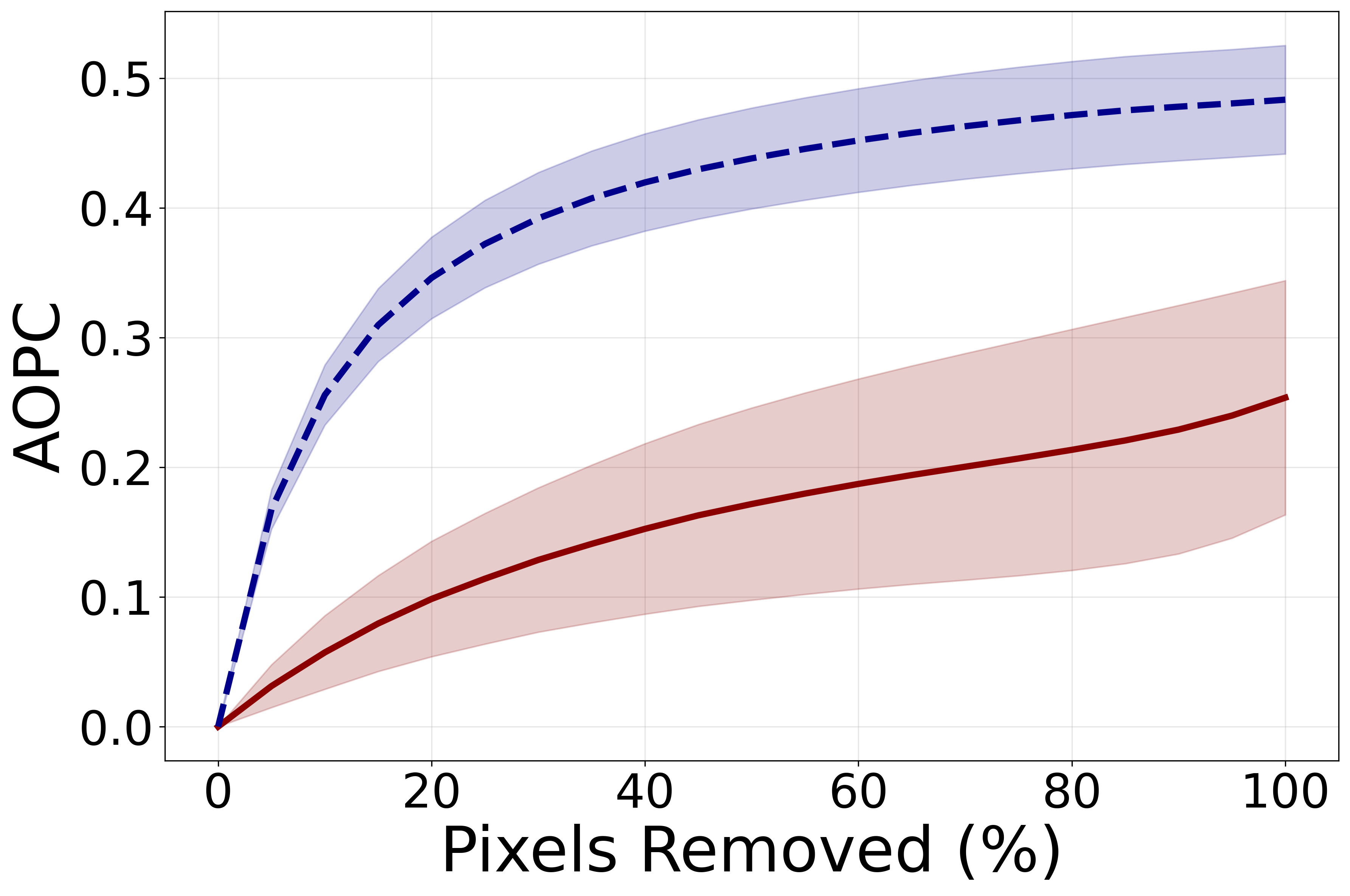} &
      \includegraphics[width=0.31\linewidth]{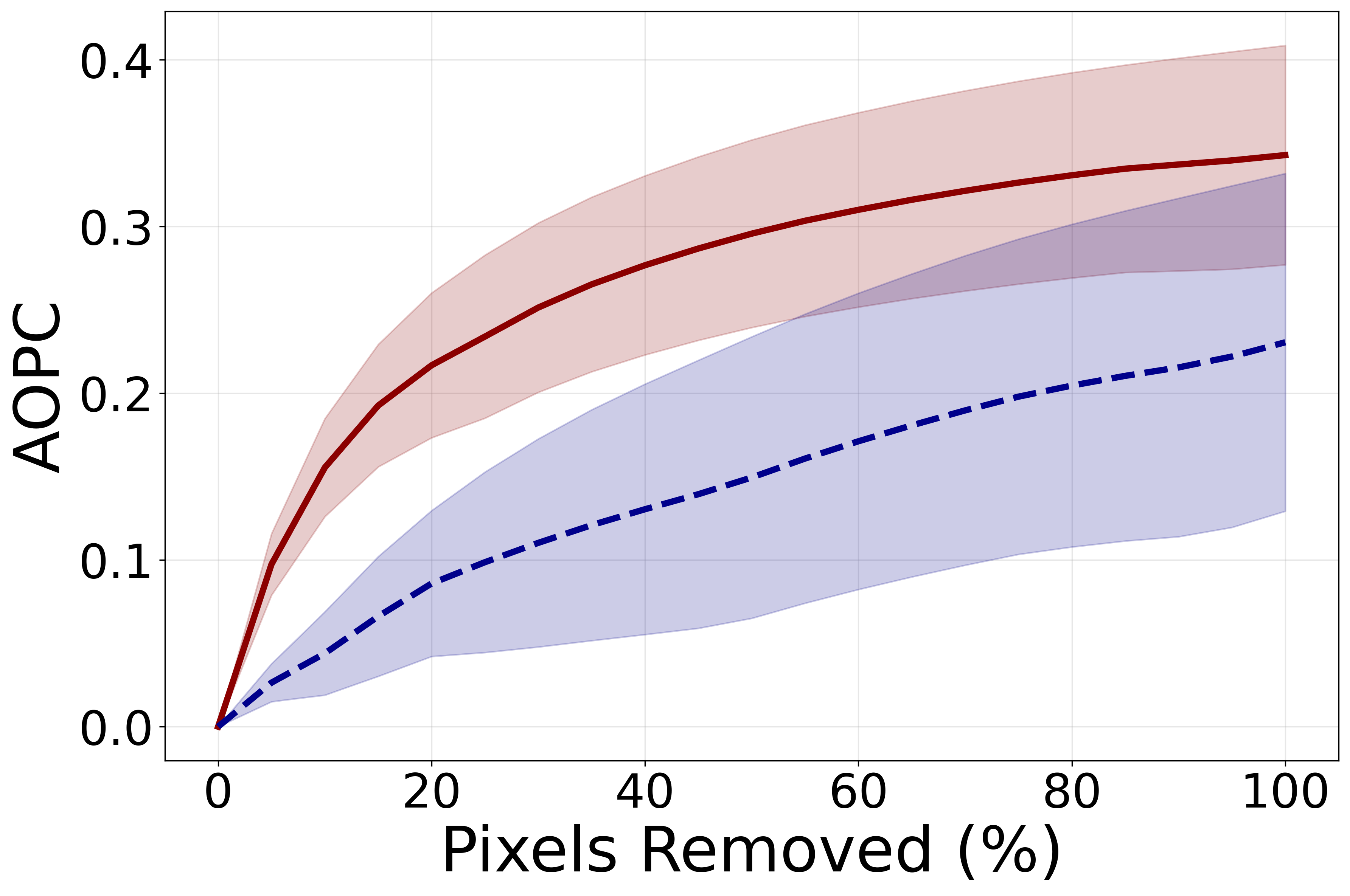} \\
  \end{tabular}
  \caption{AOPC evaluation across attribution methods. Red solid: MoRF; blue dashed: LeRF. Shaded regions show standard error of the mean across the datasets.}
  \label{fig:aggregated_attribution_comparison}
\end{figure}

Figure~\ref{fig:aggregated_attribution_comparison} presents aggregated AOPC results across all five datasets (MNIST, CIFAR10, SVHN, Imagenette, and AD). Each row corresponds to a different attribution method, and each column represents a different activation metric. The solid red lines indicate MoRF (removing high-importance pixels), while dashed blue lines show LeRF (removing low-importance pixels). Shaded regions show the standard error of the mean across datasets. For belief (left column), all methods show strong performance with clear MoRF-LeRF separation. FullGrad achieves the highest \(\Delta\)AOPC of \(0.44 \pm 0.12\), followed by SHAP (\(0.33 \pm 0.12\)) and IG (\(0.33 \pm 0.12\)), confirming effective identification of prediction-driving regions. For vacuity (middle column), negative \(\Delta\)AOPC values indicate inverse behavior: FullGrad shows \(-0.17 \pm 0.07\), IG shows \(-0.22 \pm 0.09\), and SHAP shows \(-0.23 \pm 0.10\). The consistent LeRF outperformance over MoRF confirms that vacuity attributions capture uncertainty regions where the model lacks evidence. For dissonance (right column), moderate positive \(\Delta\)AOPC values indicate identification of conflicting evidence regions: FullGrad achieves \(0.14 \pm 0.03\), SHAP achieves \(0.11 \pm 0.04\), and IG achieves \(0.09 \pm 0.03\). Together, these metrics provide complementary insights: belief highlights confident predictions, vacuity reveals insufficient evidence, and dissonance captures conflicting information, offering a comprehensive uncertainty-aware attribution framework.

\section{Discussion}
\label{sec:Discussion}

While our proposed uncertainty visualization approach effectively highlights spatial regions that contribute to model uncertainty, it operates under a global-to-local attribution principle. We leverage the additivity property of belief mass and vacuity to generate spatially varying uncertainty maps. As shown in Algorithm~\ref{alg:vacuity_full_gradcam}, we compute the total belief at each spatial location \(B_{\mathrm{total}}(x,y)\) by accumulating normalized attribution maps weighted by their corresponding belief masses \(b_{i,j}\), then derive vacuity as \(u(x,y) = 1 - B_{\mathrm{total}}(x,y)\). This yields pixel-specific vacuity values, with different regions having different uncertainty magnitudes based on the weighted combination of class-specific attribution maps.

However, this spatially varying uncertainty still depends on the global belief distribution computed from the entire image. Each class's belief mass \(b_{i,j}\) is derived from the model's output for the complete input, and the FullGrad attribution maps distribute these fixed beliefs spatially according to gradient-based importance. This differs fundamentally from computing independent uncertainty values for isolated regions through occlusion or masking methods. For example, inferring only a dog's face versus only its body would yield fundamentally different belief distributions and thus different uncertainty measures, whereas our approach distributes the uncertainty from whole-image inference across spatial locations, specifically when using FullGrad.

This distinction becomes important in complex images with multiple semantic regions or objects, each potentially contributing different levels of evidence toward various class hypotheses. Consider an image containing both a dog's face and body: if evaluated independently, the facial features might provide strong evidence with a belief of 0.9 (vacuity of 0.1) while the partially occluded body could exhibit a belief of 0.3 (vacuity of 0.7) due to missing information. Our current gradient-based approach distributes global uncertainty spatially but cannot capture these fundamentally different evidence patterns that would emerge from independent regional evaluation. While computationally efficient and theoretically grounded in FullGrad, this limitation may affect scenarios that produce substantially different belief distributions if evaluated in isolation.

A more sophisticated approach would involve image segmentation methods to partition the input into semantically meaningful regions before computing belief distributions. By applying the UAM framework to individual segments rather than the entire image, we could obtain segment-specific Dirichlet parameters and compute vacuity and dissonance for each region independently. One potential approach would combine superpixel or attention-based segmentation with local uncertainty estimation: first segment the image into coherent regions (e.g., 50--100 superpixels), then compute evidential predictions for each segment by masking and inferring on isolated regions, and finally aggregate these segment-level uncertainties into a comprehensive visualization. Methods like SHAP could be applied to ensure theoretically principled attribution, where segment contributions satisfy local additivity properties.

However, this enhancement comes with significant computational trade-offs. Computing segment-specific beliefs requires multiple forward passes through the network (one per segment or masked region). For example, 50 superpixels would require 50 forward passes compared to our current single-pass approach. Additionally, defining optimal segmentation granularity, handling segment interactions (since occluding one region may affect the evidence for others), and aggregating segment-level uncertainties into coherent visualizations pose non-trivial methodological challenges. Nevertheless, for applications requiring highly detailed, spatially-resolved uncertainty attribution in complex visual scenes—such as medical imaging with multiple lesions or autonomous driving with numerous objects—these trade-offs may be justified, and we identify this as an important avenue for extending our framework's capabilities.

\begin{figure*}[tbp]
  \centering
  \setlength{\tabcolsep}{0.5pt}  
  \renewcommand{\arraystretch}{0.5}
  \scriptsize
  \begin{tabular}{cccccccccccccc}
    Original & Scores & \multicolumn{4}{c}{FullGrad} & \multicolumn{4}{c}{IG} & \multicolumn{4}{c}{SHAP} \\
    \cmidrule(lr){3-6} \cmidrule(lr){7-10} \cmidrule(lr){11-14}
    & & Belief & Vacuity & Dissonance & Overlay & Belief & Vacuity & Dissonance & Overlay & Belief & Vacuity & Dissonance & Overlay \\
    \hline

      \includegraphics[width=0.07\linewidth]{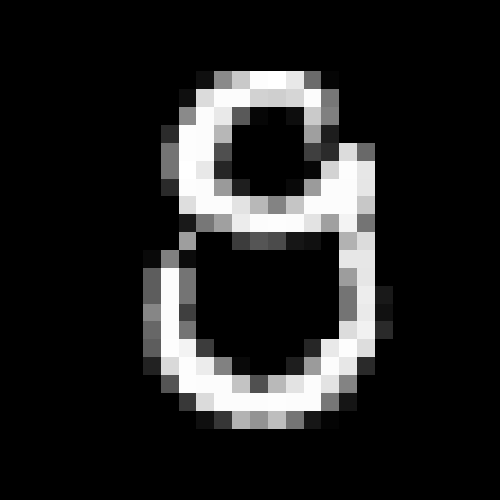} &
      \raisebox{0.4\height}{\shortstack{$b=0.76$ \\ $v=0.23$ \\ $d=0.00$}} &
      \includegraphics[width=0.07\linewidth]{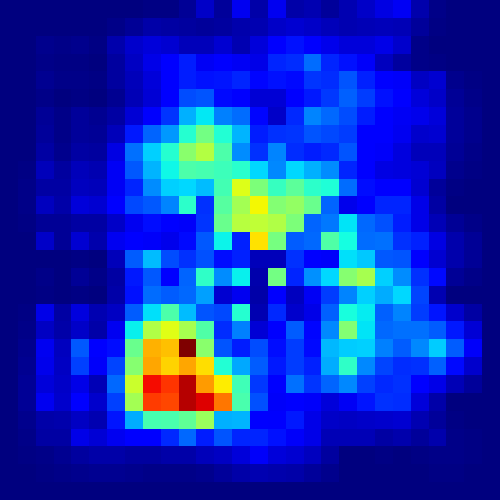} &
      \includegraphics[width=0.07\linewidth]{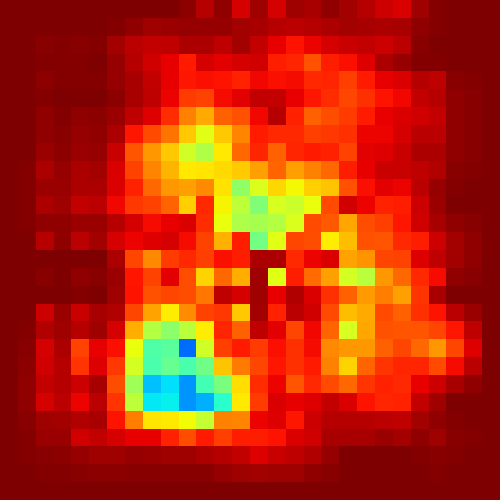} &
      \includegraphics[width=0.07\linewidth]{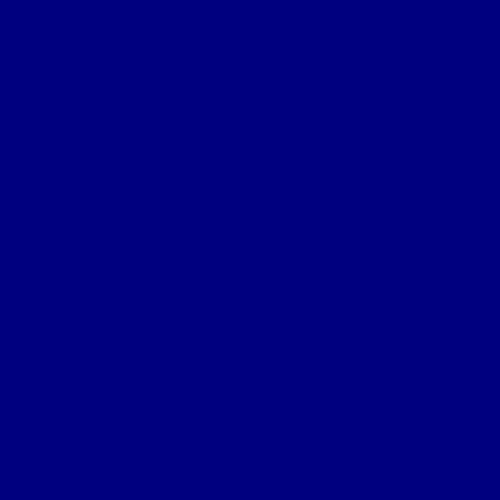} &
      \includegraphics[width=0.07\linewidth]{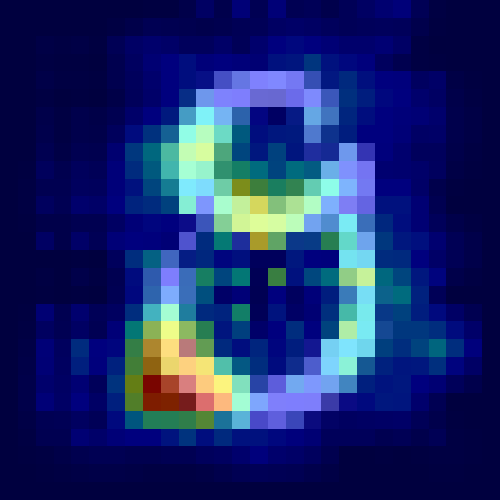} &
      \includegraphics[width=0.07\linewidth]{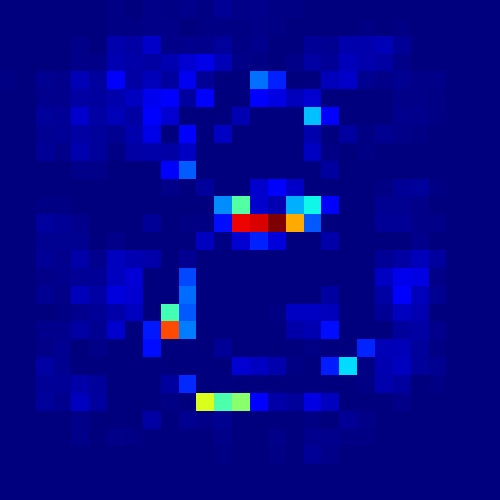} &
      \includegraphics[width=0.07\linewidth]{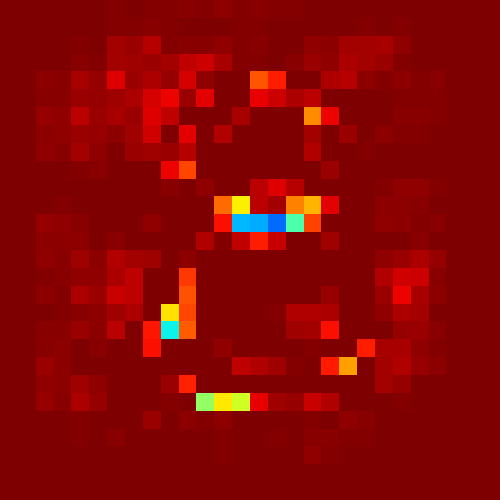} &
      \includegraphics[width=0.07\linewidth]{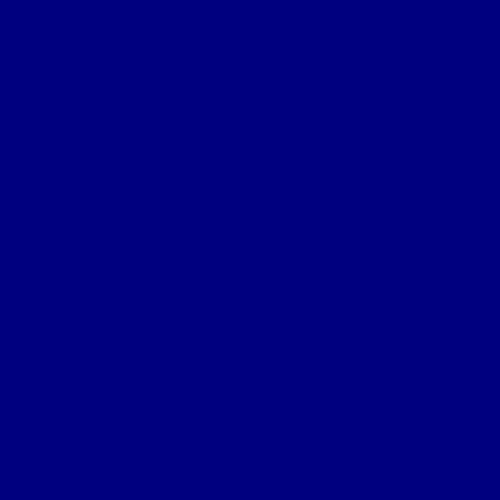} &
      \includegraphics[width=0.07\linewidth]{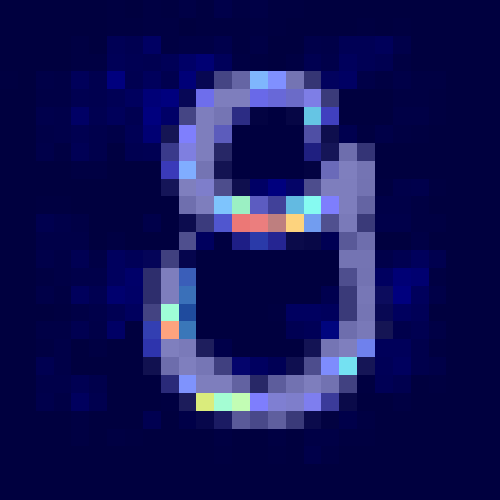} &
      \includegraphics[width=0.07\linewidth]{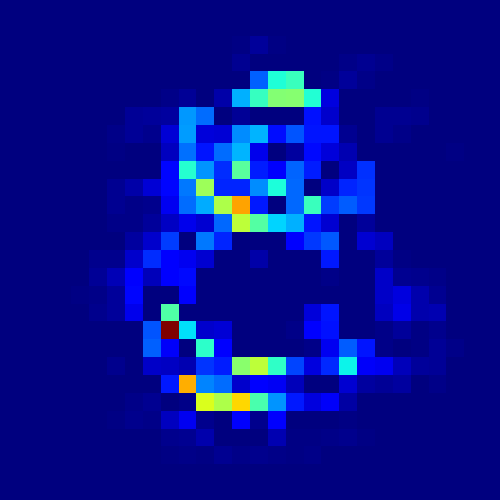} &
      \includegraphics[width=0.07\linewidth]{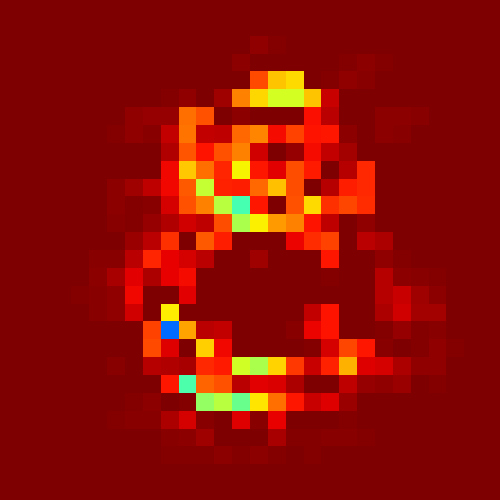} &
      \includegraphics[width=0.07\linewidth]{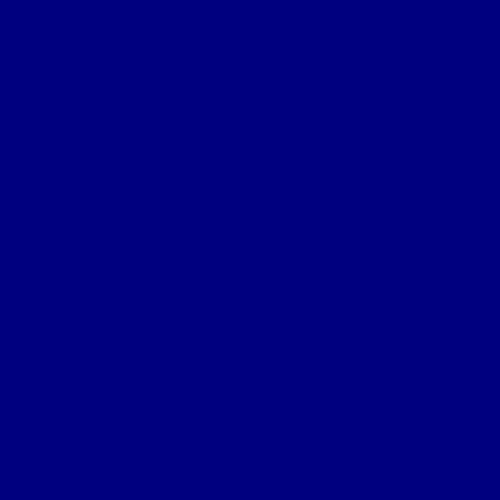} &
      \includegraphics[width=0.07\linewidth]{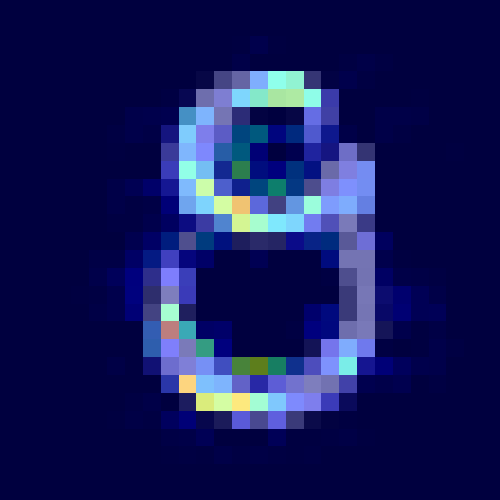} \\
  
      \includegraphics[width=0.07\linewidth]{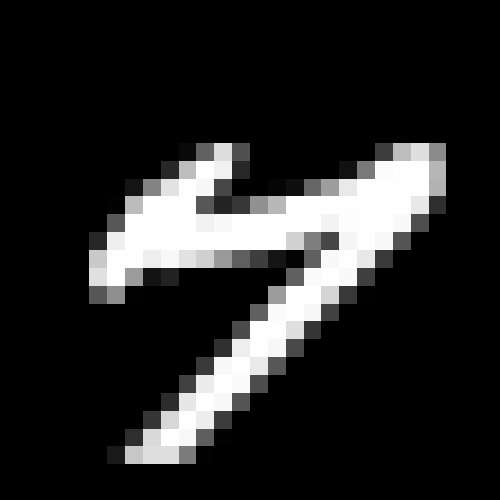} &
      \raisebox{0.4\height}{\shortstack{$b=0.57$ \\ $v=0.42$ \\ $d=0.00$}} &
      \includegraphics[width=0.07\linewidth]{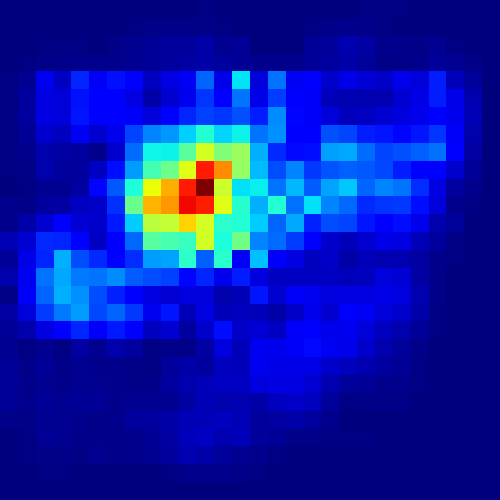} &
      \includegraphics[width=0.07\linewidth]{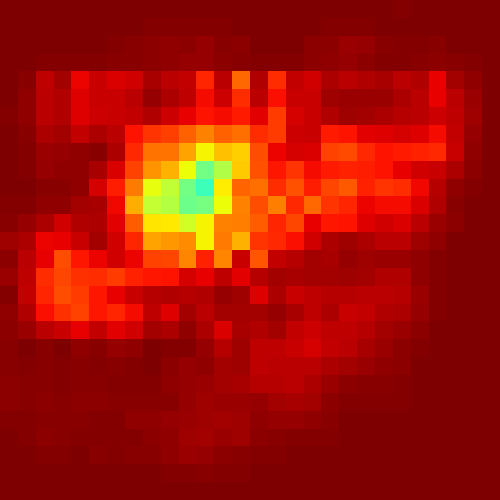} &
      \includegraphics[width=0.07\linewidth]{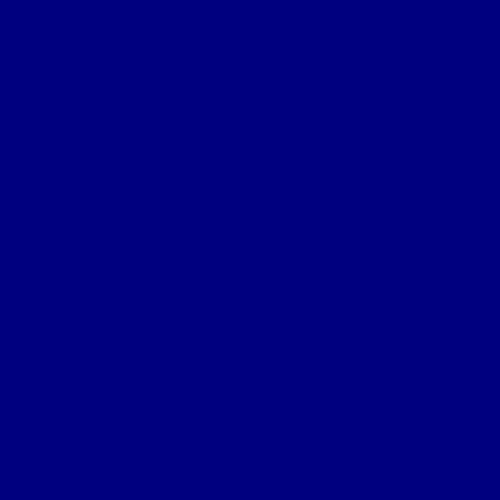} &
      \includegraphics[width=0.07\linewidth]{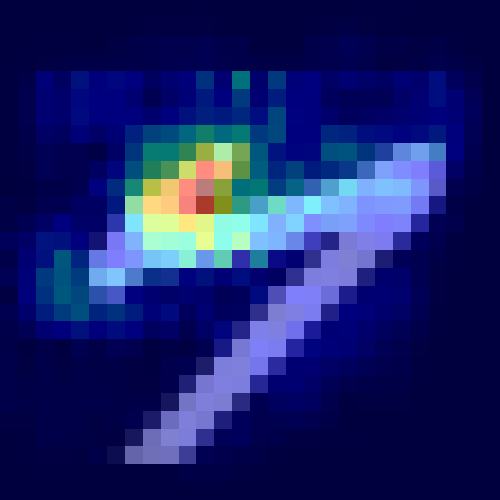} &
      \includegraphics[width=0.07\linewidth]{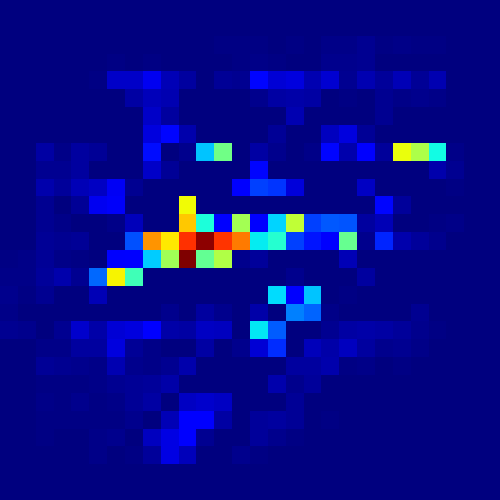} &
      \includegraphics[width=0.07\linewidth]{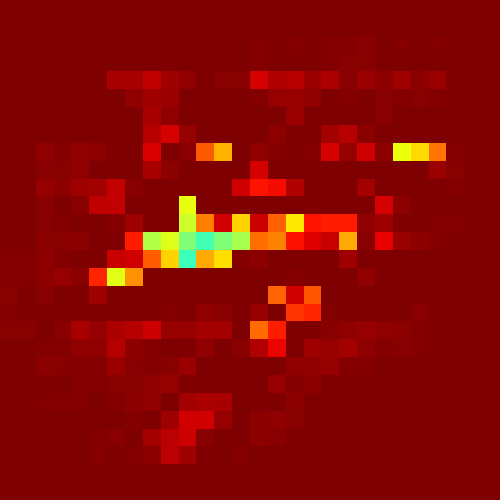} &
      \includegraphics[width=0.07\linewidth]{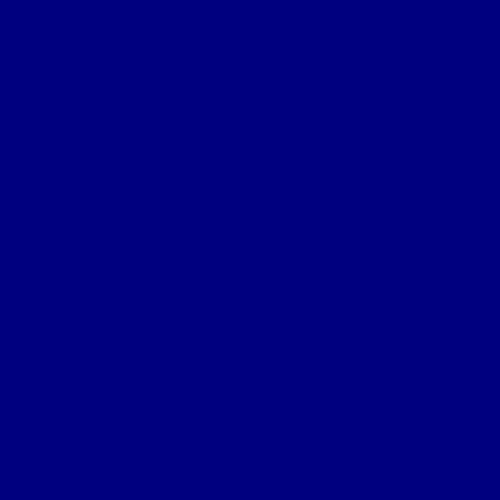} &
      \includegraphics[width=0.07\linewidth]{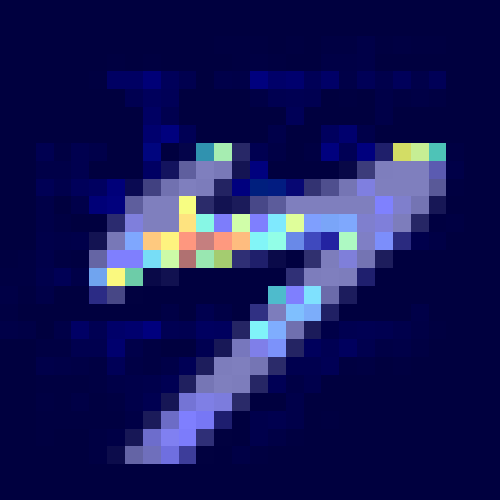} &
      \includegraphics[width=0.07\linewidth]{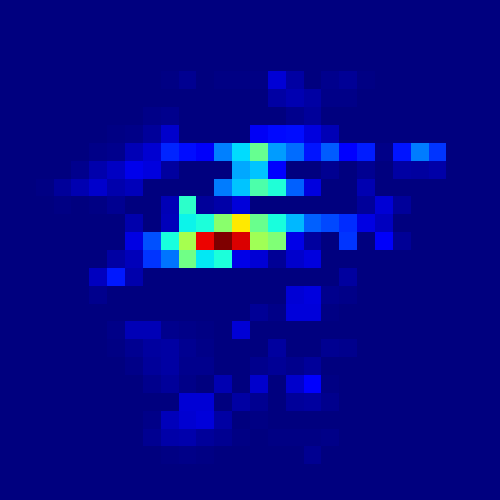} &
      \includegraphics[width=0.07\linewidth]{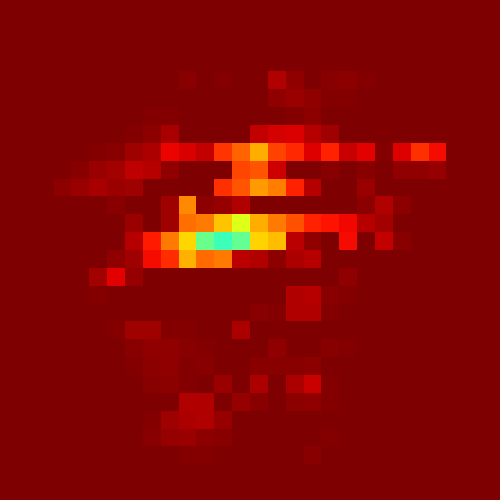} &
      \includegraphics[width=0.07\linewidth]{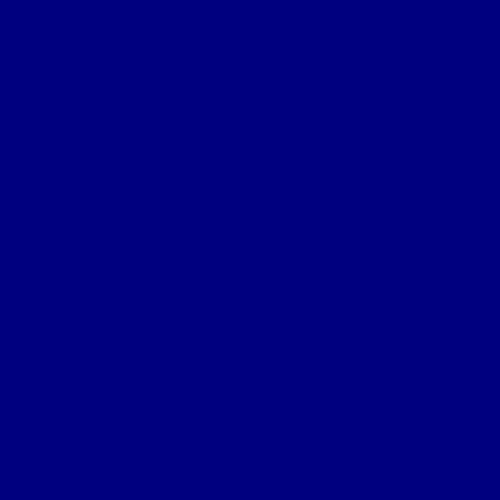} &
      \includegraphics[width=0.07\linewidth]{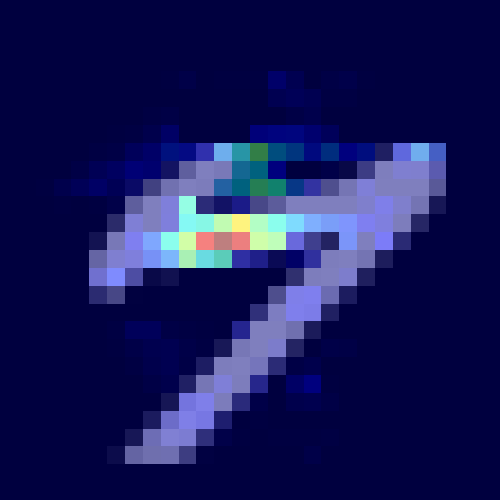} \\
      
      \includegraphics[width=0.07\linewidth]{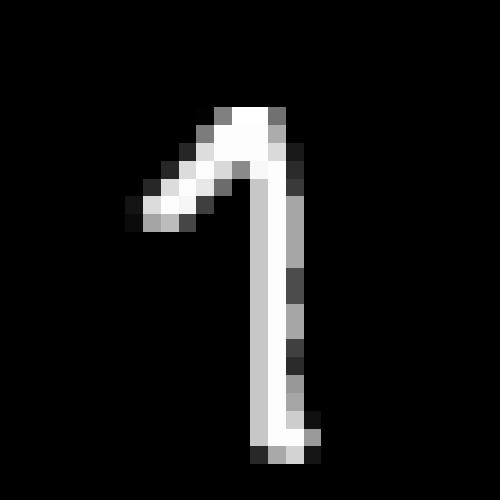} &
      \raisebox{0.4\height}{\shortstack{$b=0.90$ \\ $v=0.09$ \\ $d=0.00$}} &
      \includegraphics[width=0.07\linewidth]{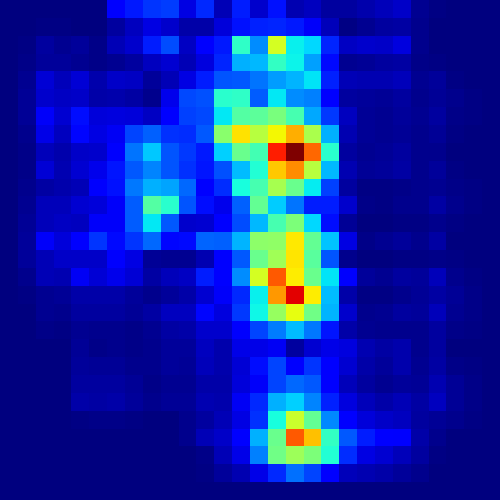} &
      \includegraphics[width=0.07\linewidth]{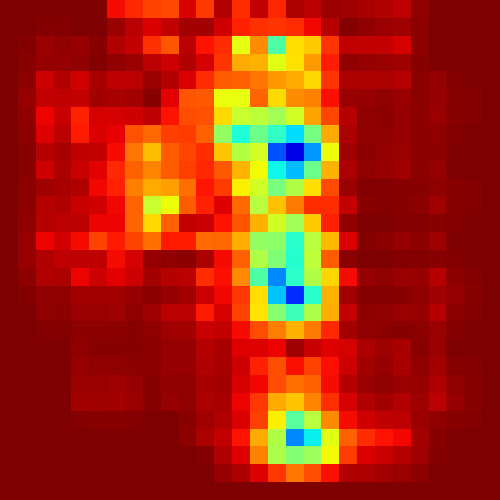} &
      \includegraphics[width=0.07\linewidth]{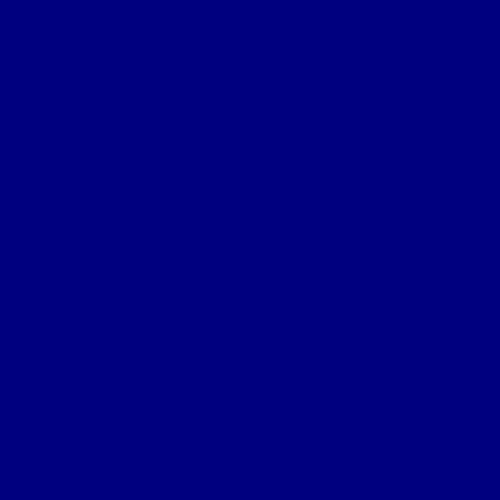} &
      \includegraphics[width=0.07\linewidth]{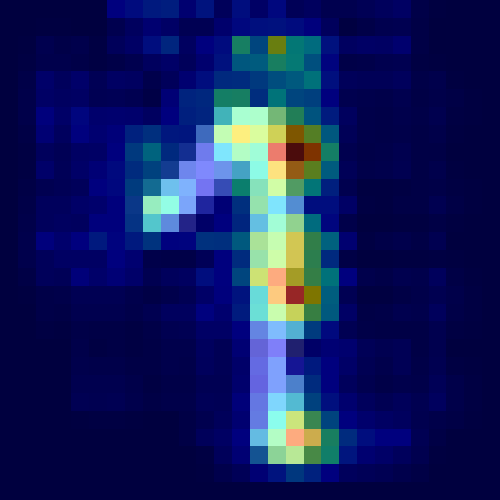} &
      \includegraphics[width=0.07\linewidth]{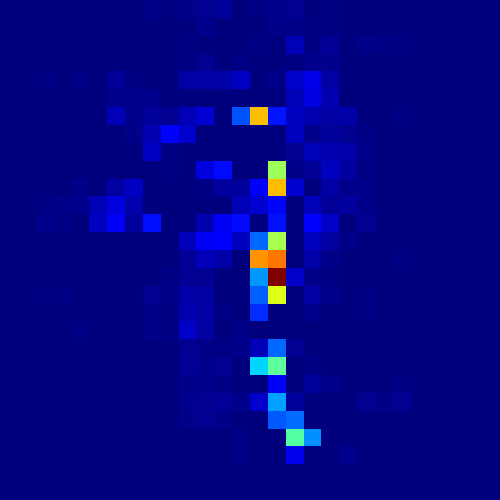} &
      \includegraphics[width=0.07\linewidth]{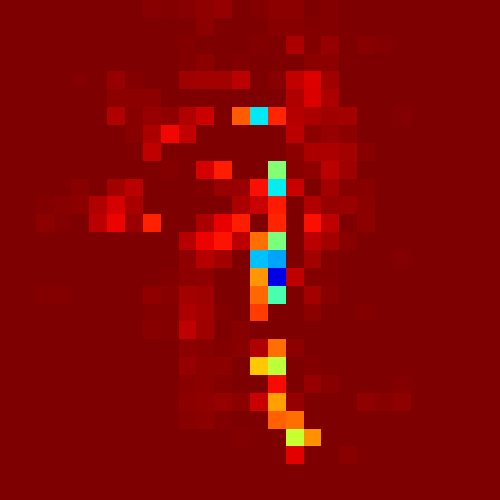} &
      \includegraphics[width=0.07\linewidth]{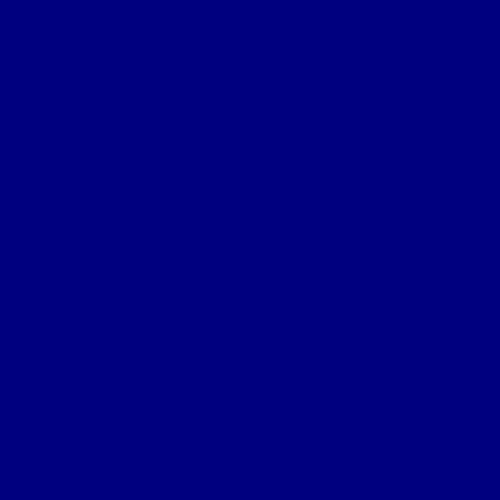} &
      \includegraphics[width=0.07\linewidth]{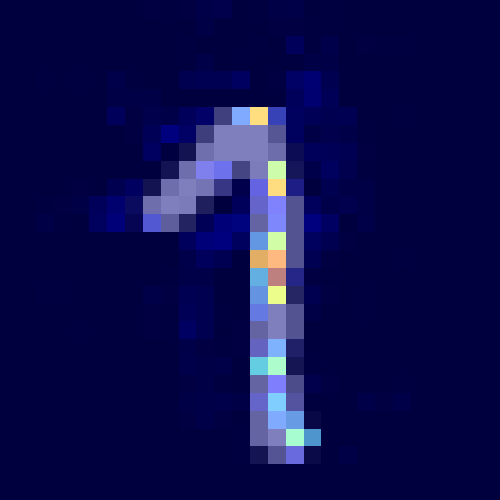} &
      \includegraphics[width=0.07\linewidth]{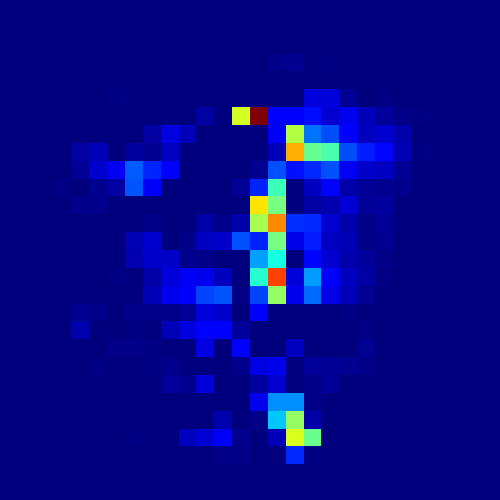} &
      \includegraphics[width=0.07\linewidth]{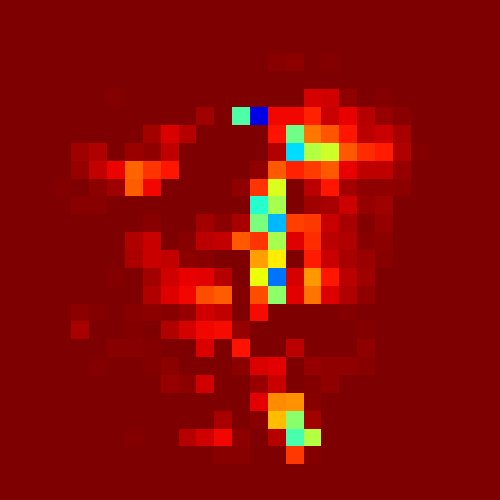} &
      \includegraphics[width=0.07\linewidth]{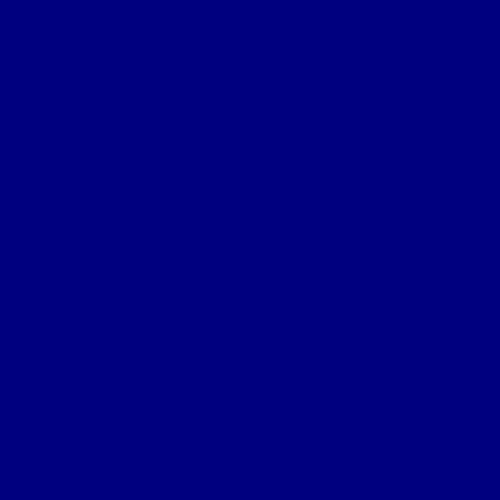} &
      \includegraphics[width=0.07\linewidth]{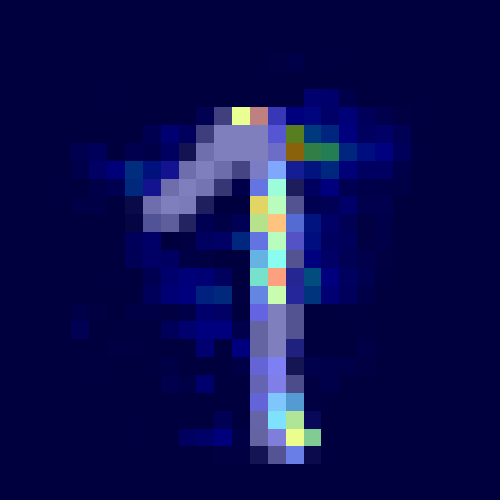} \\
      
      \includegraphics[width=0.07\linewidth]{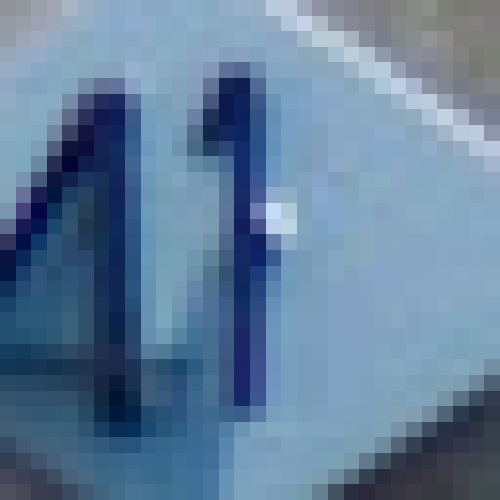} &
      \raisebox{0.4\height}{\shortstack{$b=0.55$ \\ $v=0.41$ \\ $d=0.05$}} &
      \includegraphics[width=0.07\linewidth]{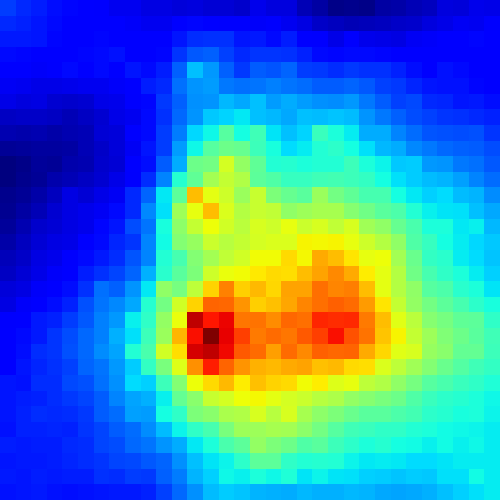} &
      \includegraphics[width=0.07\linewidth]{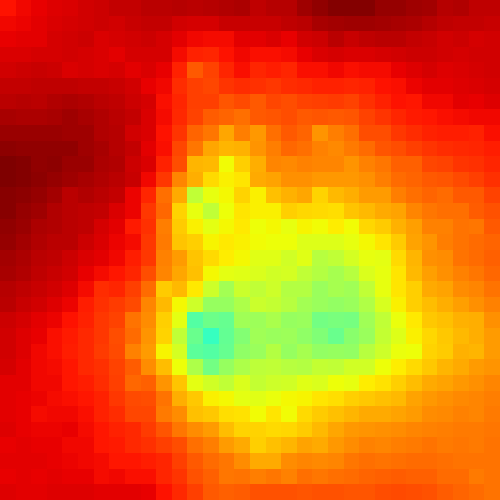} &
      \includegraphics[width=0.07\linewidth]{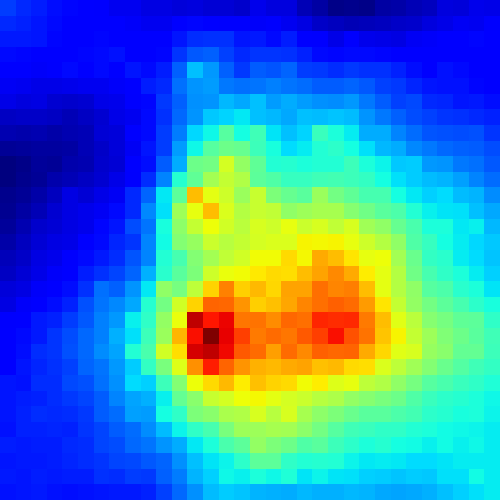} &
      \includegraphics[width=0.07\linewidth]{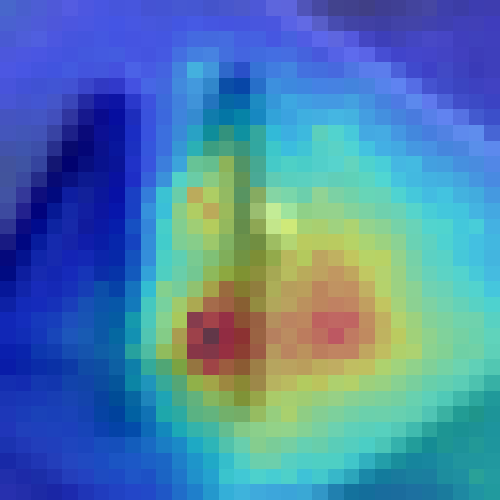} &
      \includegraphics[width=0.07\linewidth]{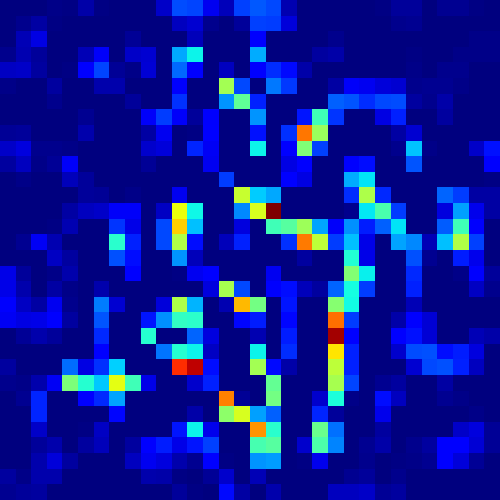} &
      \includegraphics[width=0.07\linewidth]{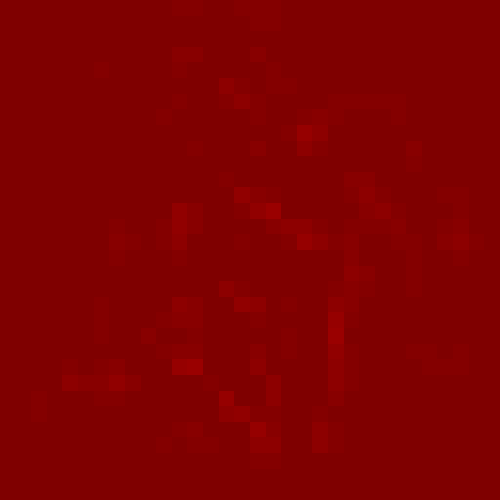} &
      \includegraphics[width=0.07\linewidth]{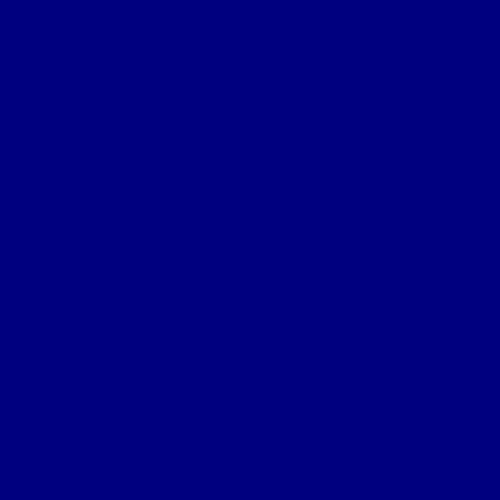} &
      \includegraphics[width=0.07\linewidth]{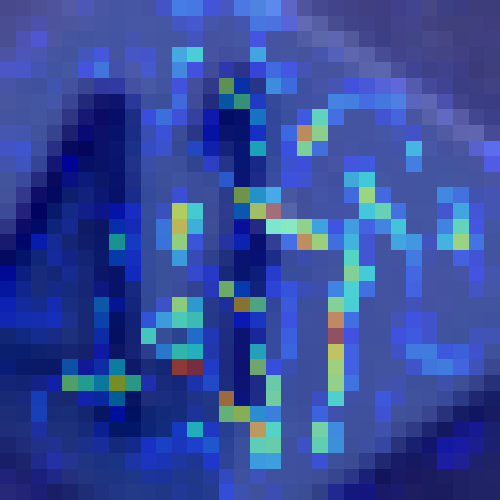} &
      \includegraphics[width=0.07\linewidth]{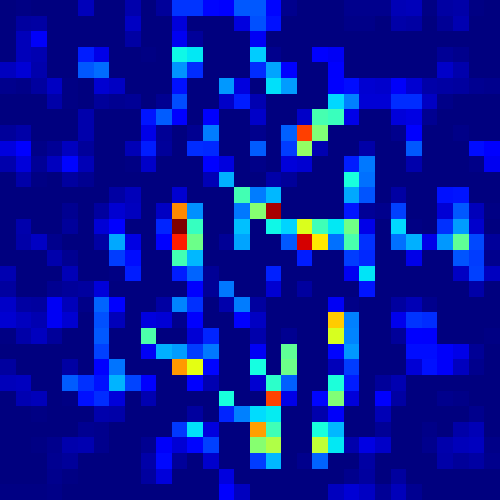} &
      \includegraphics[width=0.07\linewidth]{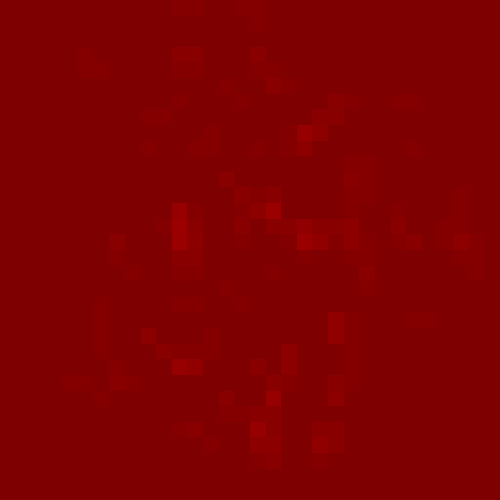} &
      \includegraphics[width=0.07\linewidth]{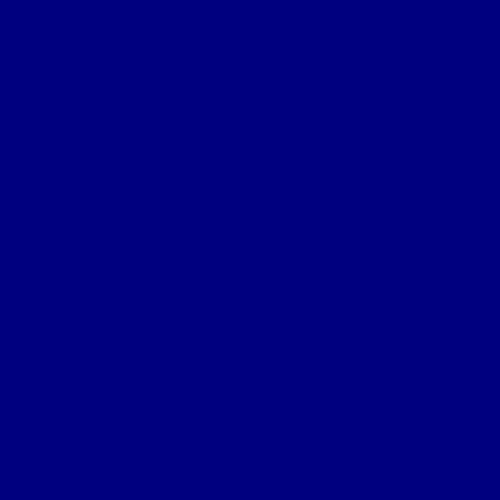} &
      \includegraphics[width=0.07\linewidth]{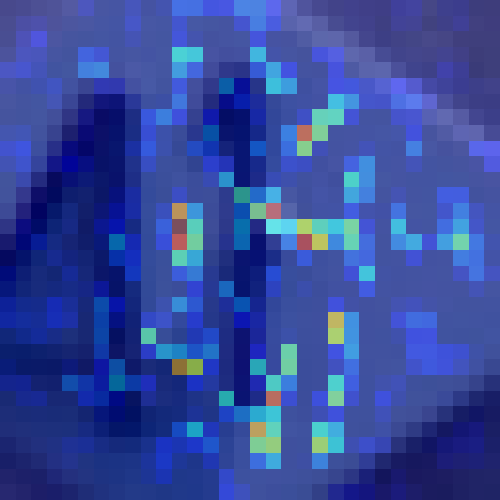} \\
      
      \includegraphics[width=0.07\linewidth]{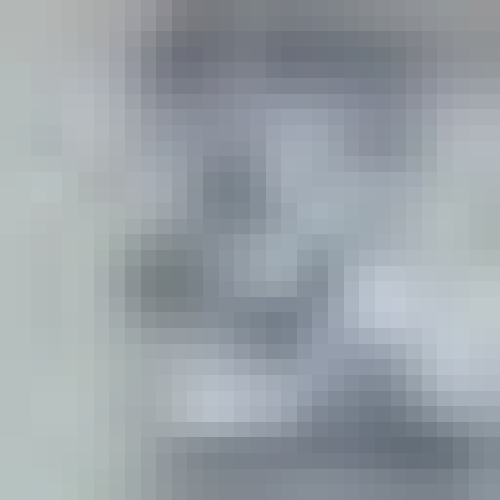} &
      \raisebox{0.4\height}{\shortstack{$b=0.18$ \\ $v=0.81$ \\ $d=0.00$}} &
      \includegraphics[width=0.07\linewidth]{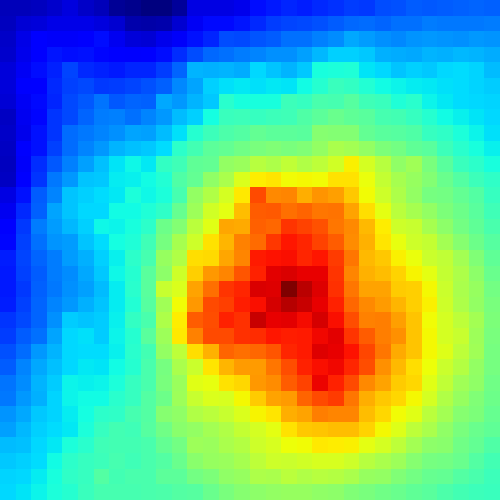} &
      \includegraphics[width=0.07\linewidth]{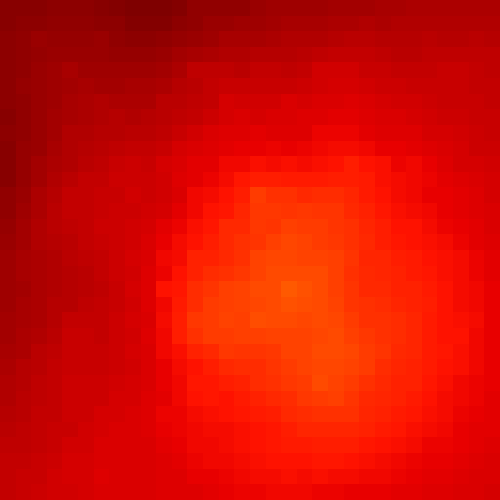} &
      \includegraphics[width=0.07\linewidth]{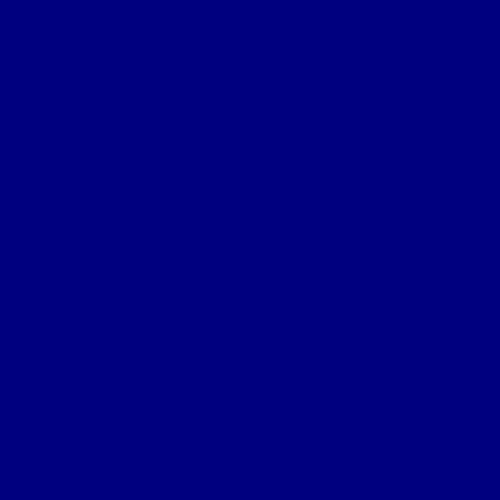} &
      \includegraphics[width=0.07\linewidth]{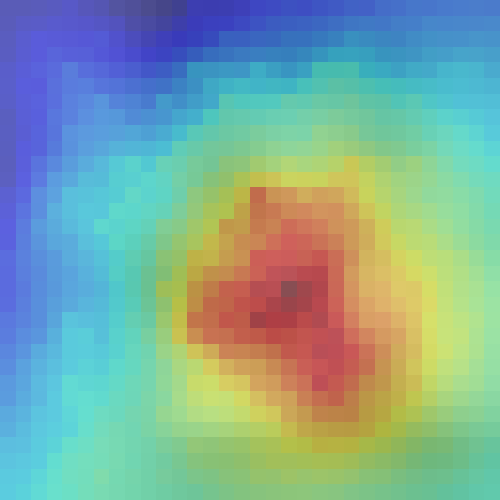} &
      \includegraphics[width=0.07\linewidth]{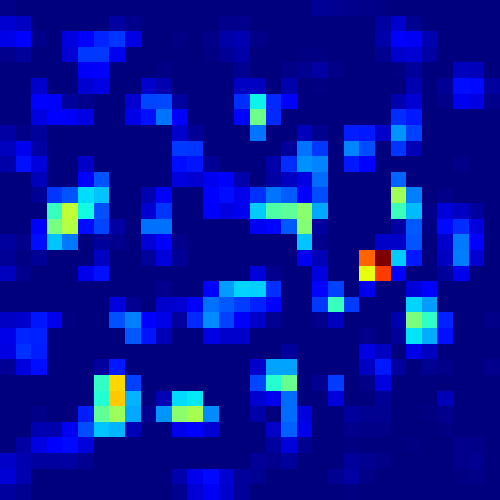} &
      \includegraphics[width=0.07\linewidth]{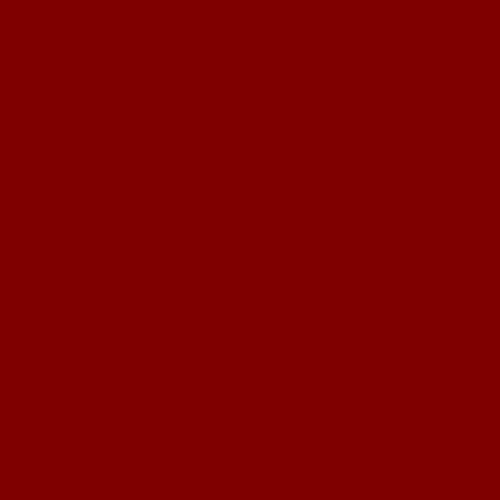} &
      \includegraphics[width=0.07\linewidth]{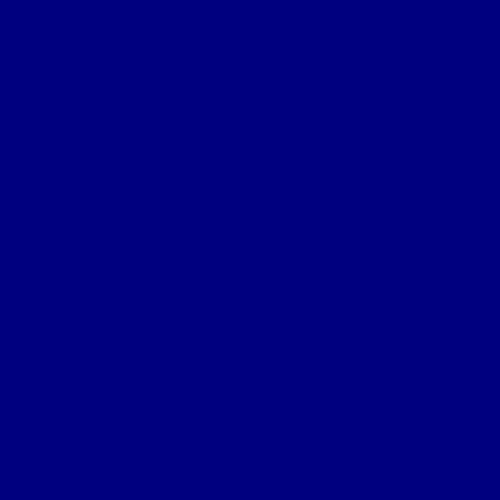} &
      \includegraphics[width=0.07\linewidth]{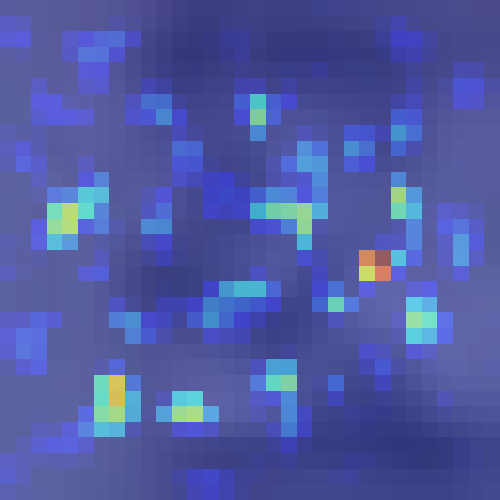} &
      \includegraphics[width=0.07\linewidth]{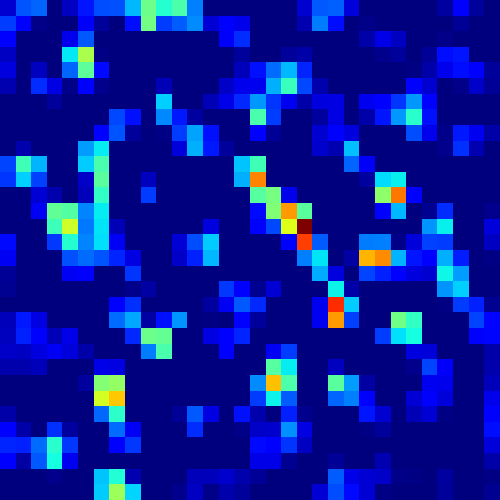} &
      \includegraphics[width=0.07\linewidth]{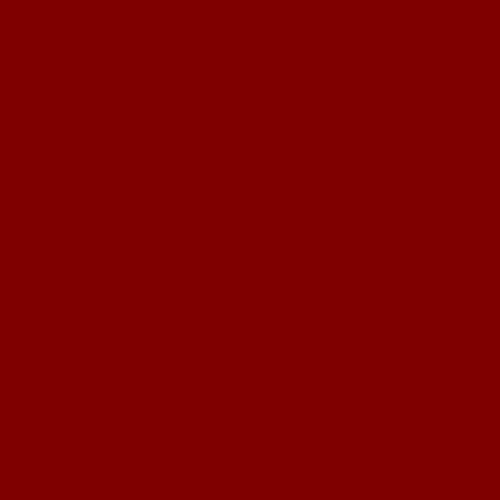} &
      \includegraphics[width=0.07\linewidth]{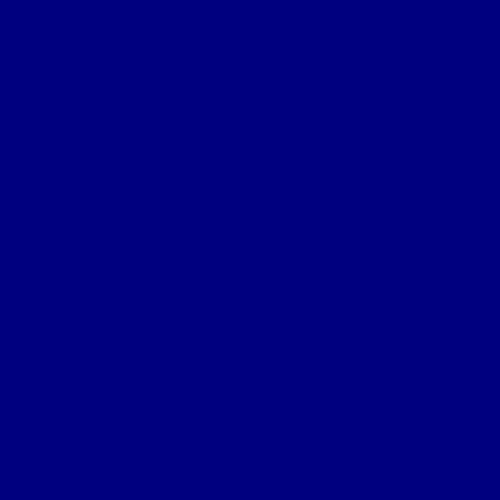} &
      \includegraphics[width=0.07\linewidth]{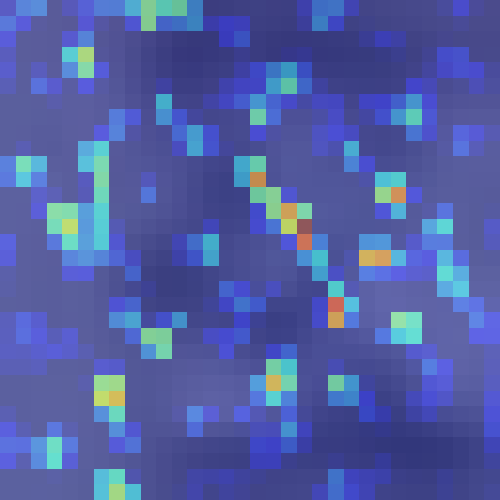} \\
      
      \includegraphics[width=0.07\linewidth]{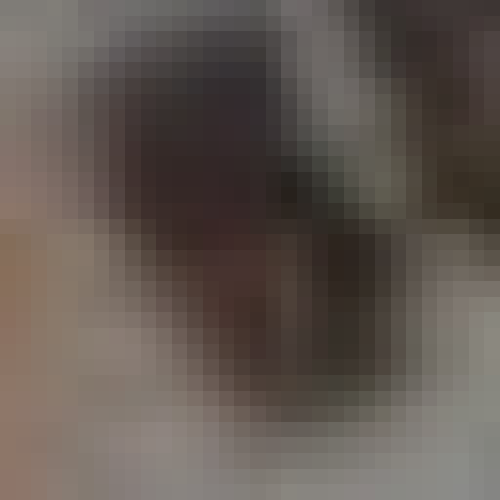} &
      \raisebox{0.4\height}{\shortstack{$b=0.41$ \\ $v=0.58$ \\ $d=0.00$}} &
      \includegraphics[width=0.07\linewidth]{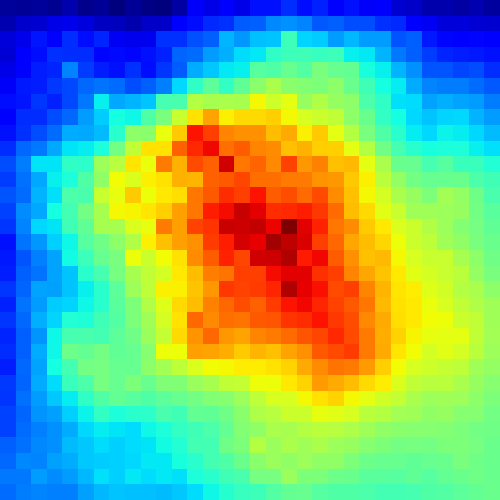} &
      \includegraphics[width=0.07\linewidth]{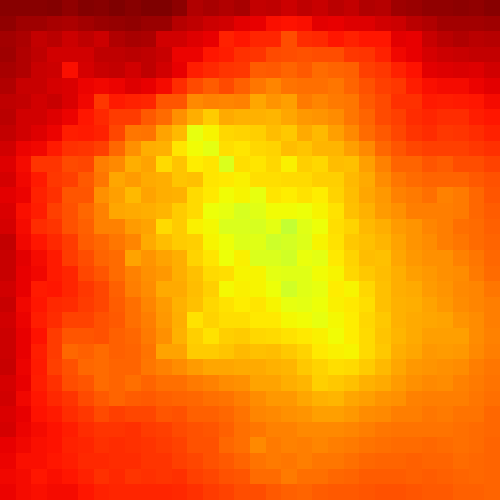} &
      \includegraphics[width=0.07\linewidth]{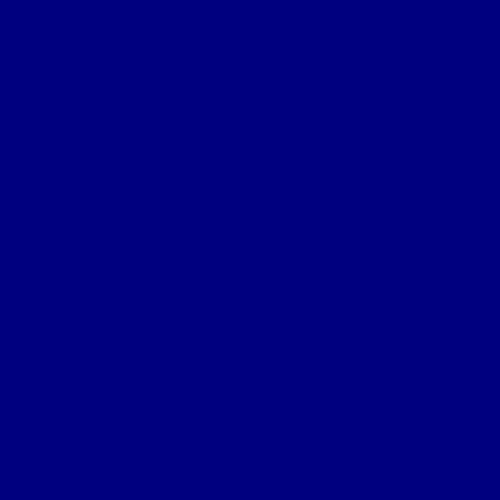} &
      \includegraphics[width=0.07\linewidth]{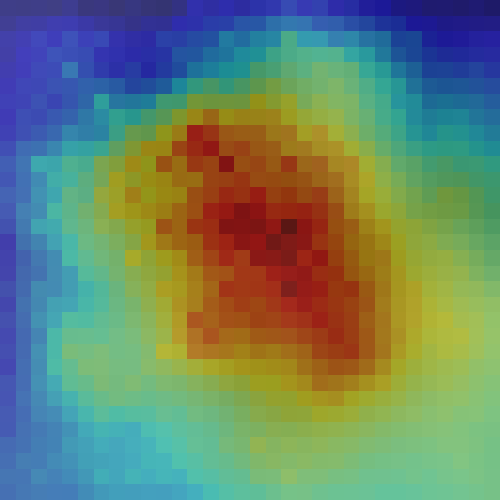} &
      \includegraphics[width=0.07\linewidth]{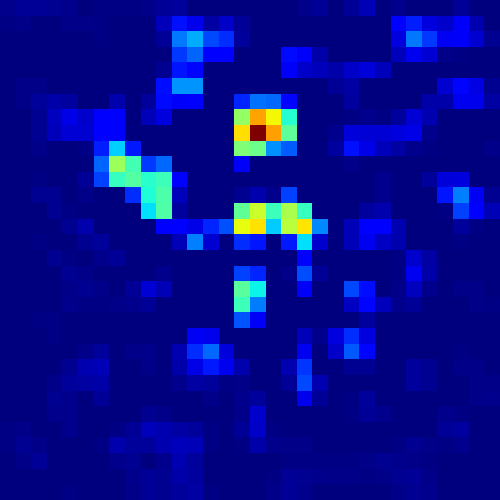} &
      \includegraphics[width=0.07\linewidth]{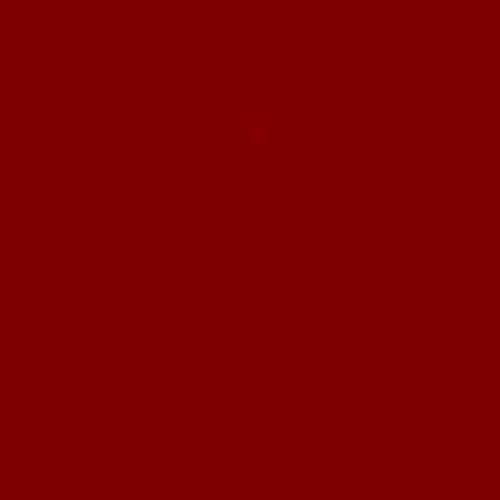} &
      \includegraphics[width=0.07\linewidth]{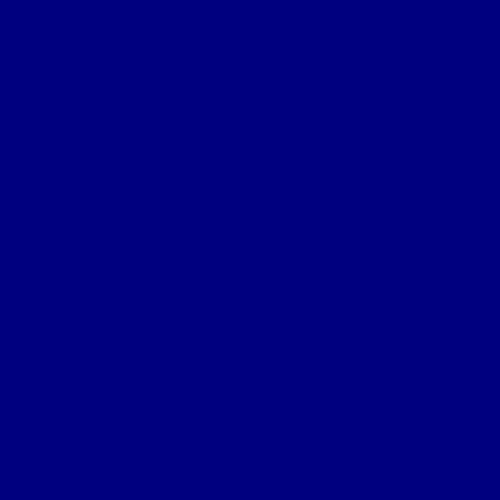} &
      \includegraphics[width=0.07\linewidth]{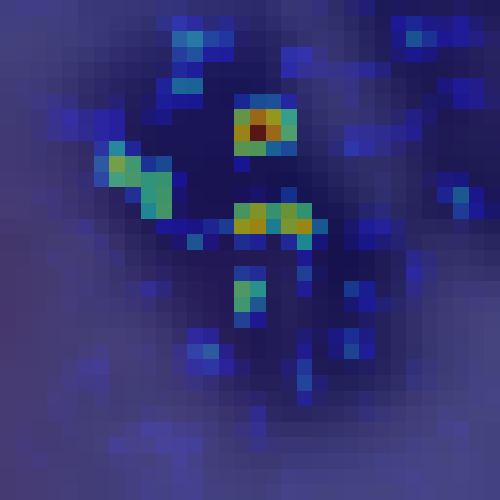} &
      \includegraphics[width=0.07\linewidth]{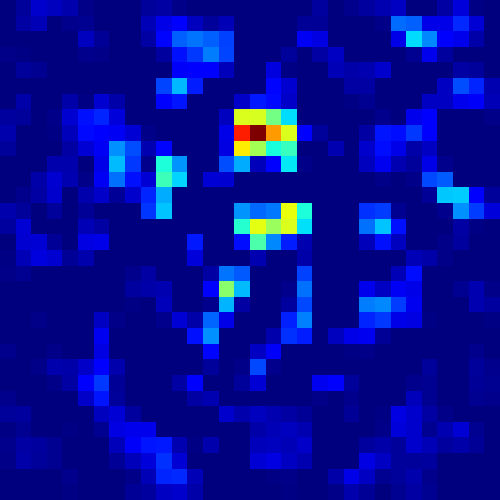} &
      \includegraphics[width=0.07\linewidth]{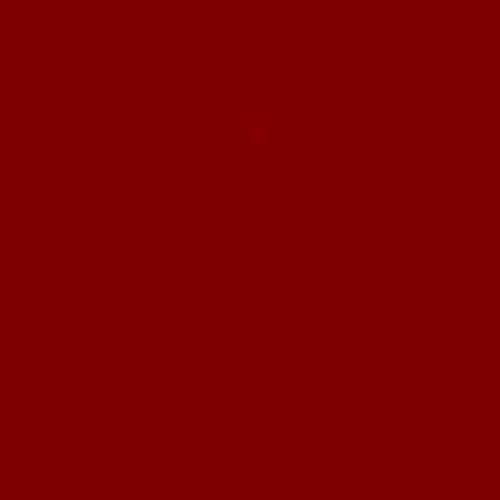} &
      \includegraphics[width=0.07\linewidth]{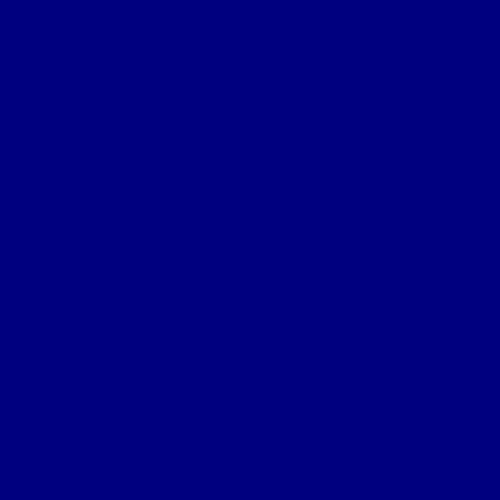} &
      \includegraphics[width=0.07\linewidth]{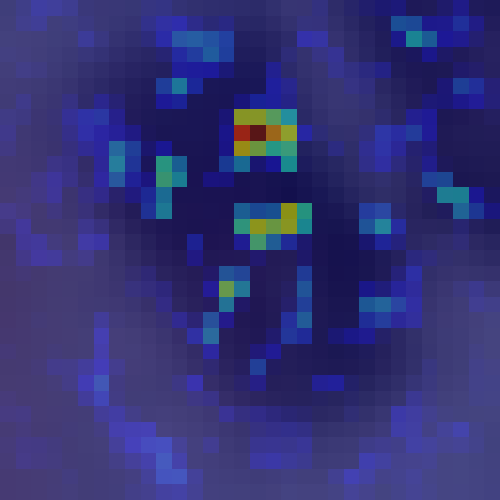} \\
      
      \includegraphics[width=0.07\linewidth]{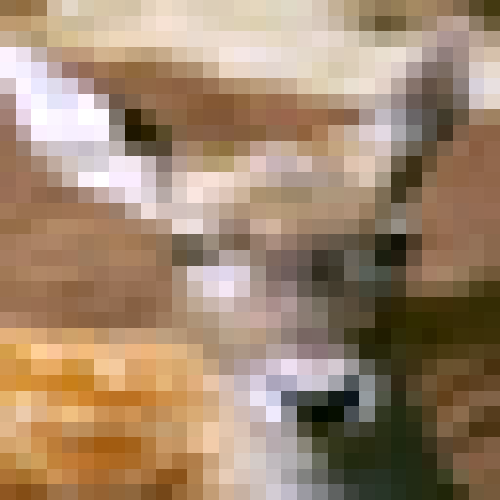} &
      \raisebox{0.4\height}{\shortstack{$b=0.42$ \\ $v=0.38$ \\ $d=0.37$}} &
      \includegraphics[width=0.07\linewidth]{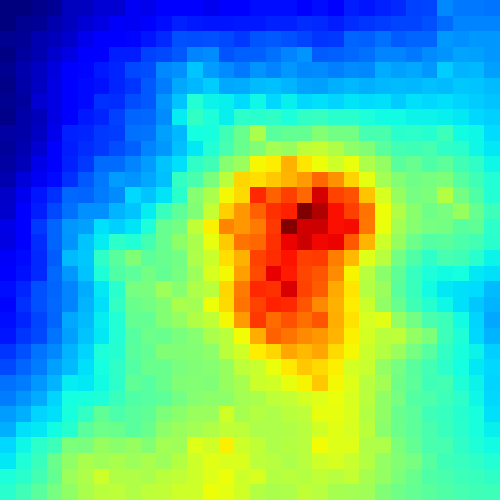} &
      \includegraphics[width=0.07\linewidth]{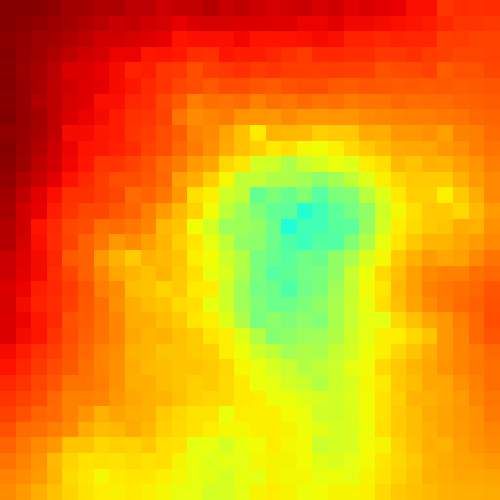} &
      \includegraphics[width=0.07\linewidth]{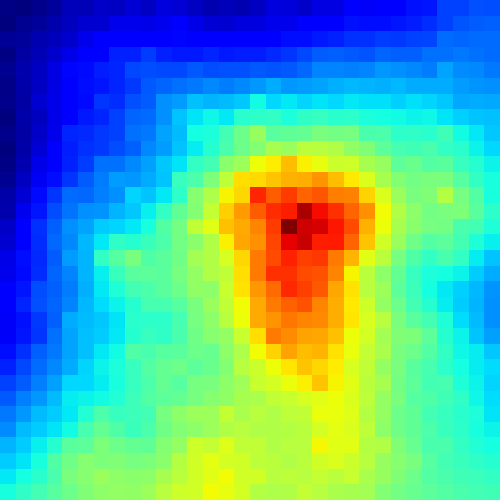} &
      \includegraphics[width=0.07\linewidth]{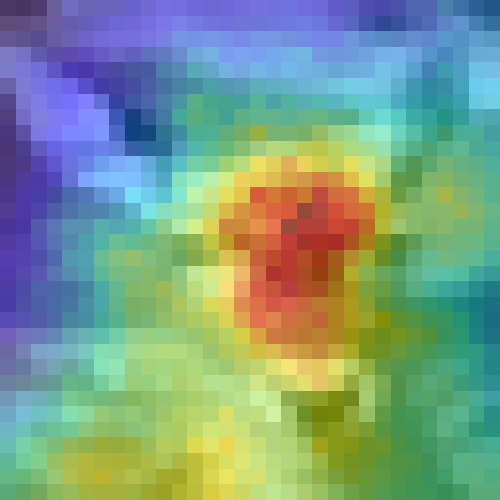} &
      \includegraphics[width=0.07\linewidth]{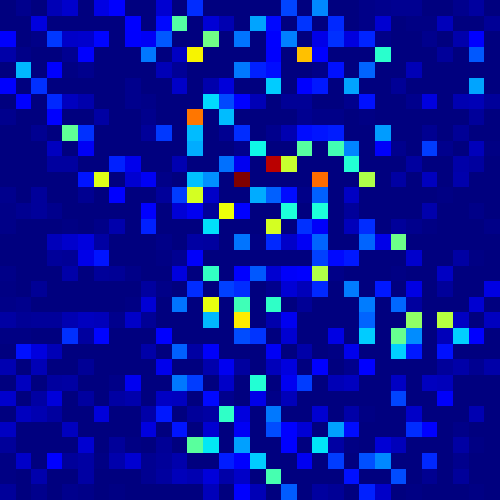} &
      \includegraphics[width=0.07\linewidth]{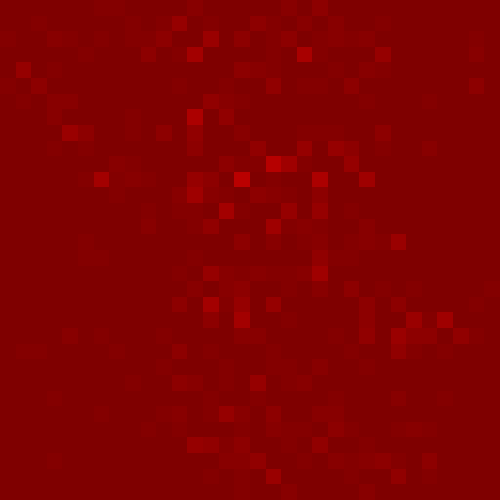} &
      \includegraphics[width=0.07\linewidth]{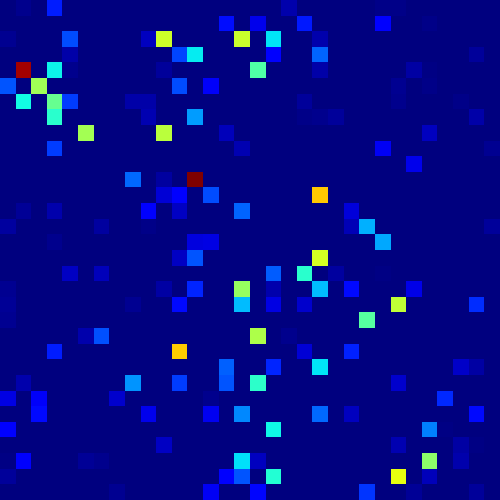} &
      \includegraphics[width=0.07\linewidth]{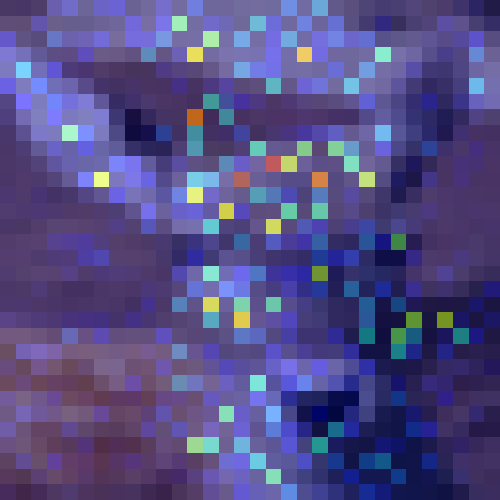} &
      \includegraphics[width=0.07\linewidth]{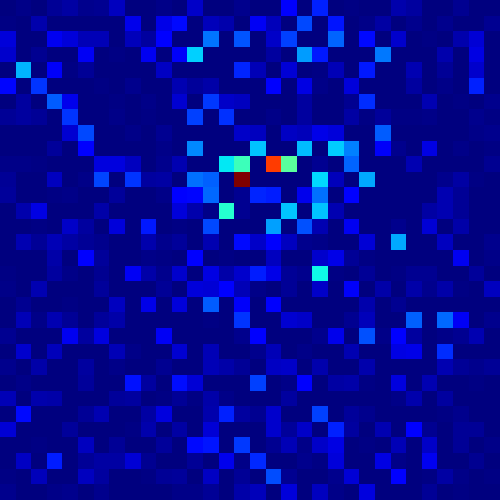} &
      \includegraphics[width=0.07\linewidth]{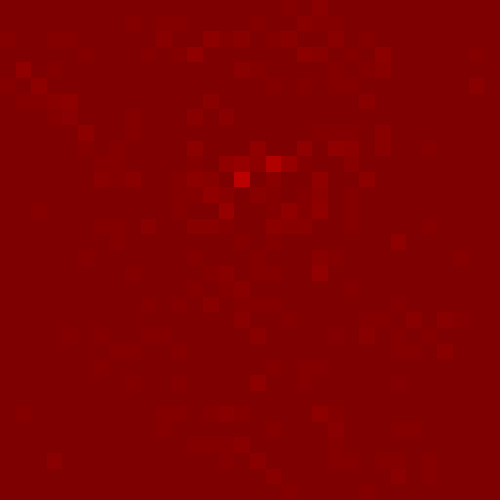} &
      \includegraphics[width=0.07\linewidth]{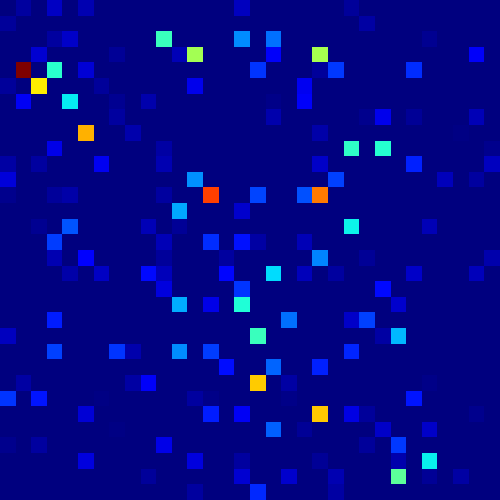} &
      \includegraphics[width=0.07\linewidth]{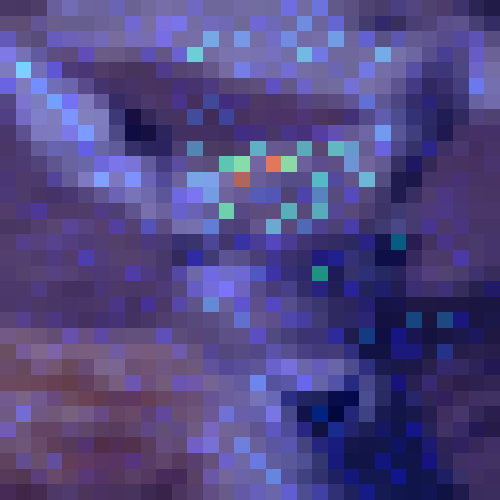} \\
    
      \includegraphics[width=0.07\linewidth]{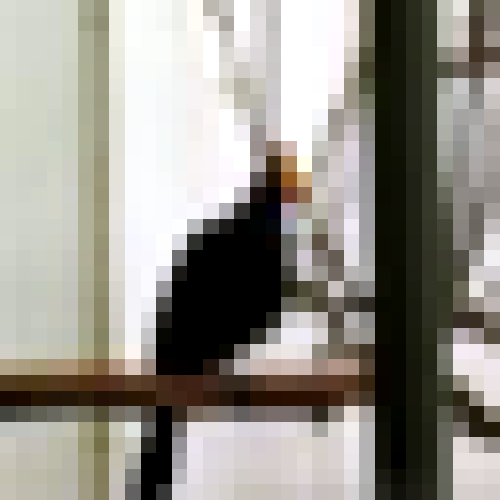} &
      \raisebox{0.4\height}{\shortstack{$b=0.68$ \\ $v=0.27$ \\ $d=0.08$}} &
      \includegraphics[width=0.07\linewidth]{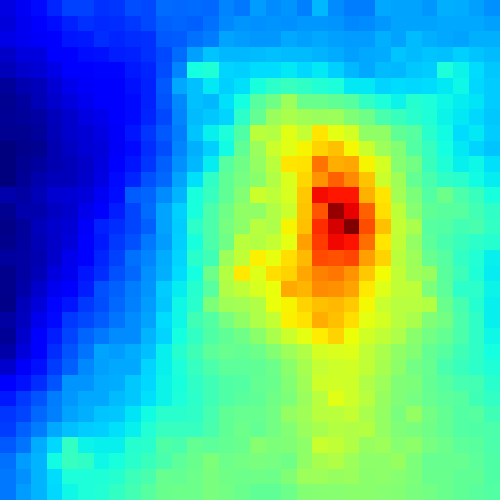} &
      \includegraphics[width=0.07\linewidth]{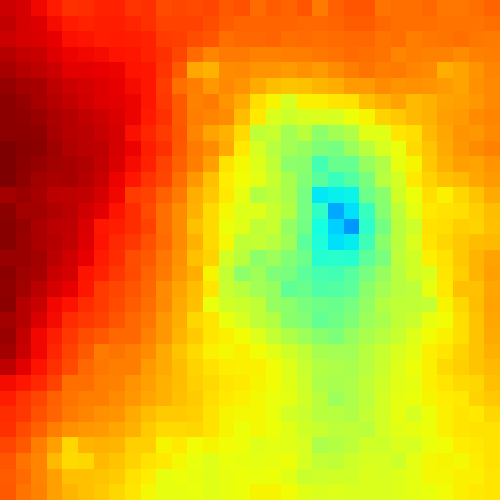} &
      \includegraphics[width=0.07\linewidth]{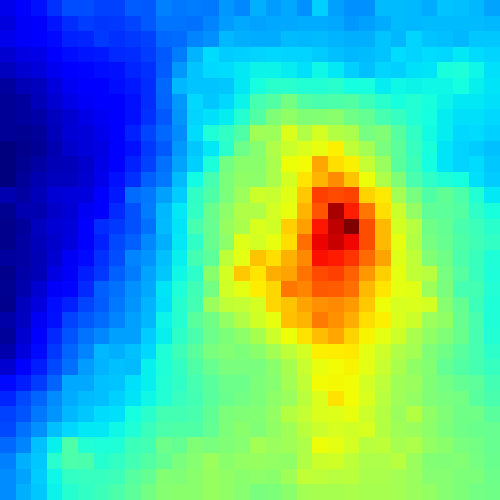} &
      \includegraphics[width=0.07\linewidth]{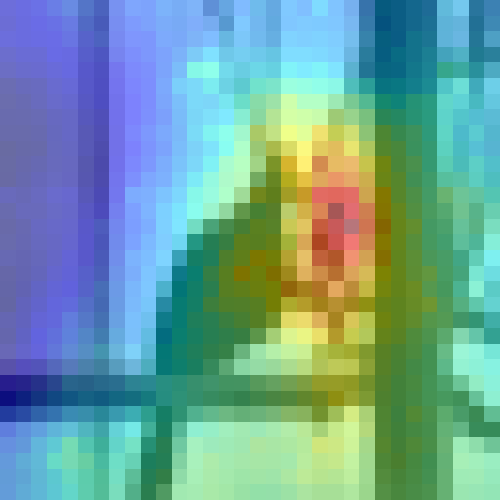} &
      \includegraphics[width=0.07\linewidth]{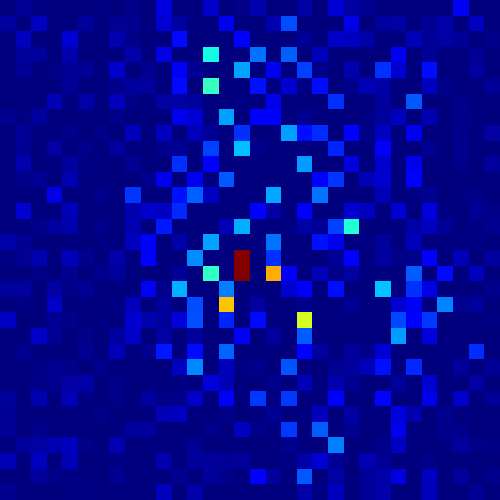} &
      \includegraphics[width=0.07\linewidth]{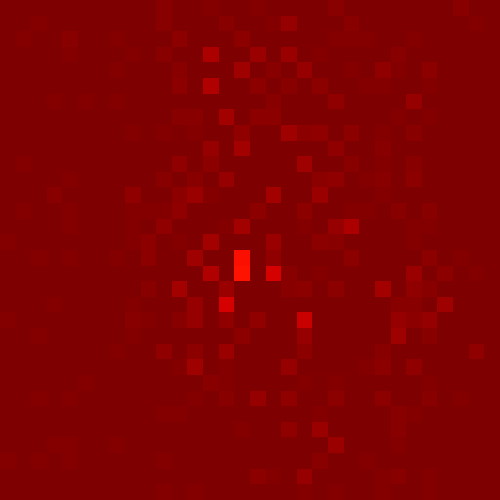} &
      \includegraphics[width=0.07\linewidth]{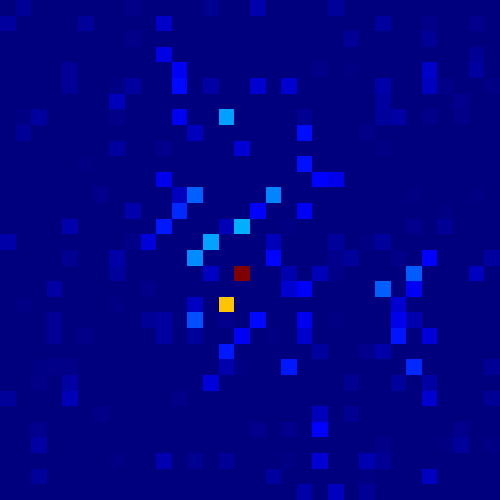} &
      \includegraphics[width=0.07\linewidth]{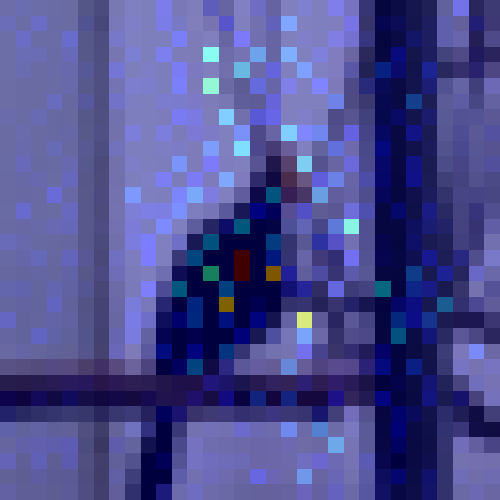} &
      \includegraphics[width=0.07\linewidth]{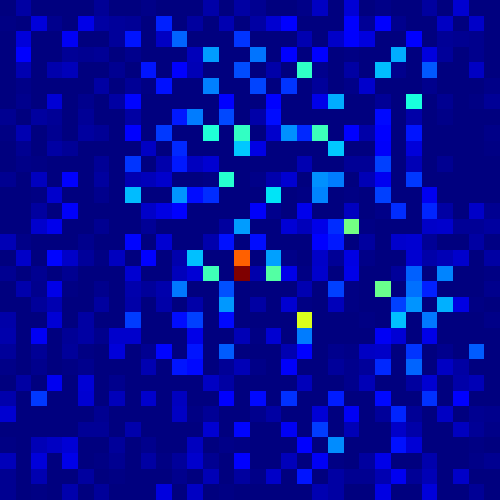} &
      \includegraphics[width=0.07\linewidth]{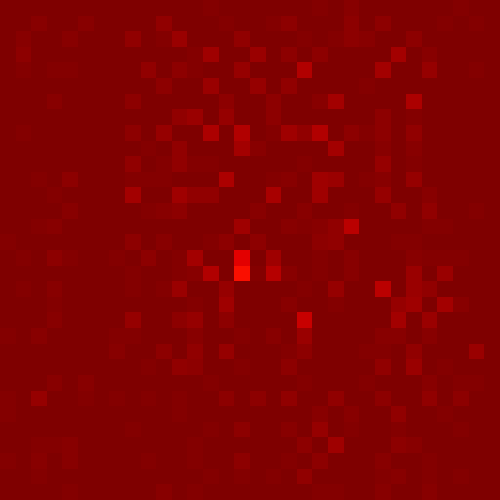} &
      \includegraphics[width=0.07\linewidth]{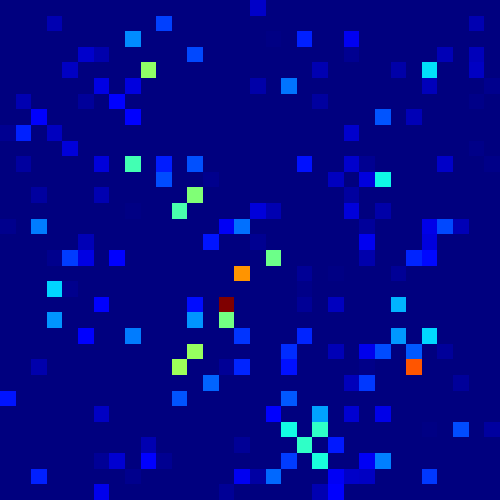} &
      \includegraphics[width=0.07\linewidth]{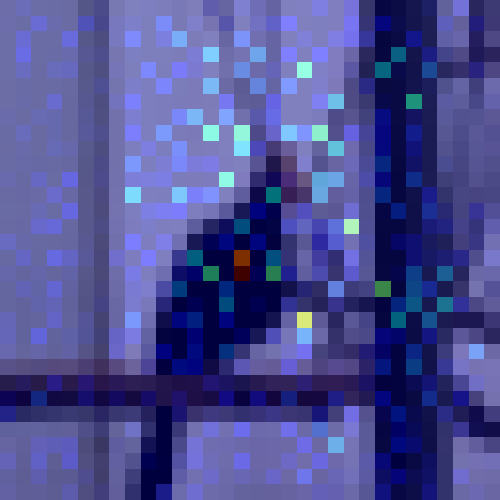} \\
    
      \includegraphics[width=0.07\linewidth]{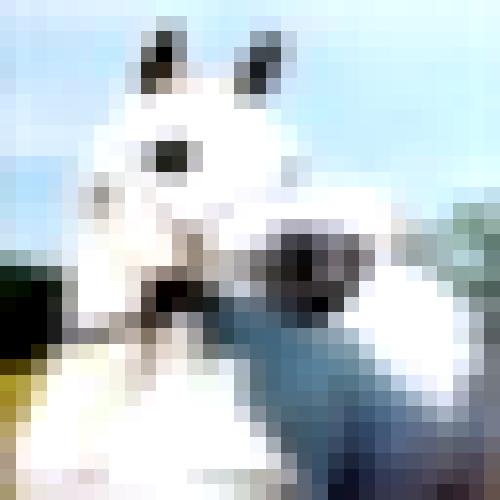} &
      \raisebox{0.4\height}{\shortstack{$b=0.70$ \\ $v=0.25$ \\ $d=0.06$}} &
      \includegraphics[width=0.07\linewidth]{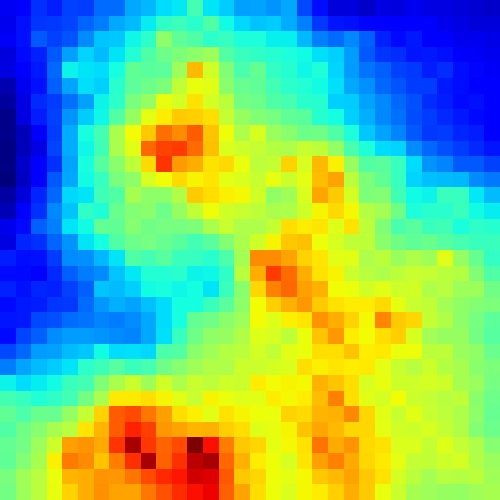} &
      \includegraphics[width=0.07\linewidth]{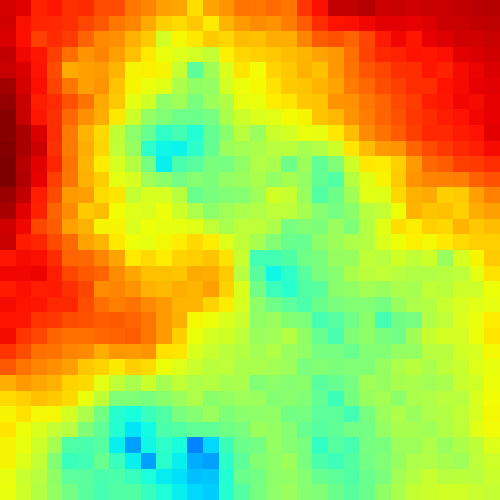} &
      \includegraphics[width=0.07\linewidth]{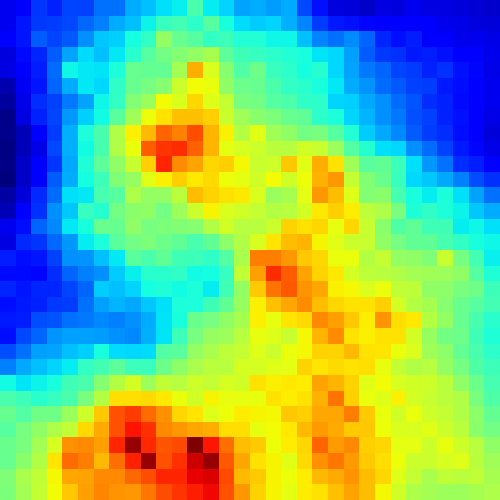} &
      \includegraphics[width=0.07\linewidth]{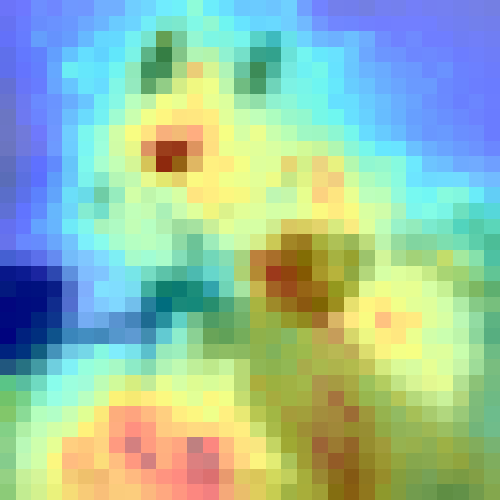} &
      \includegraphics[width=0.07\linewidth]{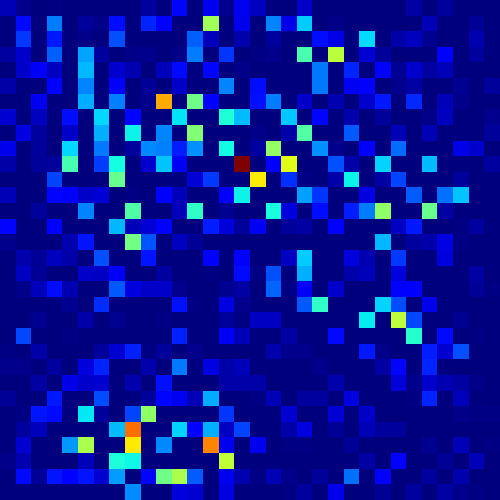} &
      \includegraphics[width=0.07\linewidth]{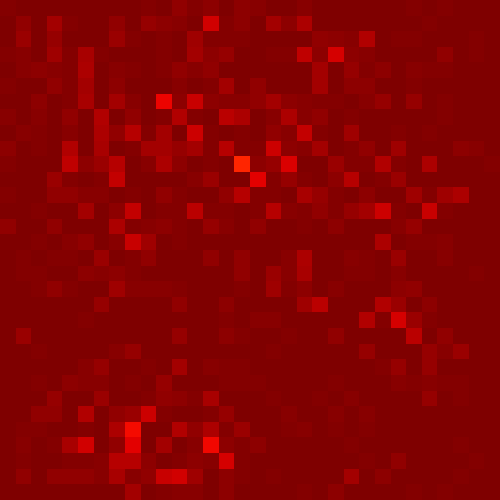} &
      \includegraphics[width=0.07\linewidth]{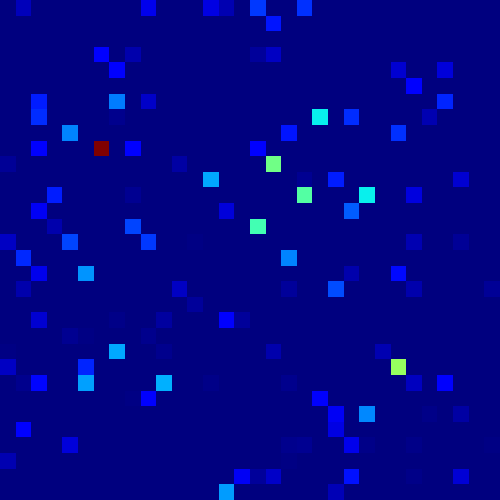} &
      \includegraphics[width=0.07\linewidth]{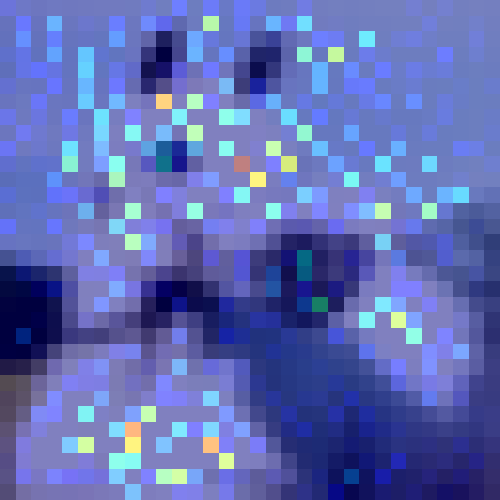} &
      \includegraphics[width=0.07\linewidth]{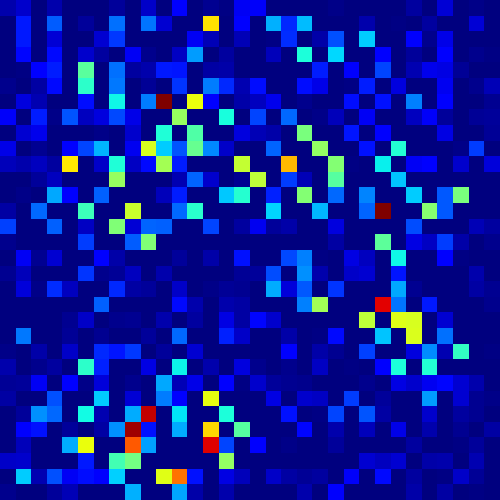} &
      \includegraphics[width=0.07\linewidth]{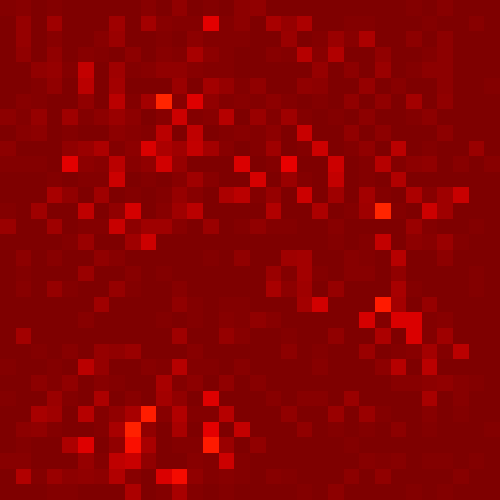} &
      \includegraphics[width=0.07\linewidth]{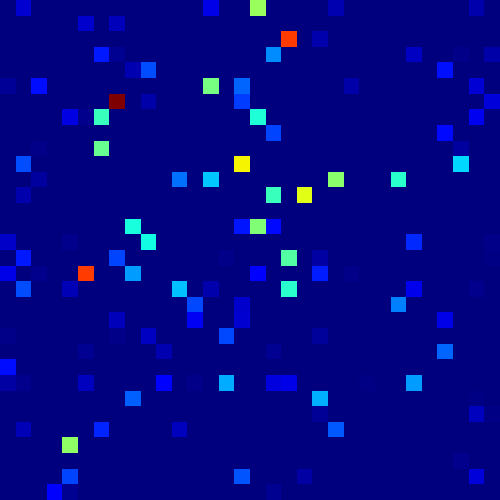} &
      \includegraphics[width=0.07\linewidth]{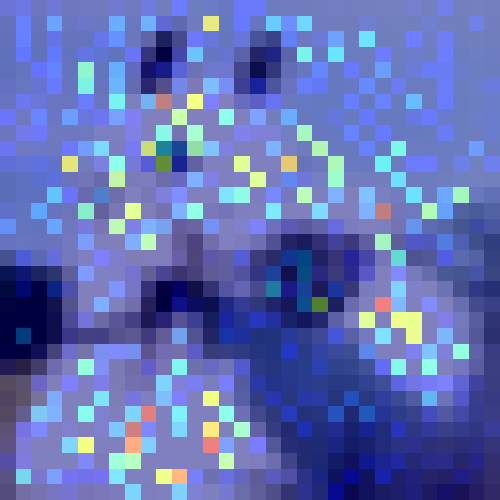} \\
            
      \includegraphics[width=0.07\linewidth]{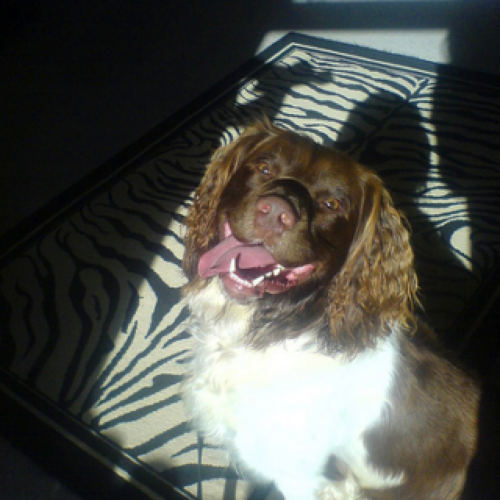} &
      \raisebox{0.4\height}{\shortstack{$b=0.13$ \\ $v=0.77$ \\ $d=0.17$}} &
      \includegraphics[width=0.07\linewidth]{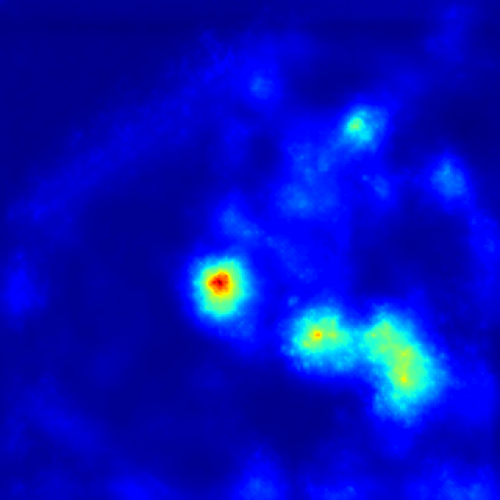} &
      \includegraphics[width=0.07\linewidth]{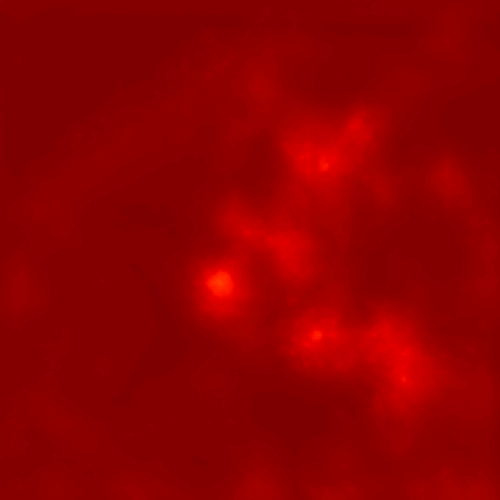} &
      \includegraphics[width=0.07\linewidth]{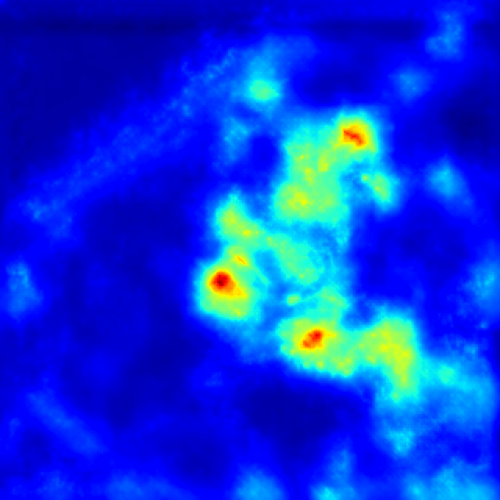} &
      \includegraphics[width=0.07\linewidth]{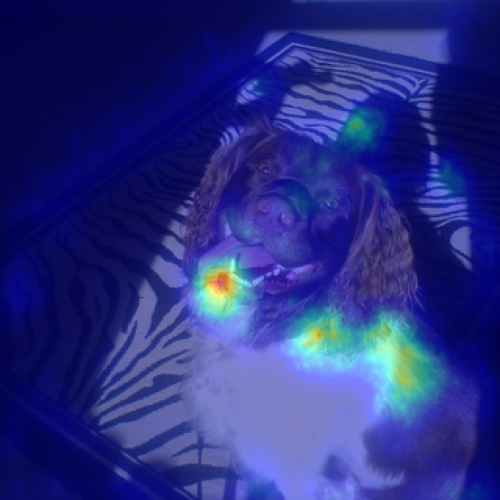} &
      \includegraphics[width=0.07\linewidth]{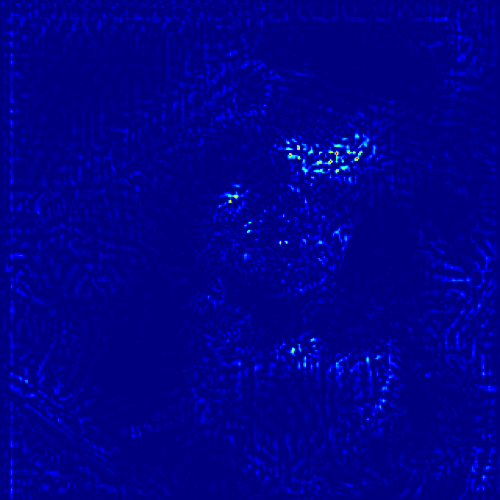} &
      \includegraphics[width=0.07\linewidth]{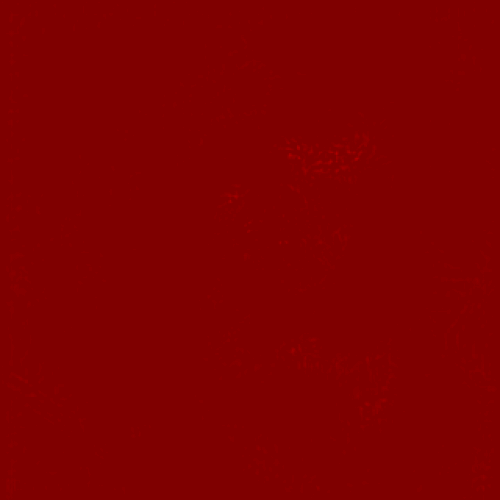} &
      \includegraphics[width=0.07\linewidth]{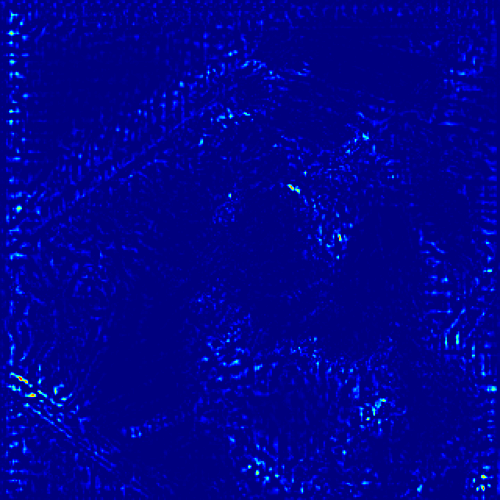} &
      \includegraphics[width=0.07\linewidth]{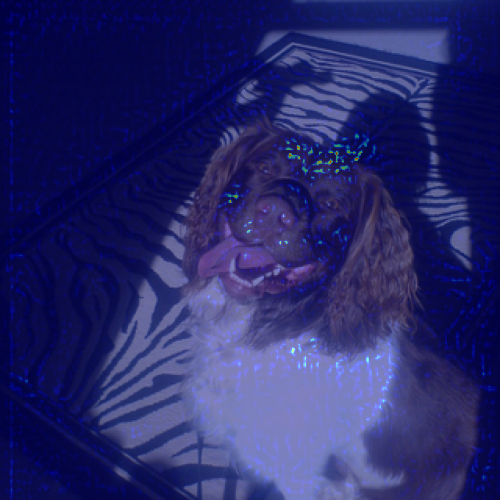} &
      \includegraphics[width=0.07\linewidth]{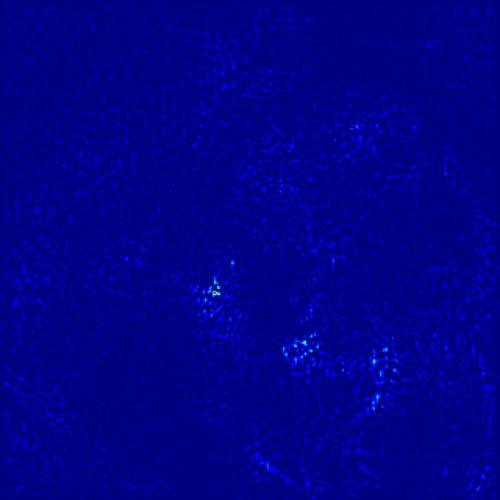} &
      \includegraphics[width=0.07\linewidth]{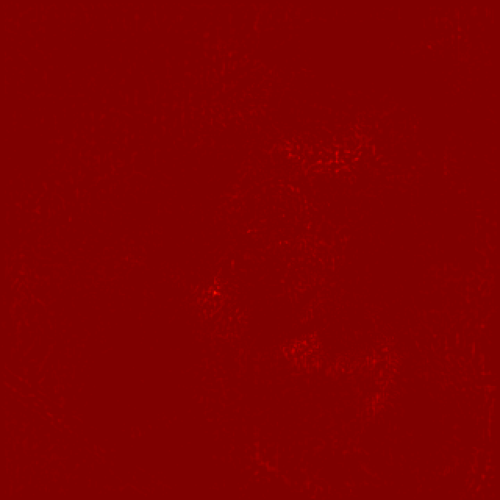} &
      \includegraphics[width=0.07\linewidth]{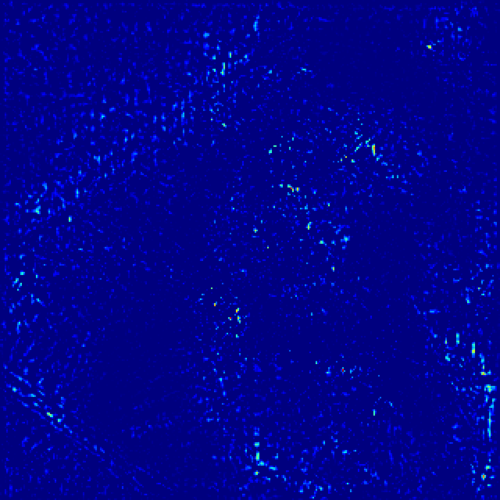} &
      \includegraphics[width=0.07\linewidth]{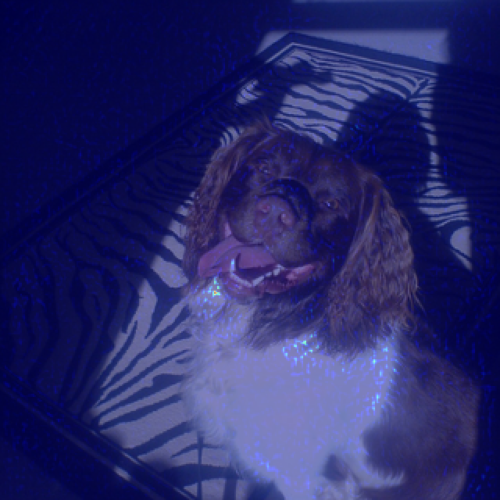} \\

      \includegraphics[width=0.07\linewidth]{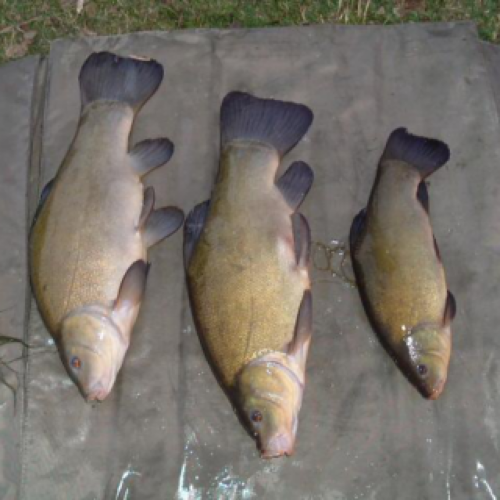} &
      \raisebox{0.4\height}{\shortstack{$b=0.42$ \\ $v=0.52$ \\ $d=0.11$}} &
      \includegraphics[width=0.07\linewidth]{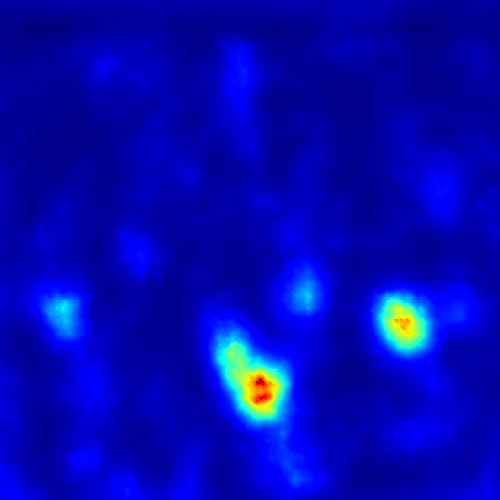} &
      \includegraphics[width=0.07\linewidth]{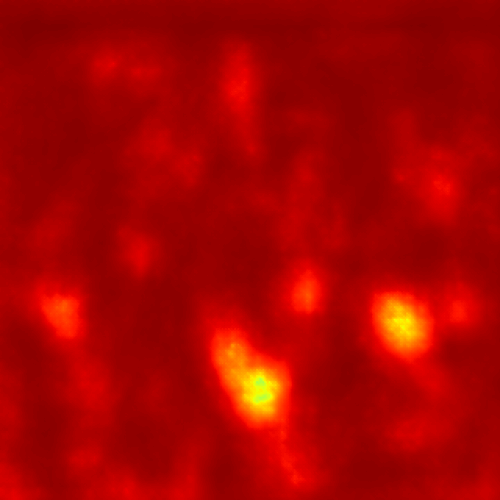} &
      \includegraphics[width=0.07\linewidth]{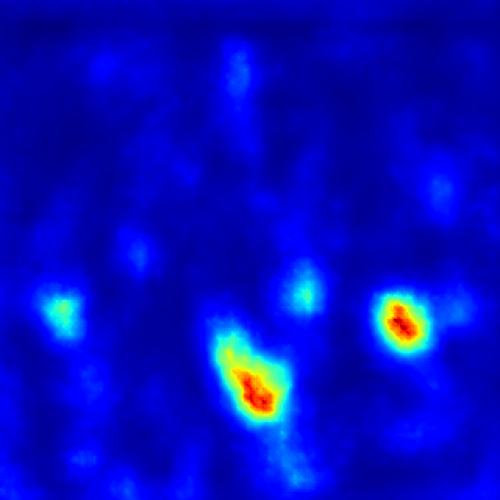} &
      \includegraphics[width=0.07\linewidth]{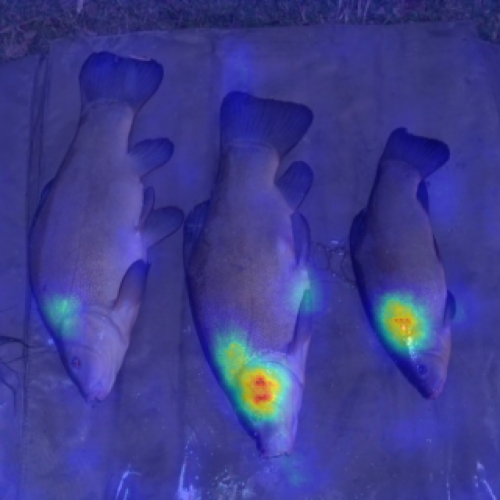} &
      \includegraphics[width=0.07\linewidth]{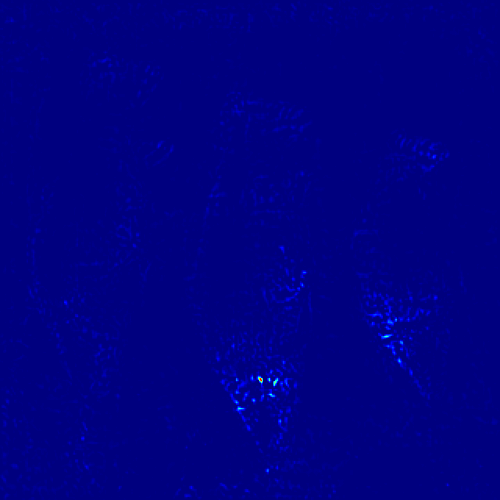} &
      \includegraphics[width=0.07\linewidth]{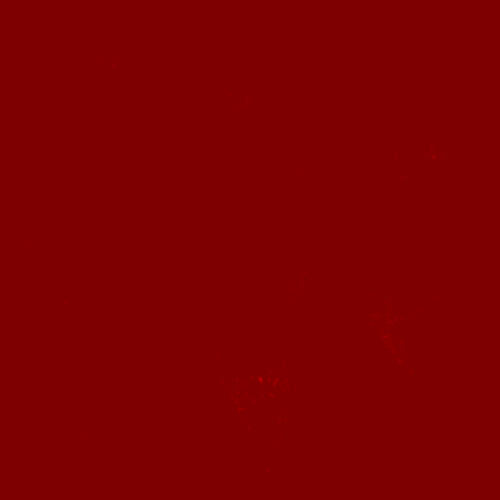} &
      \includegraphics[width=0.07\linewidth]{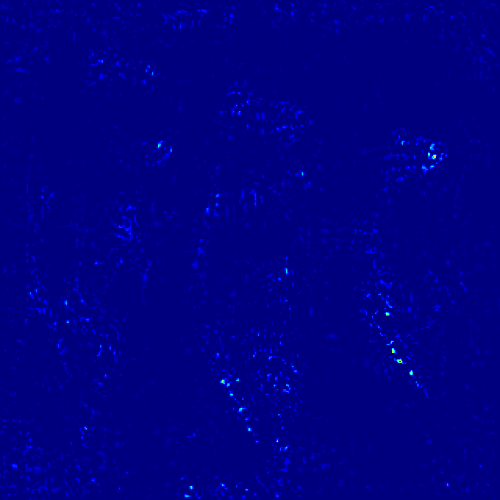} &
      \includegraphics[width=0.07\linewidth]{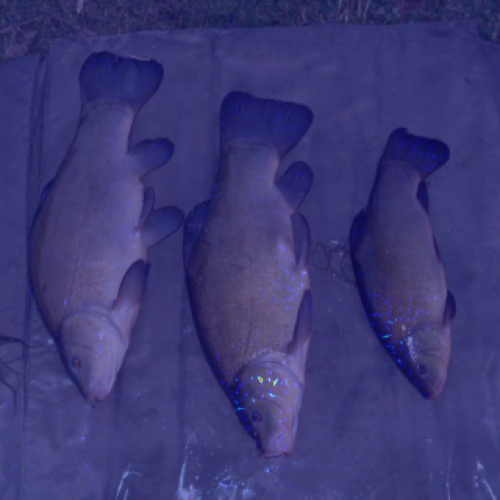} &
      \includegraphics[width=0.07\linewidth]{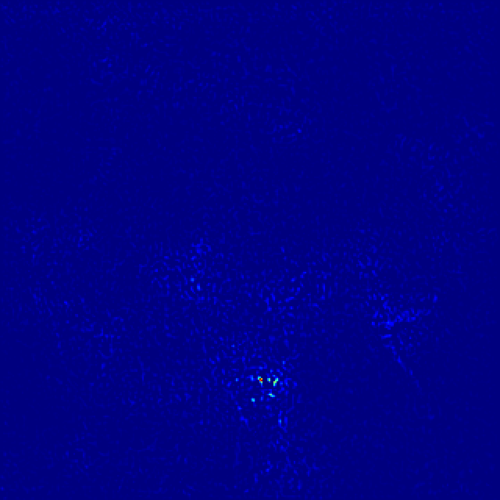} &
      \includegraphics[width=0.07\linewidth]{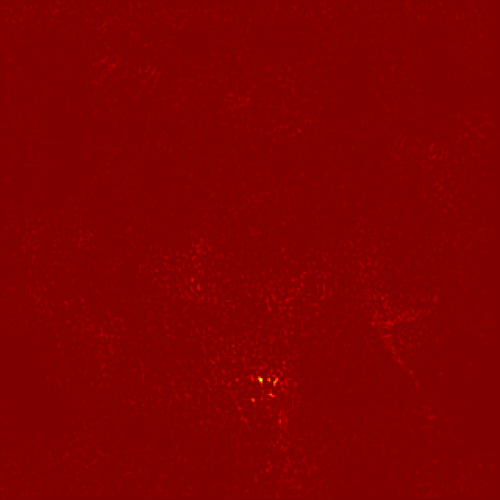} &
      \includegraphics[width=0.07\linewidth]{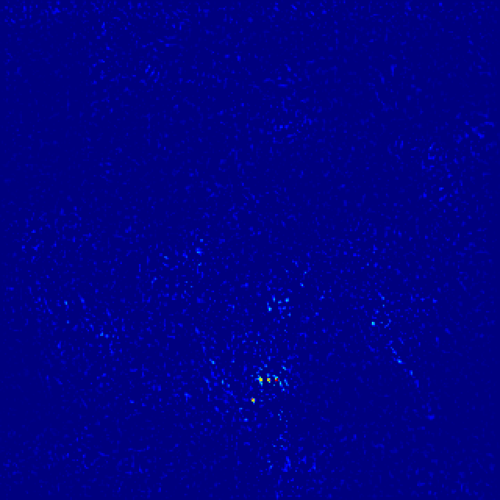} &
      \includegraphics[width=0.07\linewidth]{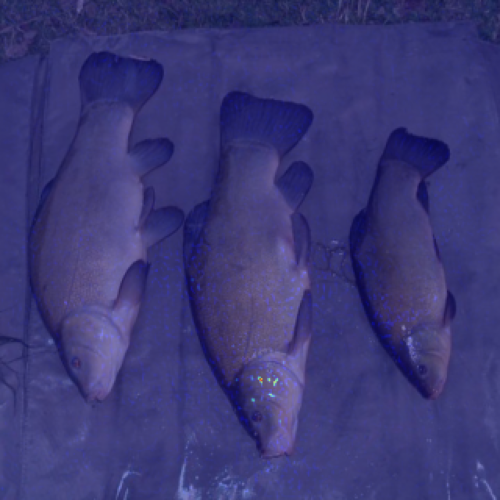} \\
      
      \includegraphics[width=0.07\linewidth]{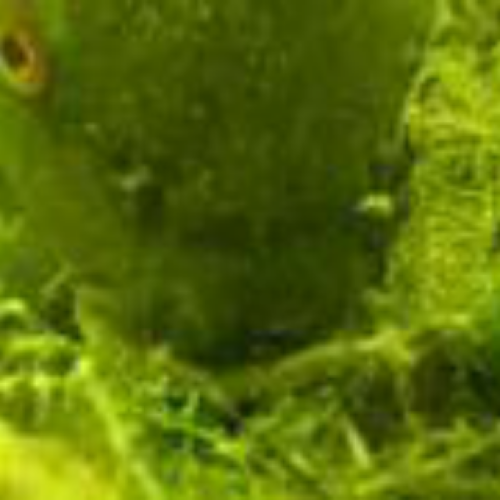} & 
      \raisebox{0.4\height}{\shortstack{$b=0.55$ \\ $v=0.44$ \\ $d=0.00$}} &
      \includegraphics[width=0.07\linewidth]{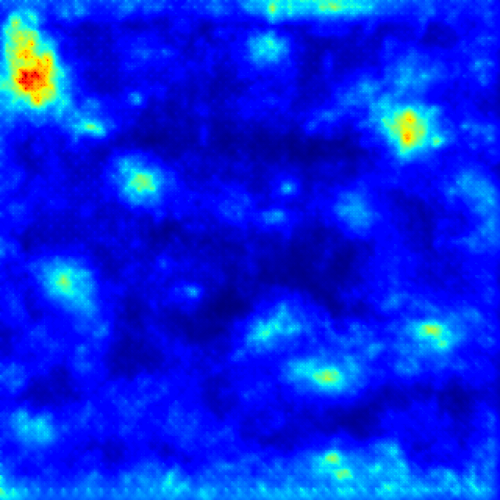} &
      \includegraphics[width=0.07\linewidth]{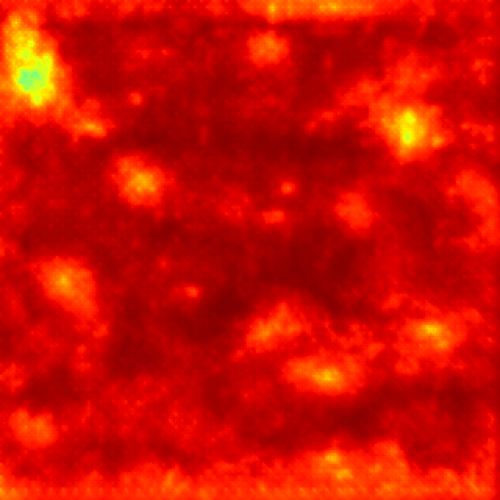} &
      \includegraphics[width=0.07\linewidth]{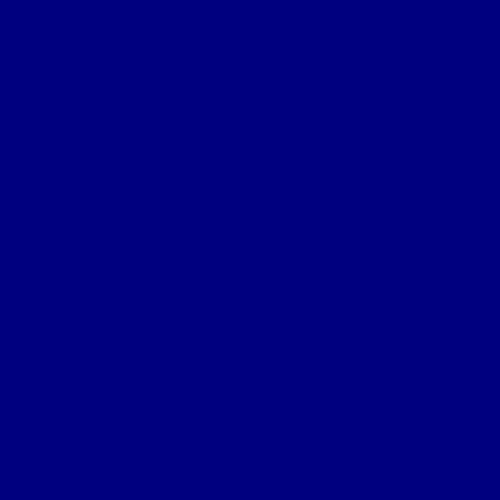} &
      \includegraphics[width=0.07\linewidth]{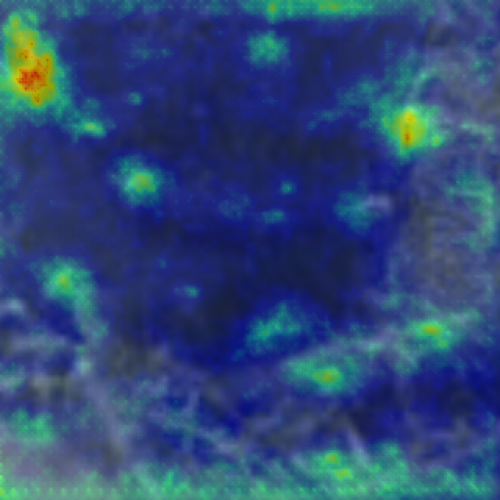} &
      \includegraphics[width=0.07\linewidth]{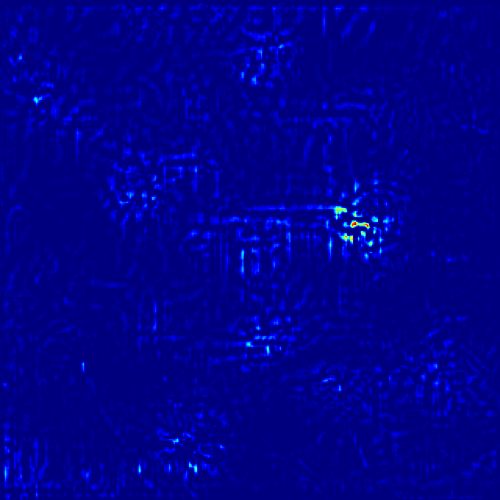} &
      \includegraphics[width=0.07\linewidth]{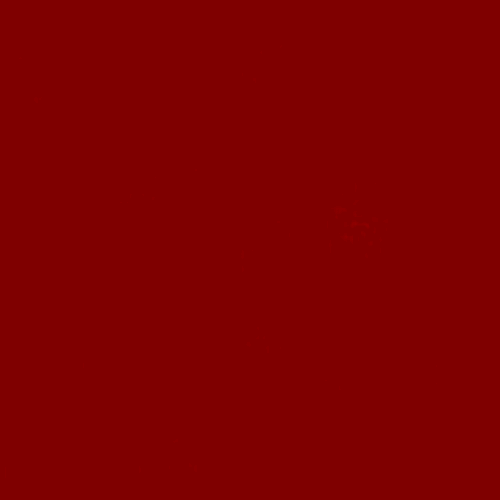} &
      \includegraphics[width=0.07\linewidth]{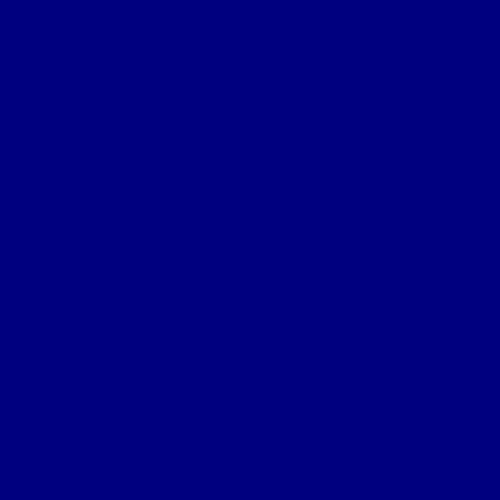} &
      \includegraphics[width=0.07\linewidth]{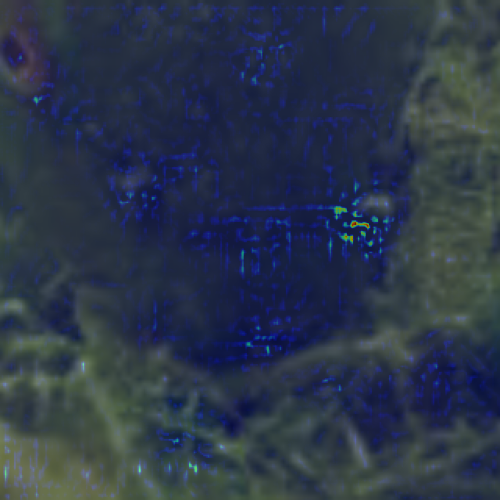} &
      \includegraphics[width=0.07\linewidth]{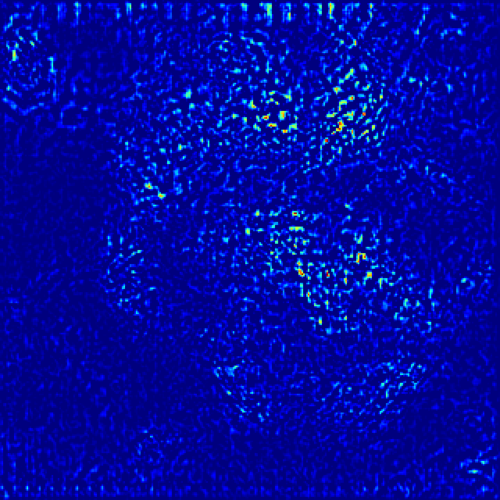} &
      \includegraphics[width=0.07\linewidth]{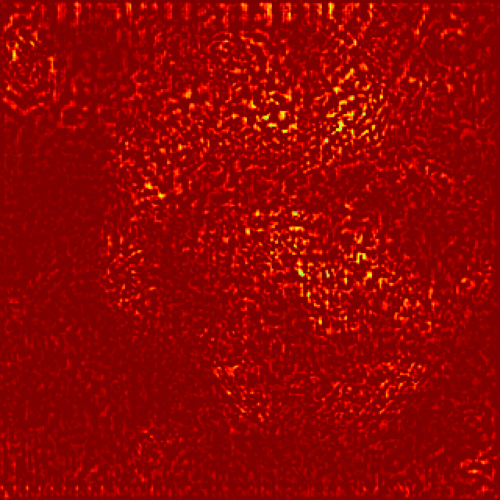} &
      \includegraphics[width=0.07\linewidth]{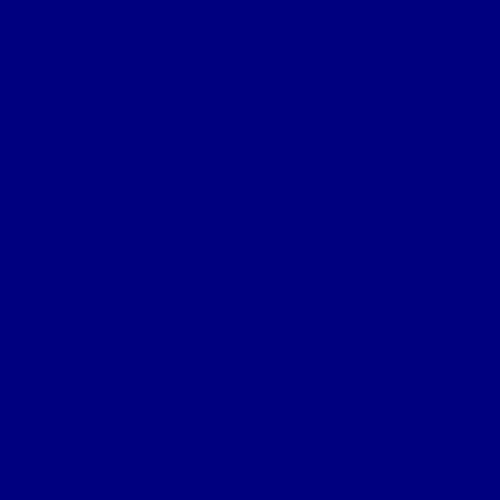} &
      \includegraphics[width=0.07\linewidth]{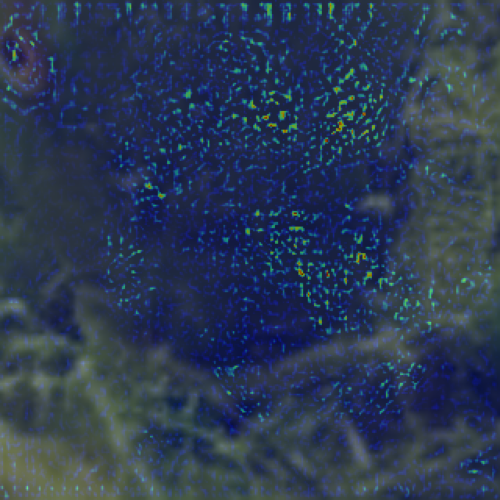} \\
  
      \includegraphics[width=0.07\linewidth]{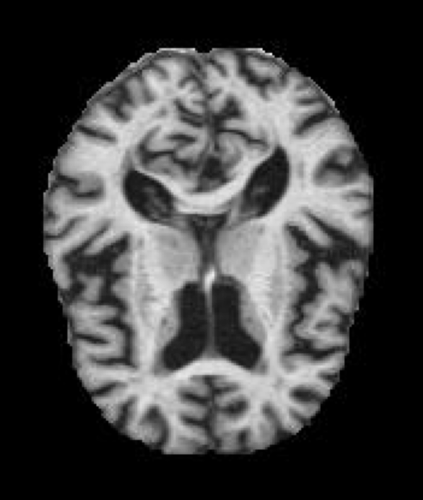} & 
      \raisebox{0.4\height}{\shortstack{$b=0.68$ \\ $v=0.31$ \\ $d=0.00$}} &
      \includegraphics[width=0.07\linewidth]{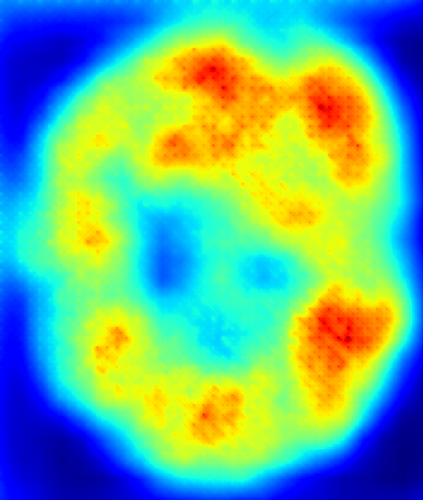} &
      \includegraphics[width=0.07\linewidth]{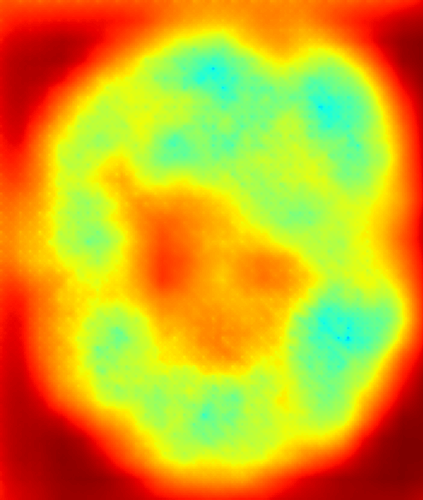} &
      \includegraphics[width=0.07\linewidth]{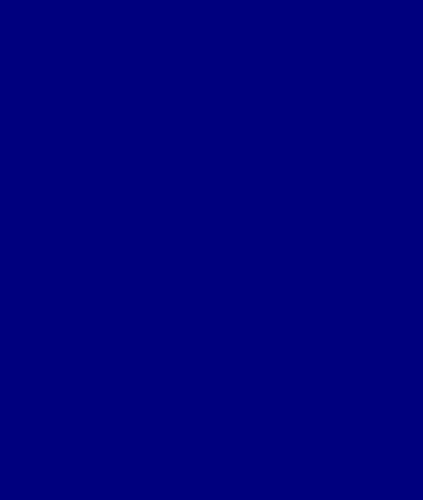} &
      \includegraphics[width=0.07\linewidth]{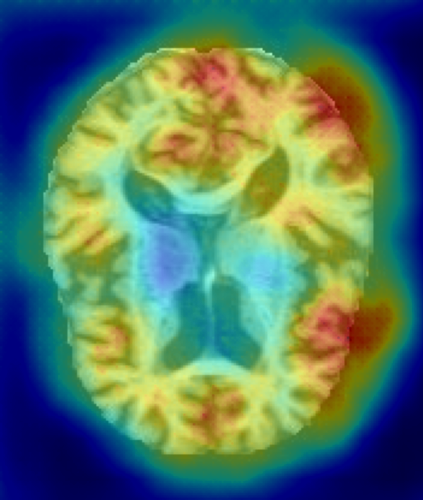} &
      \includegraphics[width=0.07\linewidth]{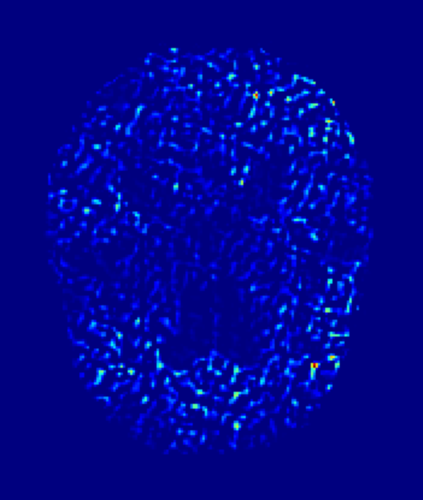} &
      \includegraphics[width=0.07\linewidth]{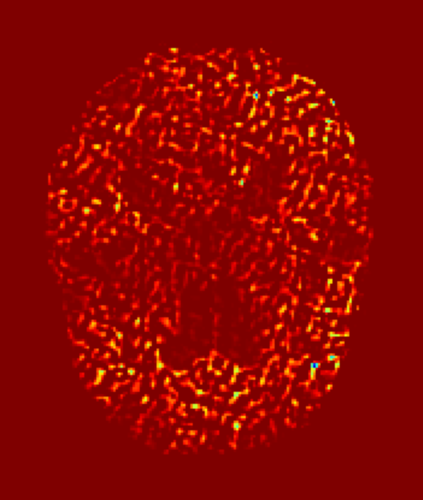} &
      \includegraphics[width=0.07\linewidth]{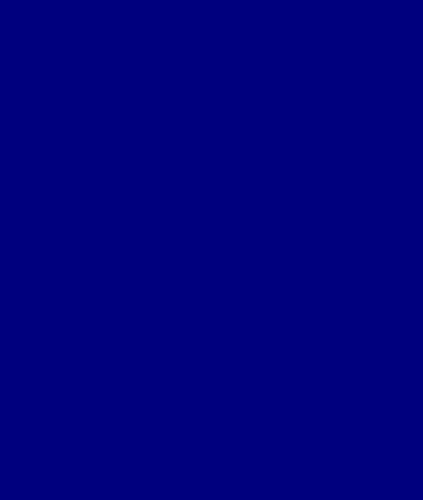} &
      \includegraphics[width=0.07\linewidth]{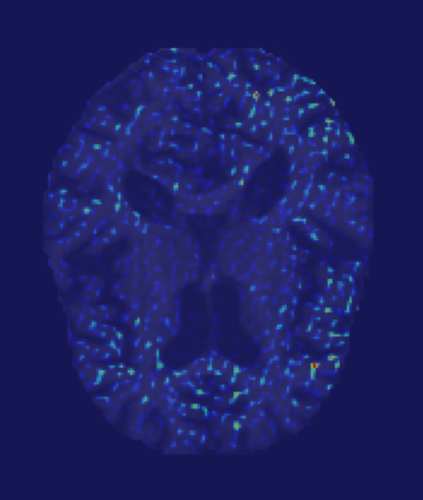} &
      \includegraphics[width=0.07\linewidth]{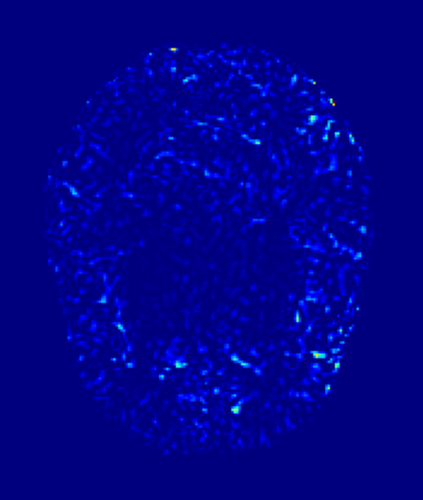} &
      \includegraphics[width=0.07\linewidth]{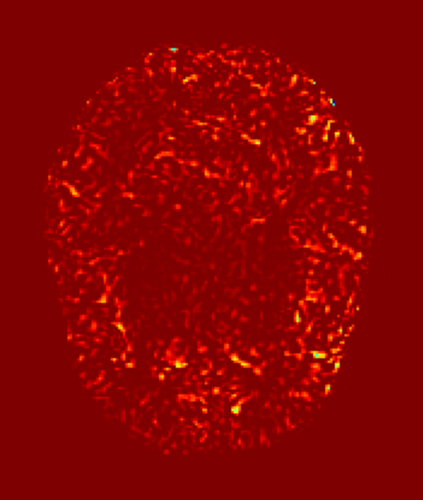} &
      \includegraphics[width=0.07\linewidth]{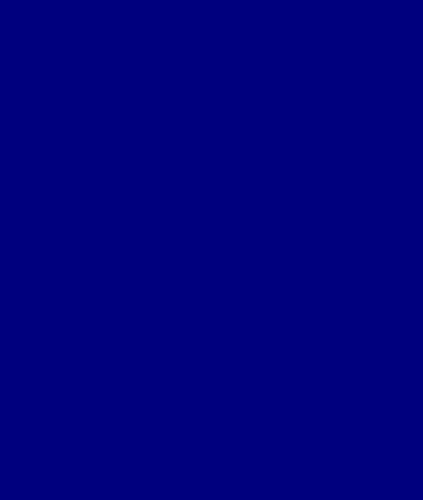} &
      \includegraphics[width=0.07\linewidth]{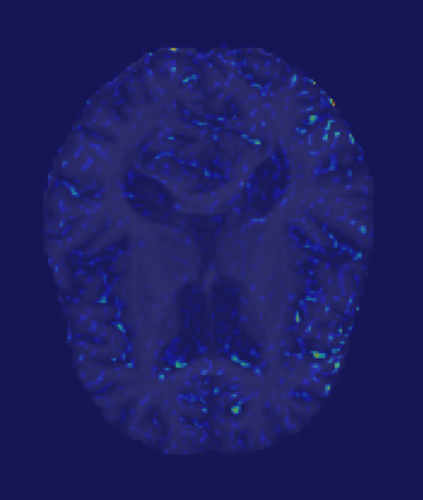} \\

      \includegraphics[width=0.07\linewidth]{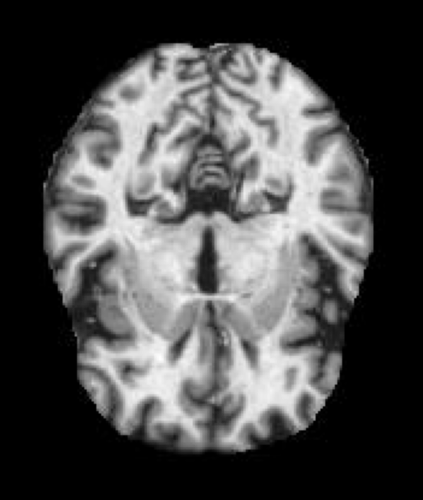} & 
      \raisebox{0.4\height}{\shortstack{$b=0.88$ \\ $v=0.11$ \\ $d=0.00$}} &
      \includegraphics[width=0.07\linewidth]{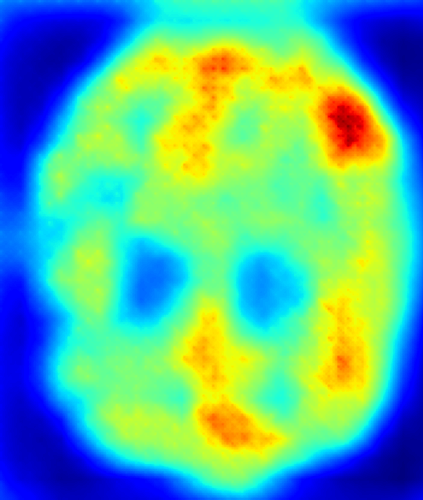} &
      \includegraphics[width=0.07\linewidth]{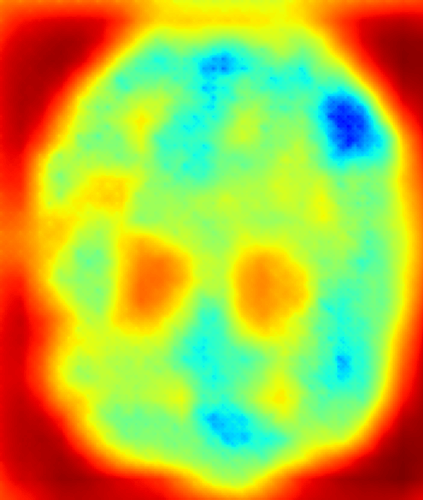} &
      \includegraphics[width=0.07\linewidth]{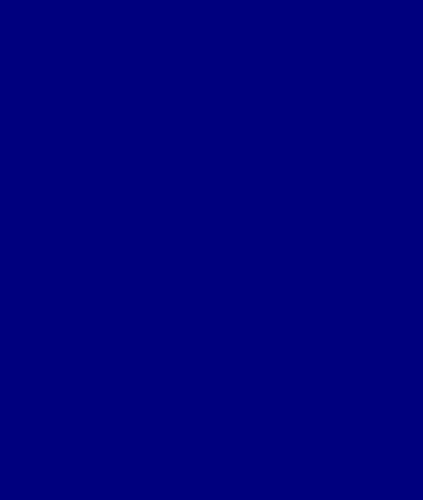} &
      \includegraphics[width=0.07\linewidth]{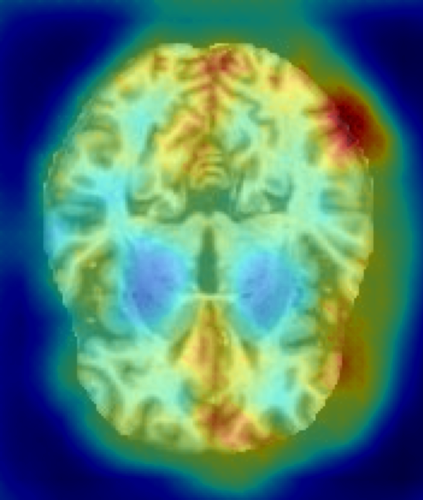} &
      \includegraphics[width=0.07\linewidth]{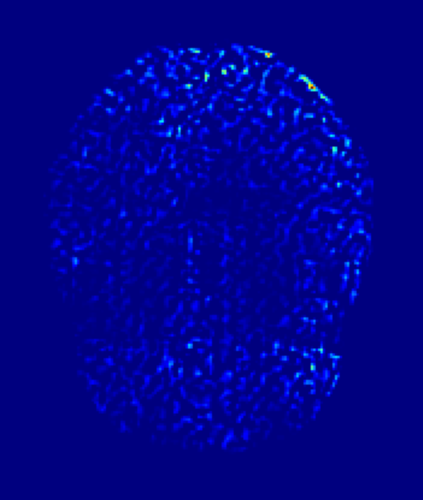} &
      \includegraphics[width=0.07\linewidth]{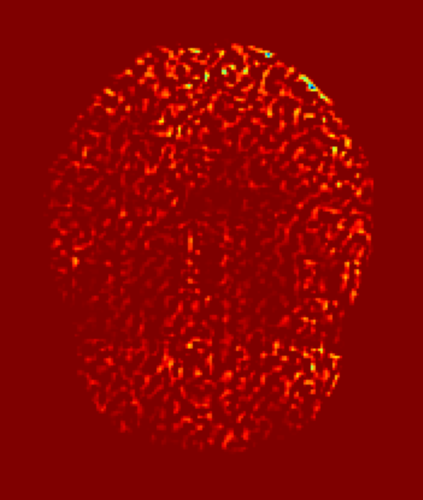} &
      \includegraphics[width=0.07\linewidth]{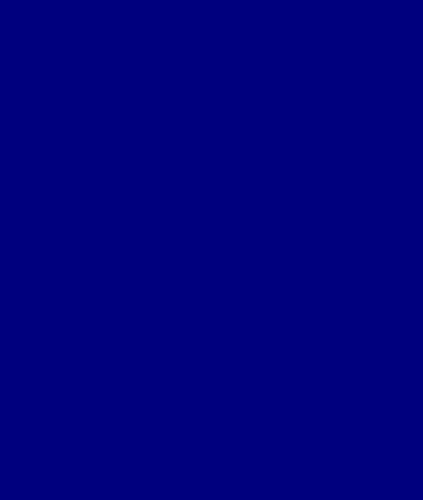} &
      \includegraphics[width=0.07\linewidth]{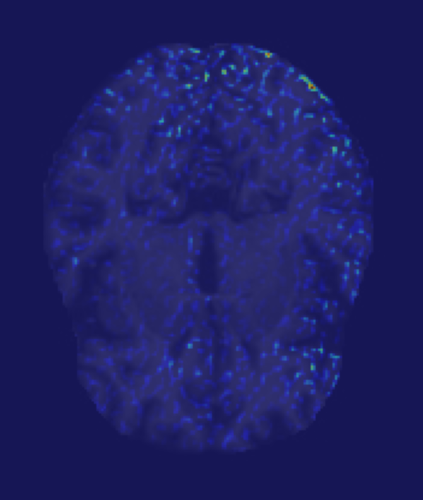} &
      \includegraphics[width=0.07\linewidth]{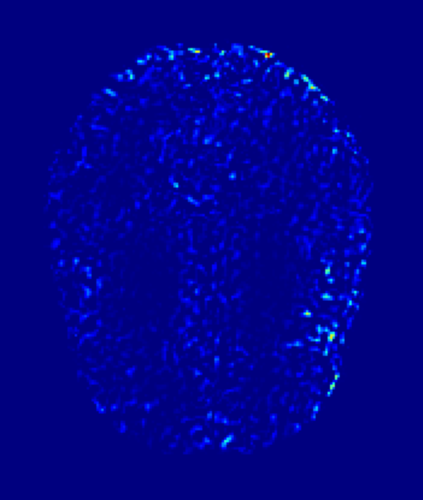} &
      \includegraphics[width=0.07\linewidth]{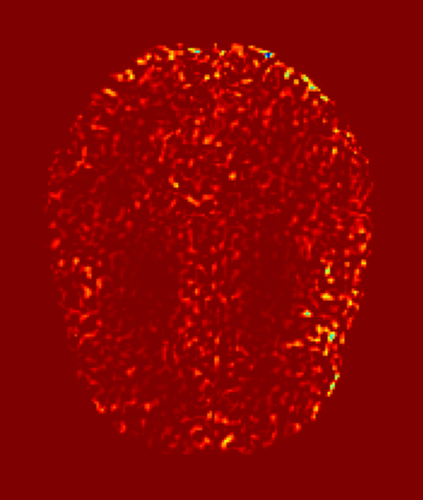} &
      \includegraphics[width=0.07\linewidth]{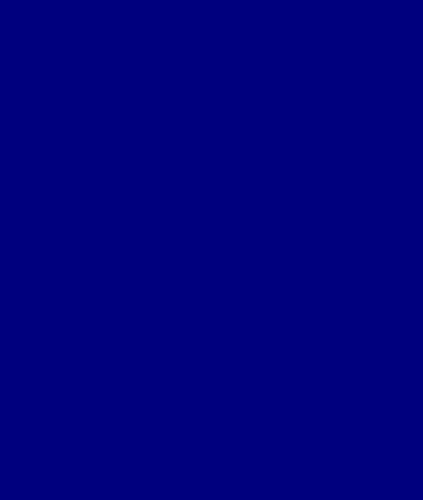} &
      \includegraphics[width=0.07\linewidth]{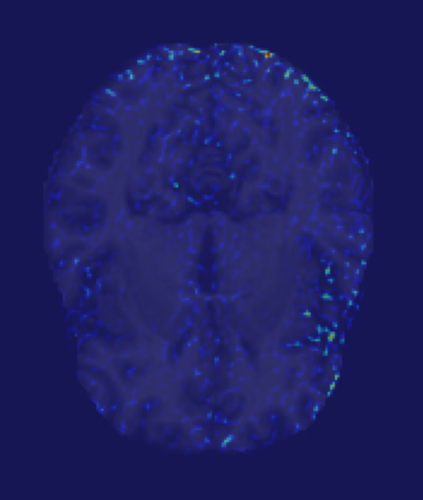} \\

      \includegraphics[width=0.07\linewidth]{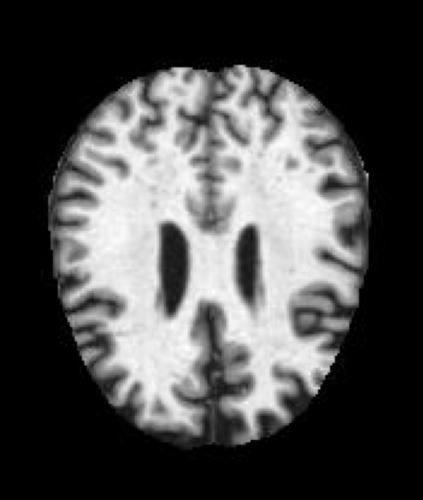} & 
      \raisebox{0.4\height}{\shortstack{$b=0.55$ \\ $v=0.44$ \\ $d=0.00$}} &
      \includegraphics[width=0.07\linewidth]{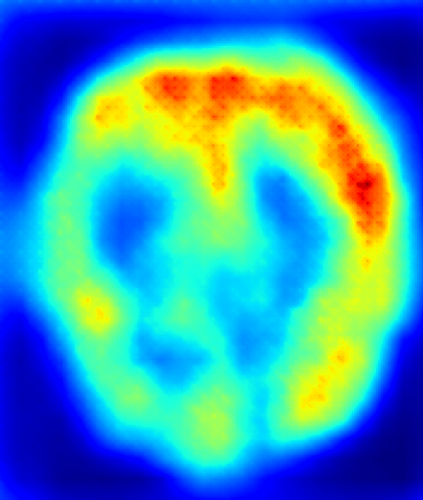} &
      \includegraphics[width=0.07\linewidth]{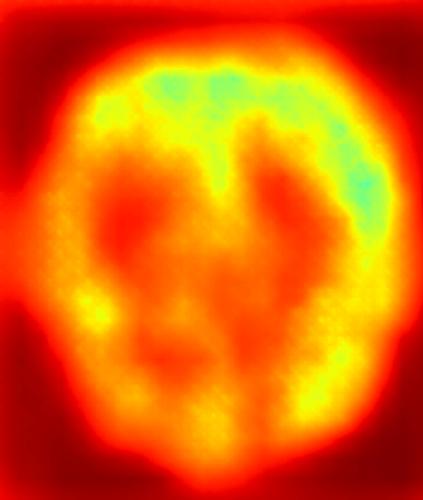} &
      \includegraphics[width=0.07\linewidth]{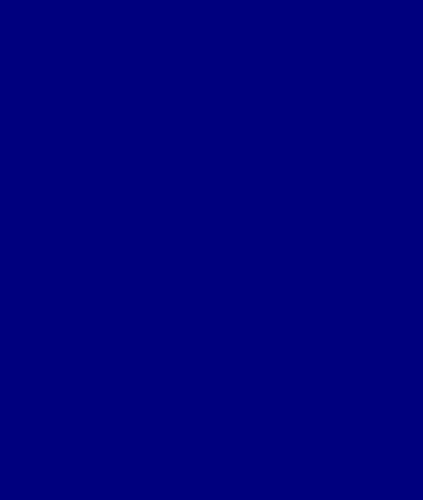} &
      \includegraphics[width=0.07\linewidth]{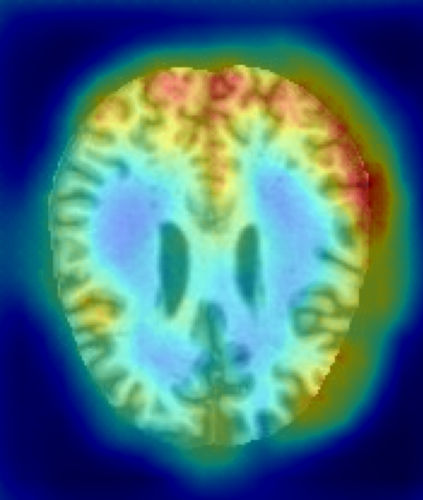} &
      \includegraphics[width=0.07\linewidth]{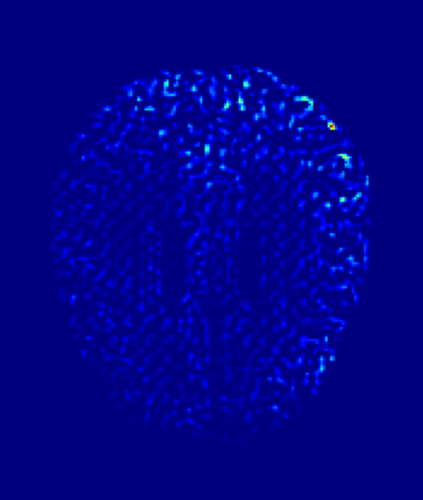} &
      \includegraphics[width=0.07\linewidth]{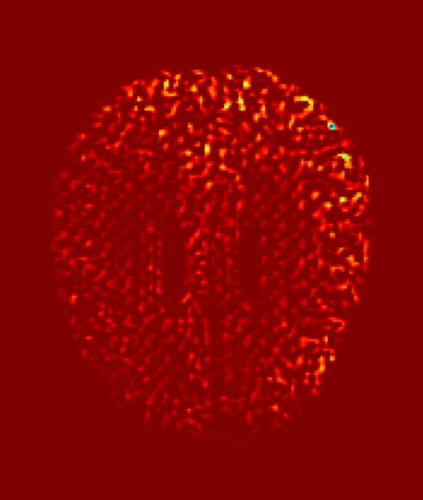} &
      \includegraphics[width=0.07\linewidth]{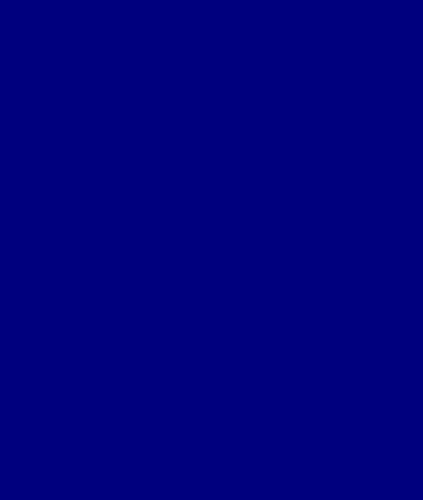} &
      \includegraphics[width=0.07\linewidth]{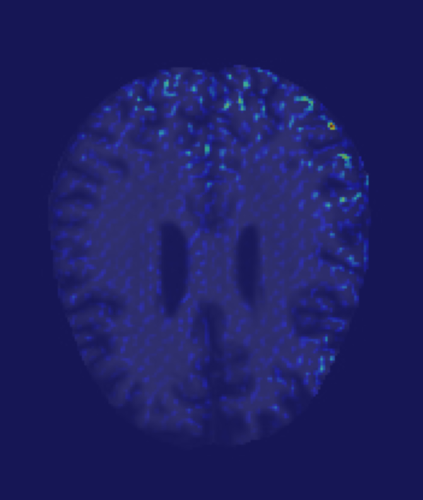} &
      \includegraphics[width=0.07\linewidth]{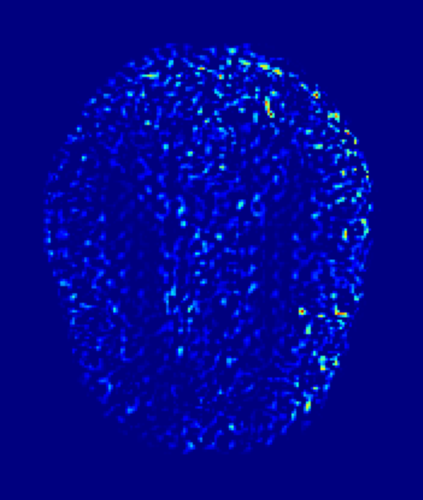} &
      \includegraphics[width=0.07\linewidth]{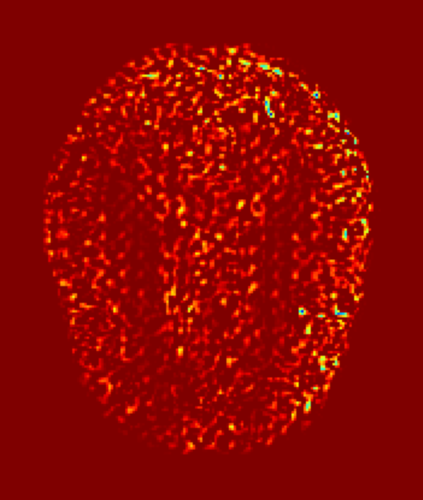} &
      \includegraphics[width=0.07\linewidth]{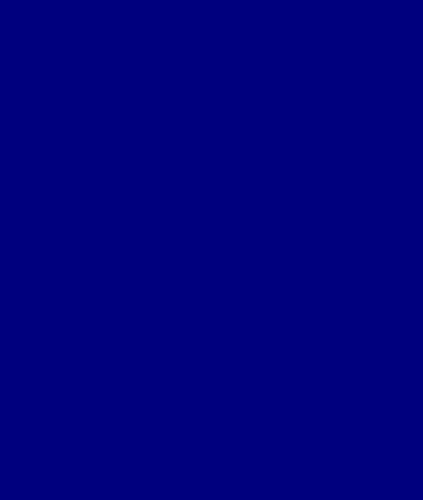} &
      \includegraphics[width=0.07\linewidth]{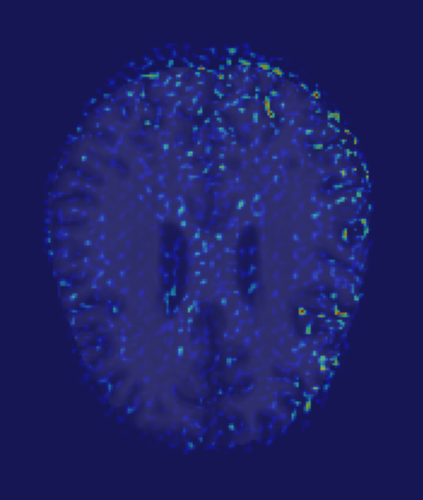} \\
      
  \end{tabular}
  \caption{Comparison of uncertainty visualization across different attribution methods and datasets using models trained with pseudo-OOD sample augmentation. Each row shows a representative sample from MNIST, SVHN, CIFAR-10, Imagenette, and AD datasets. For each sample, columns display: (left to right) original image, belief mass, vacuity, dissonance, and attribution heatmaps from three methods (FullGrad, IG, and SHAP).}
  \label{fig:examples}
\end{figure*}

In this study, we found that using FullGrad is especially effective for visualizing uncertainty in complex visual tasks. For example, when analyzing medical images (i.e., the AD dataset), it can highlight anatomical regions that contribute to diagnostic uncertainty, helping medical professionals understand where the model lacks confidence (see Figure \ref{fig:examples}). Similarly, in autonomous driving applications, the method can identify environmental features (such as unusual lighting conditions or partially occluded objects) that cause uncertainty in object detection systems. It generates more complete attributions, making it especially suitable for uncertainty visualization, as uncertainty often manifests in subtle feature interactions that simpler methods might overlook. By capturing the full context of the model's decision process, our approach provides more interpretable explanations of when and why the model becomes uncertain.

\section{Conclusion and Future Work}
\label{sec:Conclusion}

In this paper, we propose Uncertainty Activation Map (UAM), a unified framework that bridges uncertainty quantification and explainable AI by generating interpretable spatial visualizations of model uncertainty. By combining EDL with FullGrad, our approach produces theoretically grounded uncertainty activation maps that distinguish between vacuity (lack of evidence) and dissonance (conflicting evidence).

The combination of the Vacuity Activation Map and the Dissonance Activation Map provides a comprehensive visual explanation framework. Vacuity highlights regions where the model lacks sufficient evidence, appearing as areas with weak or diffuse attributions across all classes. Dissonance reveals regions with strong but conflicting evidence, where multiple high-confidence classes compete for attention. This distinction is particularly valuable for analyzing out-of-distribution inputs and ambiguous examples near class boundaries. Our extensive evaluation demonstrates that the UAM framework effectively localizes uncertainty sources across diverse visual domains. A comparative analysis showed that FullGrad's theoretical completeness provides the most reliable foundation, particularly in out-of-distribution scenarios.

Future work includes extending UAM to region-level uncertainty estimation using segmentation-based local inference, incorporating temporal dynamics for video analysis, adapting the framework to architectures beyond CNNs, and exploring additional modalities such as natural language processing and multimodal learning. Another promising direction is integrating uncertainty-aware attribution into active learning and human-AI collaborative decision-making systems for safety-critical applications.

\section*{Acknowledgments}
This work is supported by NSF under grants 2107449, 2107450, and 2107451. The research was sponsored by the Army Research Office and was accomplished under Grant Number W911NF-23-1-0217. The views and conclusions contained in this document are those of the authors and should not be interpreted as representing the official policies, either expressed or implied, of the Army Research Office or the U.S. Government. The U.S. Government is authorized to reproduce and distribute reprints for Government purposes notwithstanding any copyright notation herein.



\bibliographystyle{unsrt}  
\bibliography{bibliography}

\end{document}